\documentclass{article}

 \usepackage[main, final]{neurips_2026}

\usepackage[utf8]{inputenc} 
\usepackage[T1]{fontenc}    
\usepackage{hyperref}       
\usepackage{url}            
\usepackage{booktabs}       
\usepackage{amsfonts}       
\usepackage{nicefrac}       
\usepackage{microtype}      
\usepackage{xcolor}         

\usepackage{subcaption}
\usepackage{graphicx}
\usepackage{float}
\usepackage{amsmath}
\title{Latent Video Prediction for World Modeling: \\ An Evaluation Uncovering Intriguing Favorable Evidence}

\author{%
  Ali J Alrasheed$^{1}$\thanks{Corresponding author: \texttt{aalrasheed@student.unimelb.edu.au}} \quad
  Aryan Yazdan Parast$^{1}$ \quad
  Basim Azam$^{1}$ \quad
  James Bailey$^{2}$ \quad
  Naveed Akhtar$^{1}$ \\
  $^{1}$The University of Melbourne \quad $^{2}$Monash University
}

\begin{document}

\maketitle
\vspace{-2mm}
\begin{abstract}
Self-supervised video models are increasingly framed as world models, yet they are still evaluated almost entirely on clean video and reported as a final task score, obscuring how their representations behave under the degraded and ambiguous conditions a deployed world model must handle. We present the first systematic study of this component, analyzing four matched-capacity frontier self-supervised learning models that are strong candidates for world-model encoders—V-JEPA 2.1, V-JEPA 2, VideoPrism, and VideoMAEv2—across five robustness axes relevant to their deployment as video world models: feature discriminability, corruption robustness, fine-grained discrimination, occlusion robustness, and sensitivity to temporal direction. We run the study on Something-Something\,v2 and repeat every axis that ports on the egocentric EGTEA Gaze+, where the model ordering is unchanged. Our results reveal a distinct and consistent profile for latent-prediction models across all five axes. They degrade more gracefully under pixel corruption, preserve usable class structure rather than mere geometric stability under occlusion, capture fine-grained physical contact cues without reconstructing pixels, and uniquely encode the arrow of time. We finally test whether these representation-level differences matter for downstream world modeling, pairing each frozen encoder with an identical action-conditioned predictor and planner in a simulated manipulation environment. Only latent prediction yields near-complete task success, remains effective under degraded observations, and transfers to a manipulation task for which the predictor was never trained. Our results provide concrete new evidence that latent prediction is a promising foundation for robust world modeling.
\end{abstract}
\vspace{-3mm}
\section{Introduction}
\vspace{-2mm}
Recent self-supervised video models are increasingly positioned as world foundation models, systems that learn the dynamics of the physical world from observation. Such a system has two learned parts: an \emph{encoder} that maps observations to latent states, and a \emph{predictor} that models how those states evolve. Self-supervised video models supply the encoder, and its pretraining objective is therefore a central design choice for world modelling. The Joint Embedding Predictive Architecture (JEPA)~\citep{lecun2022path} is central to this shift. Rather than reconstructing raw pixels, JEPA learns to predict in a learned latent space, motivated by the hypothesis that latent prediction captures the causal structure of the world more effectively than pixel reconstruction.

What does it mean for such a representation to understand the world? Video understanding benchmarks offer retrieval, localisation, question answering and anticipation benchmarks, yet across all of them the video is clean and the outcome is reported as a final task score only~\citep{assran2025v, mur2026v, wang2023videomae, zhao2024videoprism}; and when these encoders are placed inside world models, the reported result is usually whether the system reached its goal. Neither shows how a representation behaves under the degraded and ambiguous conditions a deployed world model has to handle, e.g., stability under sensor noise and environmental change, tolerance to missing input, fine-grained discrimination between visually similar actions, and an internal sense of temporal order. A comparable evaluation at the level of pretraining objectives also remains missing for such models.

We address this gap through a controlled study of four public video models, V-JEPA 2.1~\citep{mur2026v}, V-JEPA 2~\citep{assran2025v}, VideoPrism~\citep{zhao2024videoprism}, and VideoMAEv2~\citep{wang2023videomae}, spanning the three dominant self-supervised paradigms: latent prediction, contrastive plus masked prediction, and pixel reconstruction. To isolate the pretraining objective, we hold three axes fixed: backbone capacity, evaluation data and protocol, and a single attentive-probe configuration of identical capacity for every encoder, so that no model is read out through a larger or better-matched head than another. Remaining factors are fixed by each public release and cannot be equalised without retraining from scratch under a shared compute budget. We further evaluate V-JEPA 1~\citep{bardes2023v}, which matches V-JEPA\,2 in architecture, capacity, objective and readout but saw an order of magnitude less pretraining data, as a direct control on data scale. Sections~\ref{sec:encoder-discriminability}--\ref{sec:temporal} measure the encoder in isolation; Section~\ref{sec:downstream} then attaches an identical action-conditioned predictor to each frozen encoder and plans with it, testing whether the representation-level differences carry through to behaviour.

Through this systematic evaluation across six major experiments that consumed approximately 2,000 A100 GPU-hours, we uncover the following major intriguing insights:
\vspace{-2mm}
\begin{figure}[t]
\centering
\begin{minipage}{0.6\linewidth}
    \centering
    \includegraphics[width=\linewidth]{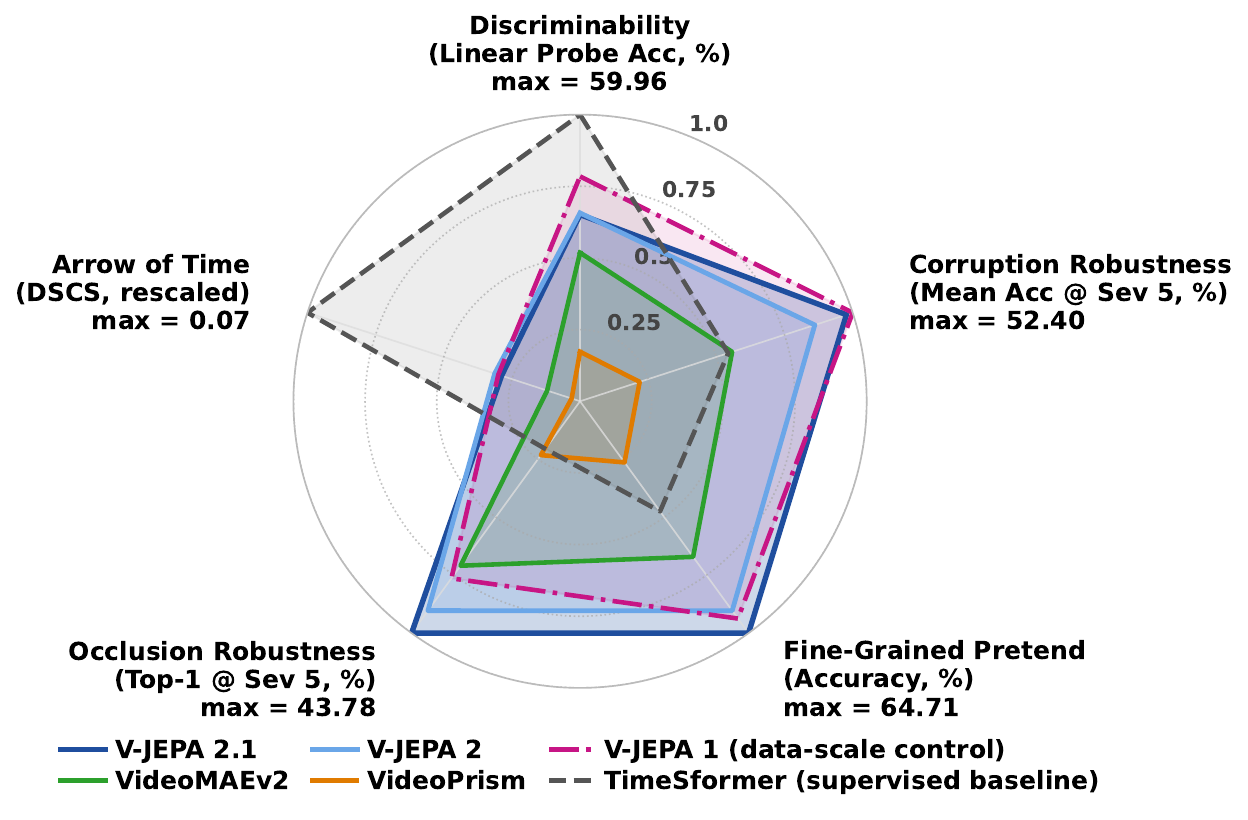}
\end{minipage}\hfill
\begin{minipage}{0.36\linewidth}
    \caption{{Multi-axis robustness profile on Something-Something v2~\citep{goyal2017something}. Higher normalised scores indicate stronger performance; raw maxima are annotated. V-JEPA variants show the most consistently balanced profiles, while TimeSformer’s competitive clean performance does not translate to corruption and occlusion robustness. V-JEPA 1 closely matches V-JEPA 2 despite substantially less pretraining data (Appendix~\ref{app:datascale}). Figures~\ref{fig:discrim}--\ref{fig:temporal_reversal} report the four primary encoders.}}
    
    \label{fig:teaser-radar}
\end{minipage}
\end{figure}

\begin{itemize}
    \item \textbf{Latent-prediction models are substantially more robust under pixel corruption.} V-JEPA\,2.1 leads on five of six corruption types and shows the slowest accuracy degradation across severities, consistent with the joint-embedding predictive objective discarding surface-level visual variation in favour of higher-order semantic structure.
    \item \textbf{Stable features are not the same as usable features for these models.} Under the most severe spatiotemporal patch dropout, VideoPrism maintains representational similarity above 0.98 yet collapses to 2.7\% top-1 accuracy, while V-JEPA\,2.1 retains 46.1\% on the same clips, showing that cosine similarity alone is a misleading measure of representational stability.
    \item \textbf{Latent prediction captures fine-grained visual cues without reconstructing pixels.} V-JEPA outperforms the pixel-reconstruction baseline on virtually every class of  `pretend actions', with the largest margins precisely on high-sensitivity classes whose discriminative signal is the absence of actual physical contact.
    \item \textbf{Latent-prediction models internalise the arrow of time substantially more strongly than the alternatives.} Under video reversal, V-JEPA models are the most likely to flip predictions to semantically antonymous classes such as pushing and pulling, achieving a Directional Semantic Coherence Score (DSCS) $2.4\times$ that of VideoMAEv2 and $9\times$ that of VideoPrism.
    \item \textbf{The advantage carries through to downstream world modelling.} With an identical action-conditioned predictor and planner on every frozen encoder, both V-JEPA models drive the planner to near-complete success while the pixel-reconstruction and contrastive encoders fail most episodes. Under corruption and occlusion only latent prediction sustains a usable success rate, and it is the only objective whose learned dynamics carry over substantially to a task it was not trained on.
\end{itemize}
\section{Related work}
\paragraph{Video self-supervised learning and world models.}
Self-supervised learning (SSL) encompasses several paradigms aimed at learning useful representations from unlabeled data. Contrastive methods~\citep{chen2020simple,grill2020bootstrap} learn view-invariant features but capture limited temporal dynamics. Generative methods~\citep{he2022masked,tong2022videomae} shift the objective to pixel reconstruction, encouraging models to capture the underlying structure of the data. More recently, Joint-Embedding Predictive Architectures (JEPA)~\citep{lecun2022path} argue that prediction should occur in learned latent spaces rather than pixel space, a philosophy instantiated for images in I-JEPA~\citep{assran2023self}, extended to video in V-JEPA~\citep{bardes2023v}, and scaled in V-JEPA 2~\citep{assran2025v} to over one million hours of internet video. This philosophy connects to the broader ambition of learning internal world models that support planning and physical reasoning~\citep{ha2018world,hafner2019dream,hafner2020mastering,hafner2023mastering,hafner2025training}. On the generative side, Cosmos~\citep{agarwal2025cosmos} and Genie 2~\citep{parker2024genie} demonstrate open-weight world foundation models and action-controllable environment generation, respectively. DINO-WM~\citep{zhou2024dino} also showed that predicting future states using latent-prediction principle enables zero-shot planning.
\paragraph{Robustness evaluation.}
Prior work on image transformers has documented substantial accuracy retention under severe occlusion and distribution shift, attributed to content-dependent global attention rather than texture bias~\citep{naseer2021intriguing,paul2022vision}, grounded in the corruption benchmarking framework of~\citet{hendrycks2019benchmarking}. \citet{caron2021emerging} further demonstrated that self-supervised ViTs develop emergent segmentation capabilities, establishing that the self-supervised objective shapes structural properties beyond task performance. However, most relevant to latent prediction, \citet{garrido2025intuitive} showed that V-JEPA acquires intuitive physics understanding through representation-space masked prediction alone, while pixel-reconstruction models perform near chance on the same benchmarks. \citet{joseph2026interpreting} provided the first interpretability study of physics representations inside V-JEPA 2 and VideoMAEv2, finding that latent prediction gives rise to qualitatively different internal organisation. On the temporal axis, prior work has collectively established that video models frequently rely on static appearance cues and that deeper layers progressively lose temporal order information~\citep{sevilla2021only, huang2018makes, yun2022time}. Despite this body of work, a comparative evaluation of how distinct video self-supervised paradigms respond to different realistic perturbations remains absent from the literature.
\section{Evaluation framework}

A representation intended for use as a video world model must do more than
classify clean clips correctly. It must remain reliable under sensor noise
and environmental change, tolerate missing input, distinguish visually similar
actions that differ only in fine physical detail, and capture how a scene
evolves over time, including the direction in which events unfold. Rather
than establishing a new clean-accuracy ranking, our goal is to characterise
how each pretraining paradigm \emph{degrades} along these axes, and whether
the choice of objective leaves a measurable structural signature that
current benchmarks miss. To this end, we hold model capacity, readout
protocol, and evaluation data fixed across all experiments, so that the
pretraining objective is the principal remaining difference between models
(release-specific factors are discussed in Appendix~\ref{app:limitations}). We describe the models and dataset before
summarising the five experiments and the findings.
\paragraph{Models.}
We compare four publicly available video foundation models at matched ViT-L capacity, each kept frozen and read out through a lightweight attentive probe~\citep{assran2023self}. The four models span the three dominant self-supervised video paradigms: joint-embedding latent prediction (V-JEPA 2.1, V-JEPA 2), contrastive plus masked prediction (VideoPrism), and pixel reconstruction (VideoMAEv2); their capacities, pretraining corpora and sources are detailed in Appendix~\ref{app:model_comparison}. In Appendix~\ref{app:training_regime}, we additionally run a competitiveness probe that asks whether frozen V-JEPA 2 features match or exceed an end-to-end fine-tuned VideoMAE and a fully supervised TimeSformer trained specifically for the task. This second comparison is restricted to public checkpoints, which dictate  the baseline choices included in the evaluations.
\paragraph{Dataset.}
Because our objective is to measure how performance degrades under controlled perturbations rather than to rank models on a new clean benchmark, the controlled variable is the perturbation, not the data distribution. For our evaluation, what matters is a benchmark whose class structure is rich enough to expose the failure modes of interest. Something-Something v2 (SSv2)~\citep{goyal2017something} meets this requirement. This large-scale dataset comprises over 220{,}000 videos across 174 fine-grained human-object interaction categories, including visually similar but temporally distinct pairs such as ``pushing something left'' versus ``pushing something right'' that force genuine temporal discrimination, and the antonym structure required for the pretend-action and reversal probes. To confirm that the findings are not specific to one benchmark, every axis that ports was re-run on EGTEA Gaze+~\citep{li2018eye}, an egocentric kitchen-activity dataset with 106 action classes, using the same frozen encoders, the same probe and the same perturbation code; the model ordering is unchanged on every axis, and Appendix~\ref{app:egtea} reports the full protocol and results.
\vspace{-3pt}
\paragraph{Evaluation axes.}
In our experiments, we probe five complementary axes of representation quality:
\begin{itemize}
  \item \textbf{Feature Discriminability.} Separability of representations on 600 videos drawn from 30 classes (20 videos each), spanning different levels of semantic difficulty.
  \item \textbf{Corruption Robustness.} Classification accuracy under six
        ImageNet-C corruptions~\citep{hendrycks2019benchmarking} (motion blur,
        snow, pixelation, impulse noise, brightness, and elastic transform)
        at three severity levels on a class-balanced set of 500 videos.
  \item \textbf{Fine-Grained Action Discrimination.} Performance on the
        pretending subset (1,992 videos, 22 classes), where distinguishing
        real from simulated physical interactions requires sensitivity to
        fine spatiotemporal contact cues.
  \item \textbf{Occlusion Robustness.} Three spatiotemporal occlusion
        paradigms at three severity levels on a 1,720-video balanced set
        spanning  174 classes.
  \item \textbf{Temporal Robustness.} Four temporal disruption paradigms
        (frame shuffling, reversal, static replacement, and noise injection)
        on a 1,738-video balanced set spanning 174 classes, measuring sensitivity to disrupted order and to
        the direction of time.
\end{itemize}

\paragraph{Outline of the results.}
The sections below report result and analysis following the experiment order listed above: feature discriminability (Section~\ref{sec:encoder-discriminability}), corruption robustness (Section~\ref{sec:corruption}), pretend versus real action discrimination (Section~\ref{sec:pretend-actions}), occlusion robustness (Section~\ref{sec:occlusion}), and temporal robustness (Section~\ref{sec:temporal}). These five sections isolate the encoder. Section~\ref{sec:downstream} then extends the evaluation to downstream world-modeling tasks, testing whether the observed advantage is retained in a more task-oriented setting. Two appendices test the alternative explanations: Appendix~\ref{app:datascale} asks whether pretraining scale rather than the objective drives the result, and Appendix~\ref{app:training_regime} whether the frozen advantage survives when the baselines are allowed to fine-tune end-to-end on the same task. The unifying question across all of these evaluations is whether the choice of pretraining objective leaves a measurable structural signature in the base representation that single-score benchmarks miss.

\section{Do latent-prediction models produce better representations?} \label{sec:encoder-discriminability}

\begin{figure}[t]
    \centering
    \includegraphics[width=\textwidth]{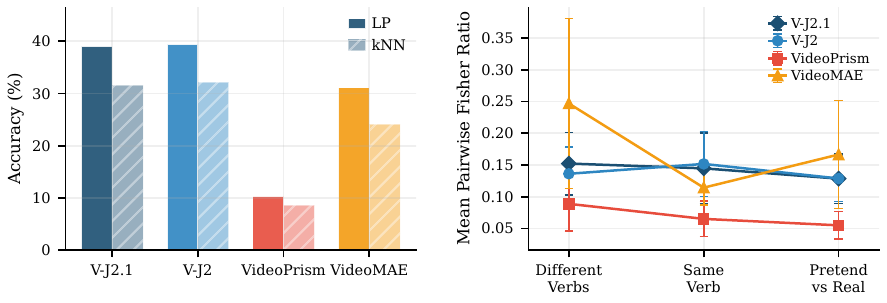}
    \caption{Encoder discriminability on  semantically stratified action classes using frozen GAP features (higher is better in both panels).
    \textbf{(Left)}~Linear probe and kNN accuracy.
    \textbf{(Right)}~Fisher discriminant ratio across three semantic categories ordered by increasing difficulty: \emph{different-verb} (distinct actions, e.g.,\ folding vs.\ stacking), \emph{same-verb} (shared verb differing only in direction or object, e.g.,\ pushing left-to-right vs.\ right-to-left), and \emph{pretend-vs-real} (visually near-identical pairs, e.g.,\ pouring vs.\ pretending to pour). Lines connect category-level means; error bars show standard deviation.}
    \label{fig:discrim}
\vspace{-8pt}
\end{figure}

We measure the discriminability of frozen video representations using global average pooled (GAP) features across the four frozen encoders. The 30 selected action classes are listed in Appendix~\ref{app:intrinsic}, stratified into three tiers of increasing semantic difficulty, i.e., different-verb actions, same-verb actions, and pretend-vs-real,  chosen so that successive tiers progressively remove the coarse appearance and motion cues a representation could exploit, isolating progressively finer-grained semantic structure.
\paragraph{V-JEPA models produce the most  discriminable   representations.}
Both V-JEPA variants achieve substantially higher linear probe accuracy than VideoMAEv2 and VideoPrism across the  semantically stratified action classes - Figure~\ref{fig:discrim} (left). Although VideoMAEv2 attains the highest global Fisher discriminant ratio - Figure~\ref{fig:discrim} (right), this does not translate into superior classification accuracy, indicating that inter-class variance alone is insufficient for linear separability. The gap between linear probe and kNN accuracy is largest for the V-JEPA models (Appendix~\ref{app:intrinsic}, Figure~\ref{fig:app_attribution}), suggesting that their representations encode direction-dependent semantic structure that a learned linear head can exploit more effectively than a nearest-neighbor classifier.
\paragraph{V-JEPA separability is consistent  across semantic categories; VideoMAE is not.}
Per-category Fisher analysis shows that V-JEPA models maintain a almost consistent  separability profile as difficulty increases from different-verb classes (coarse motion and appearance cues available) through same-verb pairs (shared verb, distinguishable only by direction or object) to pretend-vs-real pairs (separable only by fine contact cues)  - Figure~\ref{fig:discrim} (right). In contrast, VideoMAEv2 peaks on different-verb classes but drops sharply on same-verb pairs, with only partial recovery on pretend-vs-real pairs. It also exhibits higher standard deviation within the different-verb category, indicating less consistent separability across classes. Together, these results support the conclusion that V-JEPA representations are more uniform across semantic regimes. This pattern aligns with the hypothesis that joint-embedding predictive objectives capture higher-level semantic structure, whereas masked pixel-reconstruction objectives remain more sensitive to low-level visual distinctiveness. The partial recovery on pretend-vs-real pairs in VideoMAEv2 may reflect exaggerated motion cues in pretending actions that provide distinctive visual signatures despite semantic ambiguity.

\vspace{-6pt}

\section{Does latent-prediction optimize for meaning over appearance?}
\label{sec:corruption}

Section~\ref{sec:encoder-discriminability} established that V-JEPA features exhibit consistent class separability under clean conditions. We now extend our evaluations to six practical corruptions (motion blur, snow, pixelation, impulse noise, brightness, and elastic transform) at three severity levels defined in~\citet{hendrycks2019benchmarking}. Refer to  Figure~\ref{fig:corruption-examples} for representative visualizations of the employed input corruptions.

\begin{figure}[t]
\centering
\includegraphics[width=\linewidth]{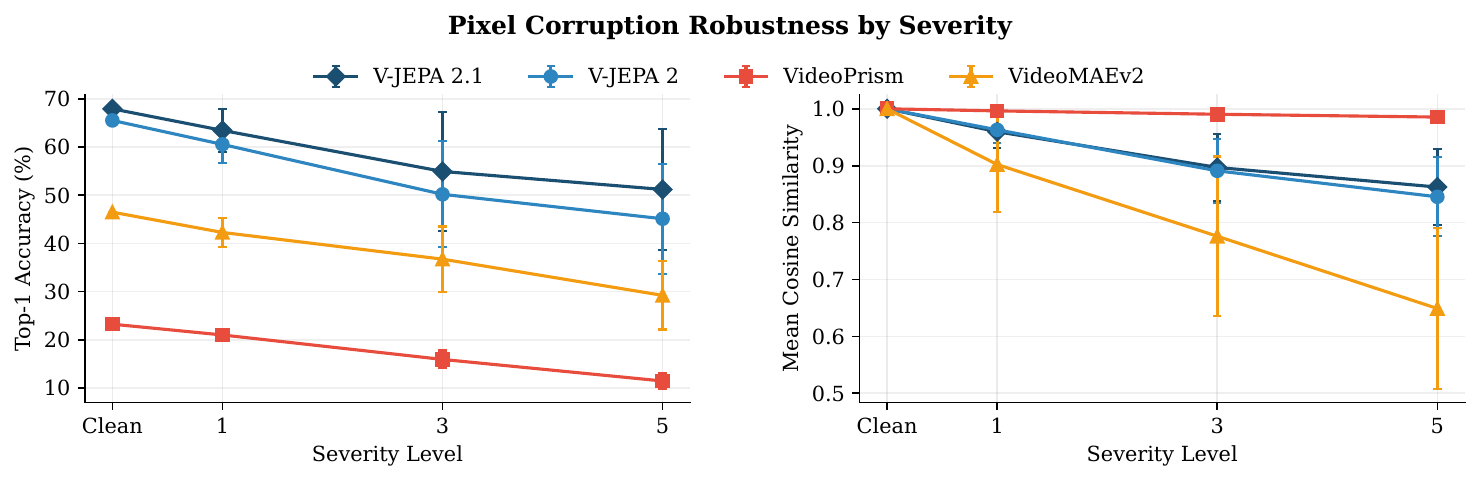}
\caption{Representational stability as a function of corruption severity, averaged across all six corruption types. \textbf{(Left)} Top-1 classification accuracy. \textbf{(Right)} cosine similarity. }
\label{fig:severity-curves}
\end{figure}

\paragraph{V-JEPA encodes meaning over appearance, yielding systematic robustness under corruption.}
Figure~\ref{fig:severity-curves} (left) presents the aggregate degradation trajectories across all corruption types. V-JEPA 2.1 maintains the highest classification accuracy at every severity level and degrades most gradually, with V-JEPA 2 following closely behind. VideoMAEv2 and VideoPrism exhibit steeper declines, with the ordering among all four models preserved across severity levels. Per-corruption analysis (Appendix~\ref{app:corruption}, Figure~\ref{fig:retention-heatmap-app}) confirms that V-JEPA 2.1 achieves the highest accuracy retention on five of six corruption types, with the largest margins on corruptions that alter pixel statistics while preserving action semantics, such as impulse noise, snow, and brightness. V-JEPA 2.1 also outperforms V-JEPA 2 on those same five corruption types and degrades more slowly than it on all six, indicating that the architectural and training refinements in the newer release enhance representational quality under distributional shift beyond clean accuracy gains alone. This resilience suggests that the joint-embedding predictive objective learns to discard surface-level visual variation in favor of higher-order temporal and semantic structure. Because V-JEPA's loss compares learned abstract representations rather than raw pixel values, two inputs that differ only in surface statistics (e.g.\ added noise or a brightness shift) can map to nearly identical targets, providing no gradient pressure to preserve those statistics. A pixel-reconstruction loss, by contrast, penalises every pixel deviation equally, forcing the encoder to allocate capacity to reproducing textures, lighting, and backgrounds, which becomes a liability when those pixel statistics are corrupted at test time.

\paragraph{Stable representations are not necessarily useful representations.}
A striking dissociation emerges in the cosine-similarity panel of Figure~\ref{fig:severity-curves} (right). VideoPrism maintains near-perfect cosine similarity between clean and corrupted features even at the highest severity, yet suffers the lowest accuracy retention among all models. A representation can thus remain geometrically stable while its class-separating structure erodes. V-JEPA exhibits the converse pattern: its cosine similarity under corruption is in fact \emph{lower} than VideoPrism's (Figure~\ref{fig:severity-curves}, right), yet it retains the highest classification accuracy (Figure~\ref{fig:severity-curves}, left). This dissociation rules out generic invariance as the source of V-JEPA's robustness and instead points to the selective preservation of information that matters for discrimination. Theoretically, VideoPrism's contrastive video-text alignment encourages the encoder to map all views of the same concept toward a shared embedding region, producing geometric stability, but this invariance pressure simultaneously compresses the intra-class variation that separates the  visually similar categories. V-JEPA does not force invariance; it preserves whatever details help predict missing content, retaining useful distinctions even if the overall representation shifts. This dissociation reproduces more cleanly still on EGTEA Gaze+, where VideoPrism holds the highest cosine similarity and the lowest accuracy retention of every encoder evaluated (Appendix~\ref{app:egtea}).

\section{Can latent prediction distinguish pretend vs real actions?}
\label{sec:pretend-actions}
We evaluate our foundation models on a pretending action set, comprising 1,992 test videos across 22 fine-grained action classes. Each class describes a simulated or failing version of a physical interaction (e.g., ``Pretending to pick something up''), forming a deliberately adversarial probe: the visual appearance, gesture kinematics, and camera viewpoint are nearly identical to their genuine counterparts, yet the defining signal is the absence of physical contact or the failure to complete the interaction. A model that relies on coarse appearance or dominant motion trajectories will be systematically misled. To characterise the nature of each action's discriminative signal, we partition pretending classes into two groups: \textit{high sensitivity} actions, whose pretend variant is detectable only through fine-grained absence-of-contact cues such as an empty grasp or fluid failing to flow; and \textit{low sensitivity} actions, discriminable from coarser whole-body or object-level motion. We refer to Appendix~\ref{app:pretend-categories} for the full categorisation of the actions.

\begin{figure}[t]
    \centering
    \includegraphics[width=\linewidth]{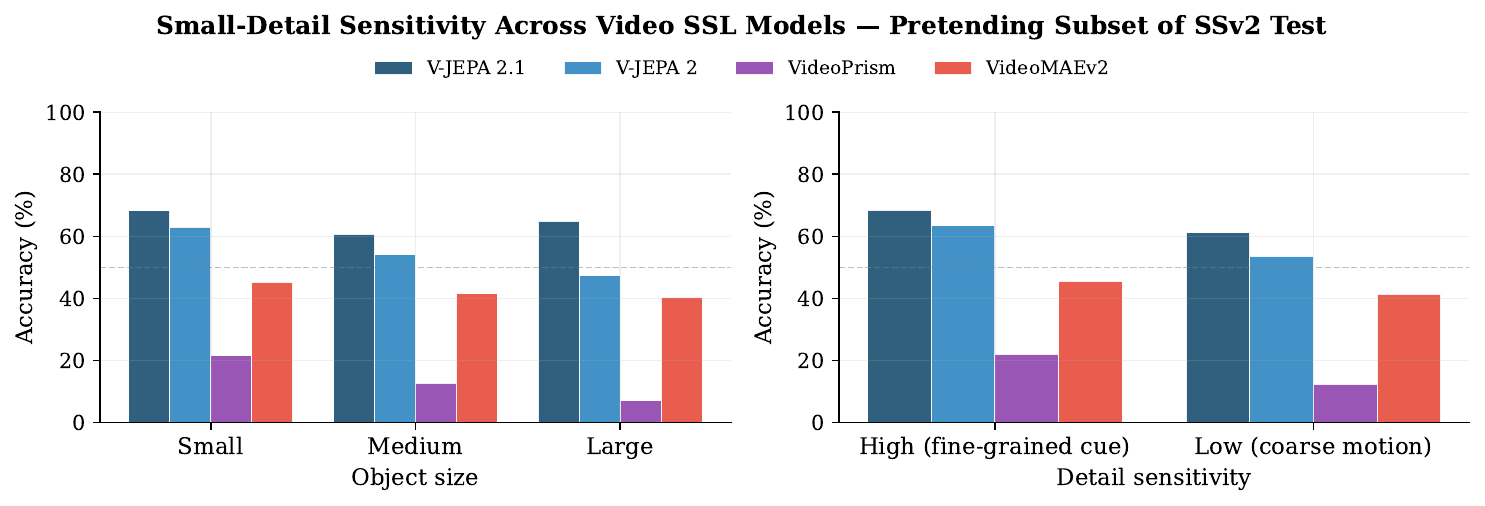}
    \caption{Evaluation on the pretend actions.
    \textbf{(Left)}~Actions are categorised based on small, medium and large sizes of the objects involved in the interaction. \textbf{(Right)}~High sensitivity actions require understanding fine-grained absence-of-contact cues, e.g., empty grasp. Low sensitivity actions are discriminable from coarser object-level motion. Appendix~\ref{app:pretend-categories} provides more details on categories.}
    \label{fig:pretend-sensitivity}
\end{figure}
\paragraph{V-JEPA captures fine-grained contact cues despite never reconstructing pixels.}
A model trained to predict in latent feature space, without ever regenerating individual pixels, might be expected to sacrifice sensitivity to fine-grained visual detail compared to one trained to reconstruct masked spatial content. The results refute this expectation. V-JEPA variants consistently outperform VideoMAEv2 across all object size categories and both detail sensitivity groups (Figure~\ref{fig:pretend-sensitivity}), and per-class comparison confirms the advantage spans virtually every individual action class rather than being driven by a few outliers (see Appendix~\ref{app:pretend-delta}). This pattern suggests that the latent predictive objective distils the spatiotemporal signatures that differentiate real from simulated contact, while pixel reconstruction allocates capacity to surface texture, diluting the contact signal.
\paragraph{The latent-space advantage is stronger   where pixel-level training is expected to help the most.}
The gap between V-JEPA and VideoMAEv2 widens on high sensitivity classes, those requiring detection of subtle absence-of-contact cues such as a fingertip producing no surface reaction or an object undergoing no deformation - Figure~\ref{fig:pretend-sensitivity} (right).
VideoMAEv2 derives minimal benefit from the fine-grained contact signal that distinguishes the two sensitivity groups, whereas V-JEPA shows a clear accuracy lift on these classes. This directly contradicts the intuition that reconstructing spatial detail confers an advantage where spatial detail matters most, and instead supports the hypothesis that joint-embedding predictive objectives capture higher-level physical semantics that pixel-level objectives fail to preserve. The residual errors are shared: the same two classes rank hardest for all four encoders, so what remains is inherent kinematic ambiguity rather than an architecture-specific failure (Appendix~\ref{app:pretend-hardest}). To predict a masked region's latent representation, V-JEPA's encoder must capture abstract state changes, including whether contact occurs and produces a physical effect. Pixel reconstruction spreads capacity across texture and colour details irrelevant to this distinction, diluting the contact signal.
\section{What happens when frames are occluded or missing?}
\label{sec:occlusion}

Any video world model deployed in realistic conditions must tolerate missing
visual information. Objects disappear behind one another, frames drop, and
sensors fail. We probe the occlusion robustness of four video encoders through
three complementary paradigms: \emph{Moving Block}, which sweeps a grey square across the spatial extent and isolates spatial robustness;
\emph{Temporal Dropout}, which freezes a contiguous block of frames and
isolates temporal robustness; and \emph{Spatiotemporal Patch Dropout}, which
zeros random 3D cuboids in the video volume and mirrors the masked prediction
objective used by V-JEPA. Each paradigm is applied at three severity levels
(parameters $\alpha, \beta, \gamma$). Full hyperparameters appear in Appendix~\ref{app:occlusion_hparams}.

\begin{figure}[t]
  \centering
  \includegraphics[width=\linewidth]{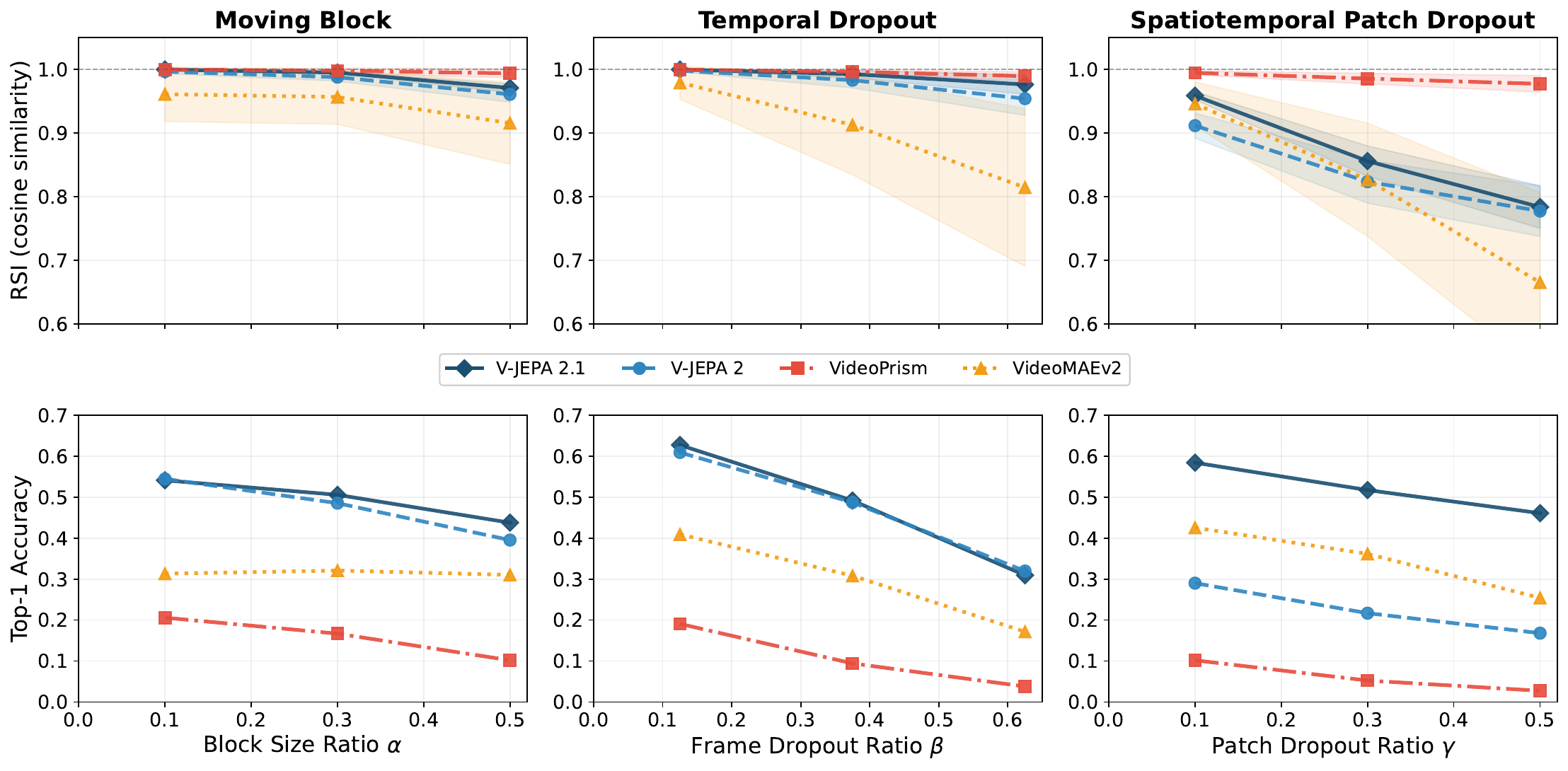}
  \caption{\textbf{Top:} Cosine similarity between clean and occluded embeddings as a function
  of occlusion severity for three paradigms.
  \textbf{Bottom:} Top-1 classification accuracy
  under the same occlusions.
  }
  \label{fig:occlusion_composite}
\end{figure}
\paragraph{Predicting in latent space generalises better to missing input
than predicting in pixel space.}
Both V-JEPA and VideoMAEv2 are pre-trained with patch masking, yet the V-JEPA
encoders produce substantially more stable features than VideoMAEv2 under all
three paradigms, and V-JEPA\,2.1 retains the highest classification accuracy at
every severity of every paradigm
(Figure~\ref{fig:occlusion_composite}). VideoMAEv2 degrades the fastest under
temporal dropout and patch dropout, exhibiting the steepest cosine-similarity slopes of all
encoders. Because these two model families differ primarily in whether the
prediction target is a latent representation or the raw pixels, this contrast
provides direct evidence that the choice of prediction space, not the
masking strategy itself, plays the key role in  occlusion robustness.
\paragraph{V-JEPA\,2.1 is the most robust encoder for downstream decision making under occlusion.}
V-JEPA\,2.1 leads in probe accuracy under all three occlusion paradigms and sustains the most graceful degradation as severity increases. Among the encoders evaluated, it provides the most dependable feature backbone for a world model that must produce correct decisions from incomplete observations.
\paragraph{Stable representations are not necessarily useful representations (occlusion).}
The dissociation we observed under corruption in Section~\ref{sec:corruption} is even more  starkly manifested in Figure~\ref{fig:occlusion_composite}. Under spatiotemporal patch dropout at maximum severity, VideoPrism's embedding shifts by less than two percent in cosine distance while its probe accuracy collapses to near chance. V-JEPA\,2.1 sits at the opposite extreme: its representations shift noticeably more, yet it retains the highest probe accuracy across all three paradigms. We also quantify this gap more concretely in per-model Decoupling Index (DI) in Figure~\ref{fig:decoupling} in the appendix.
\section{Do video models encode the arrow of time?}
\label{sec:temporal}

We probe how much each encoder relies on temporal structure by applying ten perturbation conditions from four families: frame permutation, static video (all frames replaced by a single repeated still), pure noise (Gaussian, uniform, and temporally static Gaussian), and video reversal. Results for frame permutation, static video, and pure noise are reported in Appendix~\ref{app:temporal_extra}; here we focus on the most architecturally discriminating finding.

\paragraph{Only predictive models encode directional semantics.}
Video reversal retains every frame and every spatial feature while reversing
the temporal order, providing a clean test of whether directionality is a
meaningful axis in the latent space. All encoders are used \emph{frozen},
so any directional structure observed must have been internalized during
pre-training rather than acquired by the attentive probe. We quantify this
with the Directional Semantic Coherence Score:
\[
  \mathrm{DSCS} = r_{\mathrm{sem}} \times (1 - \cos_{\mathrm{rev}}),
\]
where $r_{\mathrm{sem}}$ is the fraction of prediction changes that land on a
semantically antonymous class under reversal (e.g., \emph{pushing something
from left to right} $\to$ \emph{pulling something from right to left}) and
$(1 - \cos_{\mathrm{rev}})$ is the representational distance induced by
reversal. A high DSCS requires both that the representation shifts
meaningfully and that the shift is semantically structured.

\begin{figure}[h]
  \centering
  \includegraphics[width=\linewidth]{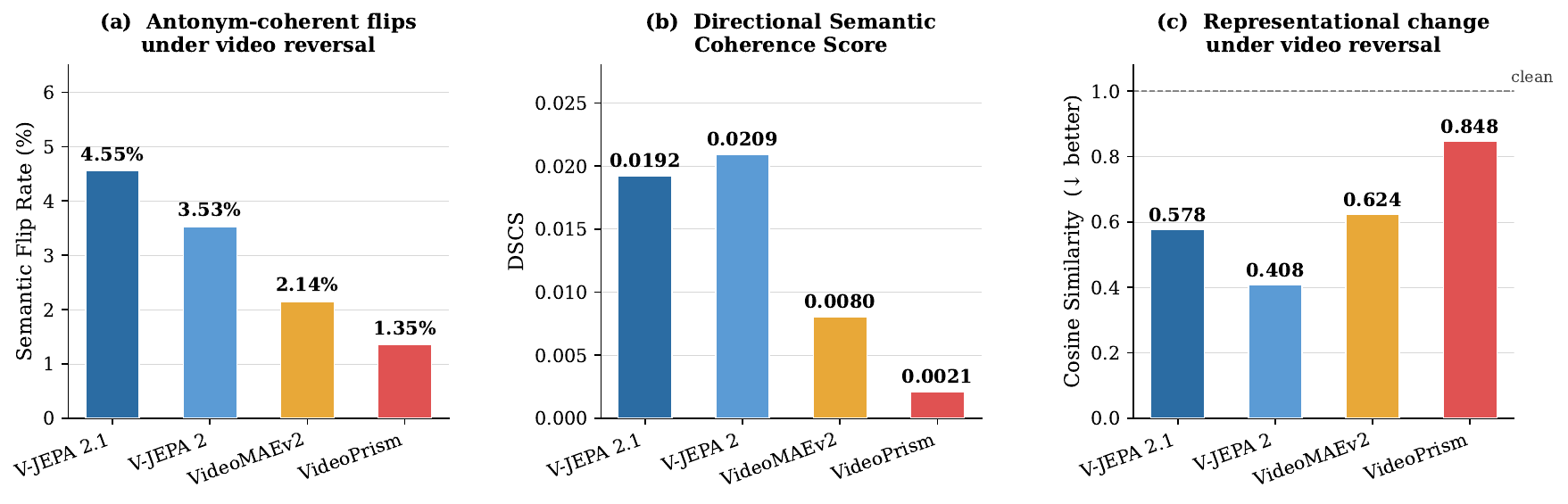}
  \caption{\textbf{Video reversal}
  \textbf{(a)}~Semantic Flip Rate ($r_{\mathrm{sem}}$): the fraction of prediction changes under
  reversal that land on a semantically antonymous class (e.g.,
  \emph{pushing} $\to$ \emph{pulling}).
  \textbf{(b)}~Directional Semantic Coherence Score (DSCS), which combines
  the semantic flip rate with the representational distance induced by
  reversal.
  \textbf{(c)}~Cosine similarity between clean and reversed embeddings.}
  \label{fig:temporal_reversal}
\end{figure}

As shown in Figure~\ref{fig:temporal_reversal}(b), the V-JEPA models achieve DSCS values $2.4\times$ those of VideoMAEv2 and $9\times$ those of VideoPrism. Because the encoder is frozen, this coherent antonym-flipping is a signature of pretraining: temporal direction is internalized as a causal concept, not merely as a source of variation to be averaged out. During pretraining, V-JEPA masks a contiguous block of future patches and predicts their latent representations from earlier context, creating an inherently directional learning signal: the encoder must represent what comes \emph{after} the visible context, not merely what co-occurs with it. The latent space acquires an oriented temporal axis separating ``pushing'' and ``pulling'' by direction rather than collapsing them. Contrastive learning suppresses this axis by treating temporal augmentations as positive pairs; pixel reconstruction dilutes it across the pixel-recovery objective. VideoPrism produces the most prediction changes yet the fewest semantically coherent ones, consistent with its stable but unusable embedding behavior observed across corruption and occlusion (Sections~\ref{sec:encoder-discriminability}--\ref{sec:occlusion}): it detects change but lacks the temporal axis to interpret it. This axis transfers almost numerically to a second dataset: on EGTEA Gaze+ every encoder's reversal cosine lands within about 0.10 of its SSv2 value, on a different domain and label space (Appendix~\ref{app:egtea}).

\section{Do these robustness gains survive downstream world modelling?}
\label{sec:downstream}

Sections~\ref{sec:encoder-discriminability}--\ref{sec:temporal} measured the encoder in
isolation. We now supply the second component of a world model, the predictor, and ask whether
the differences we measured change what an agent can do. Each encoder stays frozen; we train a
lightweight action-conditioned latent predictor that is identical across encoders, and plan with
it in the Meta-World manipulation environment~\citep{yu2020meta}. The predictor is a 4-layer causal
transformer ($\approx$3.2M parameters) that forecasts future latents conditioned on actions, and a
cross-entropy-method planner searches over action sequences with it. Two contact-rich tasks
(\texttt{drawer-open}, \texttt{door-close}) are used for training and evaluation. Perturbations reuse the corruption and occlusion families of
Sections~\ref{sec:corruption} and~\ref{sec:occlusion}, and a held-out task tests whether the
learned dynamics generalise. A scripted expert and a random policy bound every cell.
Appendix~\ref{app:downstream} gives the protocol, the full results and the controls.

\begin{figure}[b]
  \centering
  \includegraphics[width=\linewidth]{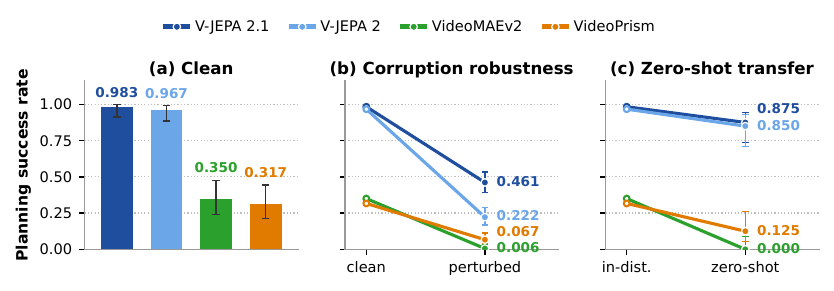}
  \caption{Closed-loop planning success with frozen encoders.
  \textbf{(a)}~Clean inputs. \textbf{(b)}~Mean success across three perturbed conditions.
  \textbf{(c)}~A held-out task, absent from every predictor's training set.
  Hollow markers repeat the clean reference from (a); whiskers are Wilson 95\% intervals.
  Significance tests are reported in Appendix~\ref{app:downstream}.}
  \label{fig:downstream}
\end{figure}

\paragraph{Latent prediction surpasses the other objectives in planning.}
On clean inputs the encoders do not rank, they divide (Figure~\ref{fig:downstream}(a)). Both
V-JEPA models drive the planner to near-complete success, while the pixel-reconstruction and
contrastive encoders leave it failing most episodes. 
\paragraph{Only latent prediction still plans when the input degrades.}
Corruption and occlusion separate the encoders completely, with V-JEPA\,2.1 degrading most
gracefully, V-JEPA\,2 following, and the non-predictive encoders near the floor of the
task: across the three perturbed conditions VideoMAEv2 completes 1 of 180 episodes and
VideoPrism 12, against 83 for V-JEPA\,2.1 (Figure~\ref{fig:downstream}(b)). The advantage is
largest under precisely the conditions a deployed world model must survive. 
\paragraph{Zero-shot: V-JEPA generalises to dynamics it was never trained on; the others do not.}
On a held-out manipulation task the V-JEPA models retain roughly nine tenths of their
in-distribution success, while VideoMAEv2 does not complete a single episode and VideoPrism
completes 5 of 40 (Figure~\ref{fig:downstream}(c)).
No predictor saw this task, so the split is carried by the frozen encoder rather than by the
tasks it was fitted to. Latent prediction yields dynamics that remain usable outside the setting
they were trained on.


\section{Conclusion}
We presented an extensive evaluation of self-supervised video representations as components of world models. Across multiple robustness axes, latent prediction produces the most consistently capable representations. These representations resist pixel corruption, tolerate occlusion, distinguish pretend actions from their real counterparts, and respond coherently to temporal reversal. These are not isolated strengths, but related consequences of how the pretraining objective shapes the representation. This advantage extends to downstream world modeling: with an identical predictor and planner paired with every frozen encoder, only latent prediction continues to plan under degraded input and transfers to a new manipulation task for which the predictor was never trained (Section~\ref{sec:downstream}). More broadly, our results suggest that conventional video-understanding benchmarks and single task scores are insufficient to capture progress in self-supervised video learning, particularly when these models are intended for world modeling, motivating multi-axis evaluation that exposes structural differences between pretraining paradigms.

\section*{Acknowledgements}
\vspace{-4pt}
Ali Alrasheed is supported by a scholarship from Aramco. Naveed Akhtar is a recipient of the Australian Research Council Discovery Early Career Researcher Award (project \# DE230101058), funded by the Australian Government. This research was also supported by The University of Melbourne’s Research Computing Services and the Petascale Campus Initiative.

\bibliographystyle{plainnat}
\bibliography{references}

\section*{NeurIPS Paper Checklist}

\begin{enumerate}

\item {\bf Claims}
    \item[] Question: Do the main claims made in the abstract and introduction accurately reflect the paper's contributions and scope?
    \item[] Answer: \answerYes{}
    \item[] Justification: The abstract and introduction explicitly enumerate five findings (corruption robustness, stability-vs-usability dissociation, pretend-action discrimination, arrow-of-time encoding, and downstream closed-loop planning) that are each supported by dedicated experimental sections (Sections~\ref{sec:encoder-discriminability}--\ref{sec:downstream}) with quantitative results, replicated on a second dataset (Appendix~\ref{app:egtea}).
    \item[] Guidelines:
    \begin{itemize}
        \item The answer \answerNA{} means that the abstract and introduction do not include the claims made in the paper.
        \item The abstract and/or introduction should clearly state the claims made, including the contributions made in the paper and important assumptions and limitations. A \answerNo{} or \answerNA{} answer to this question will not be perceived well by the reviewers.
        \item The claims made should match theoretical and experimental results, and reflect how much the results can be expected to generalize to other settings.
        \item It is fine to include aspirational goals as motivation as long as it is clear that these goals are not attained by the paper.
    \end{itemize}

\item {\bf Limitations}
    \item[] Question: Does the paper discuss the limitations of the work performed by the authors?
    \item[] Answer: \answerYes{}
    \item[] Justification: A dedicated Limitations section (Appendix~\ref{app:limitations}) discusses the factors that are not equalised across public checkpoints (input resolution, clip length, pretraining data composition) and how pretraining scale is controlled for (Appendix~\ref{app:datascale}); limitations of the closed-loop planning study are given in Appendix~\ref{app:downstream_limitations}.
    \item[] Guidelines:
    \begin{itemize}
        \item The answer \answerNA{} means that the paper has no limitation while the answer \answerNo{} means that the paper has limitations, but those are not discussed in the paper.
        \item The authors are encouraged to create a separate ``Limitations'' section in their paper.
        \item The paper should point out any strong assumptions and how robust the results are to violations of these assumptions (e.g., independence assumptions, noiseless settings, model well-specification, asymptotic approximations only holding locally). The authors should reflect on how these assumptions might be violated in practice and what the implications would be.
        \item The authors should reflect on the scope of the claims made, e.g., if the approach was only tested on a few datasets or with a few runs. In general, empirical results often depend on implicit assumptions, which should be articulated.
        \item The authors should reflect on the factors that influence the performance of the approach. For example, a facial recognition algorithm may perform poorly when image resolution is low or images are taken in low lighting. Or a speech-to-text system might not be used reliably to provide closed captions for online lectures because it fails to handle technical jargon.
        \item The authors should discuss the computational efficiency of the proposed algorithms and how they scale with dataset size.
        \item If applicable, the authors should discuss possible limitations of their approach to address problems of privacy and fairness.
        \item While the authors might fear that complete honesty about limitations might be used by reviewers as grounds for rejection, a worse outcome might be that reviewers discover limitations that aren't acknowledged in the paper. The authors should use their best judgment and recognize that individual actions in favor of transparency play an important role in developing norms that preserve the integrity of the community. Reviewers will be specifically instructed to not penalize honesty concerning limitations.
    \end{itemize}

\item {\bf Theory assumptions and proofs}
    \item[] Question: For each theoretical result, does the paper provide the full set of assumptions and a complete (and correct) proof?
    \item[] Answer: \answerNA{}
    \item[] Justification: This is a purely empirical paper; it contains no formal theorems, lemmas, or mathematical proofs.
    \item[] Guidelines:
    \begin{itemize}
        \item The answer \answerNA{} means that the paper does not include theoretical results.
        \item All the theorems, formulas, and proofs in the paper should be numbered and cross-referenced.
        \item All assumptions should be clearly stated or referenced in the statement of any theorems.
        \item The proofs can either appear in the main paper or the supplemental material, but if they appear in the supplemental material, the authors are encouraged to provide a short proof sketch to provide intuition.
        \item Inversely, any informal proof provided in the core of the paper should be complemented by formal proofs provided in appendix or supplemental material.
        \item Theorems and Lemmas that the proof relies upon should be properly referenced.
    \end{itemize}

    \item {\bf Experimental result reproducibility}
    \item[] Question: Does the paper fully disclose all the information needed to reproduce the main experimental results of the paper to the extent that it affects the main claims and/or conclusions of the paper (regardless of whether the code and data are provided or not)?
    \item[] Answer: \answerYes{}
    \item[] Justification: Full hyperparameters (probe architecture, occlusion parameters, corruption severities, random seed) are reported in the Appendix tables; the datasets (Something-Something\,v2, EGTEA Gaze+) and the Meta-World simulator are publicly available; the predictor and planner configuration is given in Appendix~\ref{app:downstream_protocol}; the compute budget ($\sim$2{,}000 A100 GPU-hours) is stated; and code will be released.
    \item[] Guidelines:
    \begin{itemize}
        \item The answer \answerNA{} means that the paper does not include experiments.
        \item If the paper includes experiments, a \answerNo{} answer to this question will not be perceived well by the reviewers: Making the paper reproducible is important, regardless of whether the code and data are provided or not.
        \item If the contribution is a dataset and\slash or model, the authors should describe the steps taken to make their results reproducible or verifiable.
        \item Depending on the contribution, reproducibility can be accomplished in various ways. For example, if the contribution is a novel architecture, describing the architecture fully might suffice, or if the contribution is a specific model and empirical evaluation, it may be necessary to either make it possible for others to replicate the model with the same dataset, or provide access to the model. In general. releasing code and data is often one good way to accomplish this, but reproducibility can also be provided via detailed instructions for how to replicate the results, access to a hosted model (e.g., in the case of a large language model), releasing of a model checkpoint, or other means that are appropriate to the research performed.
        \item While NeurIPS does not require releasing code, the conference does require all submissions to provide some reasonable avenue for reproducibility, which may depend on the nature of the contribution. For example
        \begin{enumerate}
            \item If the contribution is primarily a new algorithm, the paper should make it clear how to reproduce that algorithm.
            \item If the contribution is primarily a new model architecture, the paper should describe the architecture clearly and fully.
            \item If the contribution is a new model (e.g., a large language model), then there should either be a way to access this model for reproducing the results or a way to reproduce the model (e.g., with an open-source dataset or instructions for how to construct the dataset).
            \item We recognize that reproducibility may be tricky in some cases, in which case authors are welcome to describe the particular way they provide for reproducibility. In the case of closed-source models, it may be that access to the model is limited in some way (e.g., to registered users), but it should be possible for other researchers to have some path to reproducing or verifying the results.
        \end{enumerate}
    \end{itemize}

\item {\bf Open access to data and code}
    \item[] Question: Does the paper provide open access to the data and code, with sufficient instructions to faithfully reproduce the main experimental results, as described in supplemental material?
    \item[] Answer: \answerYes{}
    \item[] Justification: Evaluation code and scripts will be released as open-source supplemental material; Something-Something\,v2 and EGTEA Gaze+ are publicly available benchmarks, Meta-World is an open-source simulator, and all evaluated model checkpoints are publicly accessible.
    \item[] Guidelines:
    \begin{itemize}
        \item The answer \answerNA{} means that paper does not include experiments requiring code.
        \item Please see the NeurIPS code and data submission guidelines (\url{https://neurips.cc/public/guides/CodeSubmissionPolicy}) for more details.
        \item While we encourage the release of code and data, we understand that this might not be possible, so \answerNo{} is an acceptable answer. Papers cannot be rejected simply for not including code, unless this is central to the contribution (e.g., for a new open-source benchmark).
        \item The instructions should contain the exact command and environment needed to run to reproduce the results. See the NeurIPS code and data submission guidelines (\url{https://neurips.cc/public/guides/CodeSubmissionPolicy}) for more details.
        \item The authors should provide instructions on data access and preparation, including how to access the raw data, preprocessed data, intermediate data, and generated data, etc.
        \item The authors should provide scripts to reproduce all experimental results for the new proposed method and baselines. If only a subset of experiments are reproducible, they should state which ones are omitted from the script and why.
        \item At submission time, to preserve anonymity, the authors should release anonymized versions (if applicable).
        \item Providing as much information as possible in supplemental material (appended to the paper) is recommended, but including URLs to data and code is permitted.
    \end{itemize}

\item {\bf Experimental setting/details}
    \item[] Question: Does the paper specify all the training and test details (e.g., data splits, hyperparameters, how they were chosen, type of optimizer) necessary to understand the results?
    \item[] Answer: \answerYes{}
    \item[] Justification: Dataset splits, attentive-probe architecture (depth, heads, MLP ratio), kNN probe settings, corruption severities, occlusion hyperparameters, and random seed are all reported in the main paper and Appendices~\ref{app:probe_training}, \ref{app:occlusion_hparams_table}, \ref{app:temporal_hparams} and~\ref{app:downstream_protocol}.
    \item[] Guidelines:
    \begin{itemize}
        \item The answer \answerNA{} means that the paper does not include experiments.
        \item The experimental setting should be presented in the core of the paper to a level of detail that is necessary to appreciate the results and make sense of them.
        \item The full details can be provided either with the code, in appendix, or as supplemental material.
    \end{itemize}

\item {\bf Experiment statistical significance}
    \item[] Question: Does the paper report error bars suitably and correctly defined or other appropriate information about the statistical significance of the experiments?
    \item[] Answer: \answerYes{}
    \item[] Justification: Wilcoxon signed-rank tests with reported $p$-values ($p<0.01$) are provided for occlusion comparisons in Appendix~\ref{app:occlusion_extra}; standard deviation error bars are shown on the Fisher discriminant ratio plots in Figure~\ref{fig:discrim}; degradation slope regressions report $R^2>0.82$; closed-loop planning success rates carry Wilson 95\% intervals (Figure~\ref{fig:downstream}) and Fisher exact tests (Appendix~\ref{app:downstream}).
    \item[] Guidelines:
    \begin{itemize}
        \item The answer \answerNA{} means that the paper does not include experiments.
        \item The authors should answer \answerYes{} if the results are accompanied by error bars, confidence intervals, or statistical significance tests, at least for the experiments that support the main claims of the paper.
        \item The factors of variability that the error bars are capturing should be clearly stated (for example, train/test split, initialization, random drawing of some parameter, or overall run with given experimental conditions).
        \item The method for calculating the error bars should be explained (closed form formula, call to a library function, bootstrap, etc.)
        \item The assumptions made should be given (e.g., Normally distributed errors).
        \item It should be clear whether the error bar is the standard deviation or the standard error of the mean.
        \item It is OK to report 1-sigma error bars, but one should state it. The authors should preferably report a 2-sigma error bar than state that they have a 96\% CI, if the hypothesis of Normality of errors is not verified.
        \item For asymmetric distributions, the authors should be careful not to show in tables or figures symmetric error bars that would yield results that are out of range (e.g., negative error rates).
        \item If error bars are reported in tables or plots, the authors should explain in the text how they were calculated and reference the corresponding figures or tables in the text.
    \end{itemize}

\item {\bf Experiments compute resources}
    \item[] Question: For each experiment, does the paper provide sufficient information on the computer resources (type of compute workers, memory, time of execution) needed to reproduce the experiments?
    \item[] Answer: \answerYes{}
    \item[] Justification: The paper reports the total compute budget ($\sim$2{,}000 A100 GPU-hours) in Section~1, with a breakdown in Appendix~\ref{app:probe_training}: $\sim$450 A100 GPU-hours for SSv2 attentive probes, $\sim$850 for the SSv2 perturbation suites, $\sim$400 for EGTEA Gaze+, and $\sim$300 for predictor training and closed-loop planning.
    \item[] Guidelines:
    \begin{itemize}
        \item The answer \answerNA{} means that the paper does not include experiments.
        \item The paper should indicate the type of compute workers CPU or GPU, internal cluster, or cloud provider, including relevant memory and storage.
        \item The paper should provide the amount of compute required for each of the individual experimental runs as well as estimate the total compute.
        \item The paper should disclose whether the full research project required more compute than the experiments reported in the paper (e.g., preliminary or failed experiments that didn't make it into the paper).
    \end{itemize}

\item {\bf Code of ethics}
    \item[] Question: Does the research conducted in the paper conform, in every respect, with the NeurIPS Code of Ethics \url{https://neurips.cc/public/EthicsGuidelines}?
    \item[] Answer: \answerYes{}
    \item[] Justification: The study uses only publicly available datasets and pre-trained model checkpoints, involves no human subjects, collects no personal data, and raises no known ethical concerns.
    \item[] Guidelines:
    \begin{itemize}
        \item The answer \answerNA{} means that the authors have not reviewed the NeurIPS Code of Ethics.
        \item If the authors answer \answerNo, they should explain the special circumstances that require a deviation from the Code of Ethics.
        \item The authors should make sure to preserve anonymity (e.g., if there is a special consideration due to laws or regulations in their jurisdiction).
    \end{itemize}

\item {\bf Broader impacts}
    \item[] Question: Does the paper discuss both potential positive societal impacts and negative societal impacts of the work performed?
    \item[] Answer: \answerNA{}
    \item[] Justification: This paper contributes evaluation methodology for video representation learning. It does not introduce a deployable system or a new model, and there is no direct pathway from the findings to harmful applications.
    \item[] Guidelines:
    \begin{itemize}
        \item The answer \answerNA{} means that there is no societal impact of the work performed.
        \item If the authors answer \answerNA{} or \answerNo, they should explain why their work has no societal impact or why the paper does not address societal impact.
        \item Examples of negative societal impacts include potential malicious or unintended uses (e.g., disinformation, generating fake profiles, surveillance), fairness considerations (e.g., deployment of technologies that could make decisions that unfairly impact specific groups), privacy considerations, and security considerations.
        \item The conference expects that many papers will be foundational research and not tied to particular applications, let alone deployments. However, if there is a direct path to any negative applications, the authors should point it out. For example, it is legitimate to point out that an improvement in the quality of generative models could be used to generate Deepfakes for disinformation. On the other hand, it is not needed to point out that a generic algorithm for optimizing neural networks could enable people to train models that generate Deepfakes faster.
        \item The authors should consider possible harms that could arise when the technology is being used as intended and functioning correctly, harms that could arise when the technology is being used as intended but gives incorrect results, and harms following from (intentional or unintentional) misuse of the technology.
        \item If there are negative societal impacts, the authors could also discuss possible mitigation strategies (e.g., gated release of models, providing defenses in addition to attacks, mechanisms for monitoring misuse, mechanisms to monitor how a system learns from feedback over time, improving the efficiency and accessibility of ML).
    \end{itemize}

\item {\bf Safeguards}
    \item[] Question: Does the paper describe safeguards that have been put in place for responsible release of data or models that have a high risk for misuse (e.g., pre-trained language models, image generators, or scraped datasets)?
    \item[] Answer: \answerNA{}
    \item[] Justification: No new generative models, pre-trained models, or scraped datasets are released; the paper releases only evaluation code and uses existing public benchmarks.
    \item[] Guidelines:
    \begin{itemize}
        \item The answer \answerNA{} means that the paper poses no such risks.
        \item Released models that have a high risk for misuse or dual-use should be released with necessary safeguards to allow for controlled use of the model, for example by requiring that users adhere to usage guidelines or restrictions to access the model or implementing safety filters.
        \item Datasets that have been scraped from the Internet could pose safety risks. The authors should describe how they avoided releasing unsafe images.
        \item We recognize that providing effective safeguards is challenging, and many papers do not require this, but we encourage authors to take this into account and make a best faith effort.
    \end{itemize}

\item {\bf Licenses for existing assets}
    \item[] Question: Are the creators or original owners of assets (e.g., code, data, models), used in the paper, properly credited and are the license and terms of use explicitly mentioned and properly respected?
    \item[] Answer: \answerYes{}
    \item[] Justification: Something-Something\,v2, EGTEA Gaze+, Meta-World, V-JEPA\,2.1, V-JEPA\,2, V-JEPA\,1, VideoPrism, VideoMAEv2, VideoMAE, and TimeSformer are all cited with their original publications; all are publicly released under open-access or open-source terms consistent with academic use.
    \item[] Guidelines:
    \begin{itemize}
        \item The answer \answerNA{} means that the paper does not use existing assets.
        \item The authors should cite the original paper that produced the code package or dataset.
        \item The authors should state which version of the asset is used and, if possible, include a URL.
        \item The name of the license (e.g., CC-BY 4.0) should be included for each asset.
        \item For scraped data from a particular source (e.g., website), the copyright and terms of service of that source should be provided.
        \item If assets are released, the license, copyright information, and terms of use in the package should be provided. For popular datasets, \url{paperswithcode.com/datasets} has curated licenses for some datasets. Their licensing guide can help determine the license of a dataset.
        \item For existing datasets that are re-packaged, both the original license and the license of the derived asset (if it has changed) should be provided.
        \item If this information is not available online, the authors are encouraged to reach out to the asset's creators.
    \end{itemize}

\item {\bf New assets}
    \item[] Question: Are new assets introduced in the paper well documented and is the documentation provided alongside the assets?
    \item[] Answer: \answerYes{}
    \item[] Justification: New evaluation protocols, semantically stratified class splits, and metrics (DSCS, Decoupling Index, Spatial Grounding Index, Temporal Consistency Bonus) are fully described in the main paper and Appendix; code implementing them will be released with documentation.
    \item[] Guidelines:
    \begin{itemize}
        \item The answer \answerNA{} means that the paper does not release new assets.
        \item Researchers should communicate the details of the dataset\slash code\slash model as part of their submissions via structured templates. This includes details about training, license, limitations, etc.
        \item The paper should discuss whether and how consent was obtained from people whose asset is used.
        \item At submission time, remember to anonymize your assets (if applicable). You can either create an anonymized URL or include an anonymized zip file.
    \end{itemize}

\item {\bf Crowdsourcing and research with human subjects}
    \item[] Question: For crowdsourcing experiments and research with human subjects, does the paper include the full text of instructions given to participants and screenshots, if applicable, as well as details about compensation (if any)?
    \item[] Answer: \answerNA{}
    \item[] Justification: The paper does not involve crowdsourcing or research with human subjects; all experiments use pre-existing annotations from the Something-Something\,v2 and EGTEA Gaze+ benchmarks, or a simulator.
    \item[] Guidelines:
    \begin{itemize}
        \item The answer \answerNA{} means that the paper does not involve crowdsourcing nor research with human subjects.
        \item Including this information in the supplemental material is fine, but if the main contribution of the paper involves human subjects, then as much detail as possible should be included in the main paper.
        \item According to the NeurIPS Code of Ethics, workers involved in data collection, curation, or other labor should be paid at least the minimum wage in the country of the data collector.
    \end{itemize}

\item {\bf Institutional review board (IRB) approvals or equivalent for research with human subjects}
    \item[] Question: Does the paper describe potential risks incurred by study participants, whether such risks were disclosed to the subjects, and whether Institutional Review Board (IRB) approvals (or an equivalent approval/review based on the requirements of your country or institution) were obtained?
    \item[] Answer: \answerNA{}
    \item[] Justification: No human subjects are involved in this research; IRB approval is not applicable.
    \item[] Guidelines:
    \begin{itemize}
        \item The answer \answerNA{} means that the paper does not involve crowdsourcing nor research with human subjects.
        \item Depending on the country in which research is conducted, IRB approval (or equivalent) may be required for any human subjects research. If you obtained IRB approval, you should clearly state this in the paper.
        \item We recognize that the procedures for this may vary significantly between institutions and locations, and we expect authors to adhere to the NeurIPS Code of Ethics and the guidelines for their institution.
        \item For initial submissions, do not include any information that would break anonymity (if applicable), such as the institution conducting the review.
    \end{itemize}

\item {\bf Declaration of LLM usage}
    \item[] Question: Does the paper describe the usage of LLMs if it is an important, original, or non-standard component of the core methods in this research? Note that if the LLM is used only for writing, editing, or formatting purposes and does \emph{not} impact the core methodology, scientific rigor, or originality of the research, declaration is not required.
    \item[] Answer: \answerNA{}
    \item[] Justification: LLMs were used solely for writing and editing assistance and do not form any part of the core methodology; per NeurIPS policy, such usage does not require declaration.
    \item[] Guidelines:
    \begin{itemize}
        \item The answer \answerNA{} means that the core method development in this research does not involve LLMs as any important, original, or non-standard components.
        \item Please refer to our LLM policy in the NeurIPS handbook for what should or should not be described.
    \end{itemize}

\end{enumerate}

\clearpage

\appendix

This appendix provides extended experimental details and additional analyses for each section of the main paper. Appendix~\ref{app:limitations} states the limitations of the study. We then report the two experiments that test the alternative explanations for our results: Appendix~\ref{app:egtea} repeats the study on a second dataset in a different domain, and Appendix~\ref{app:datascale} asks whether pretraining scale rather than the objective accounts for the ordering. Appendix~\ref{app:training_regime} then reports the comparison between frozen latent-prediction features and task-adapted baselines. Appendix~\ref{app:model_comparison} gives global methodological context (encoder selection criteria, the attentive-probe protocol, and details of the fine-tuned and supervised baselines). The next sections mirror the order of the five main-paper evaluation axes: Feature Discriminability (Appendix~\ref{app:intrinsic}), Corruption Robustness (Appendix~\ref{app:corruption}), Fine-Grained Action Discrimination (Appendix~\ref{app:pretend}), Occlusion Robustness (Appendix~\ref{app:occlusion_hparams}), and Temporal Robustness (Appendix~\ref{app:temporal_extra}). Appendix~\ref{app:downstream} documents the closed-loop world-modelling experiment of Section~\ref{sec:downstream}, including the protocol, validation checks, full results, and the controls that establish what the planner relies on.

\section{Limitations}
\label{app:limitations}

Our evaluation provides a broad and controlled assessment across two datasets from different domains, six perturbation families, and a closed-loop planning study. To isolate the role of the pretraining objective as closely as possible, we hold backbone capacity, probe design, evaluation clips, and perturbation code identical across models, while directly measuring pretraining scale (Appendix~\ref{app:datascale}). Other factors, such as input resolution and clip length, are inherited from each public release and therefore differ across checkpoints. Fully isolating the pretraining objective from these release-specific choices would require retraining every model under a shared compute, data, and input configuration, which is beyond our scope. We therefore interpret our findings as evidence consistent with the pretraining objective contributing to the observed differences, rather than as an absolute causal effect measured in isolation.

\section{A second dataset: EGTEA Gaze+}
\label{app:egtea}

The main paper reports Something-Something\,v2. To test whether its conclusions are
properties of the pretraining objectives rather than of one benchmark, we repeated every
axis that ports onto EGTEA Gaze+~\citep{li2018eye}, an egocentric kitchen-activity dataset.
Domain, camera viewpoint, label space and recording setup all differ from SSv2. The encoder
ordering does not change on any axis, and the dissociation between representational
stability and usability reproduces more cleanly than on SSv2.

\subsection{Protocol}
\label{app:egtea_protocol}

We use split 1: 8{,}299 training clips and a class-balanced evaluation set of 1{,}653 clips
spanning 106 verb-object action classes, the closest available match to SSv2's 174. The
frozen encoders, the attentive probe (depth 2, 8 heads, MLP ratio 2.0, $\sim$15.9M
parameters, identical for every encoder as on SSv2) and the perturbation code are exactly
those used on SSv2; only the dataset changes. Each encoder runs at its released input resolution and clip length. The suite
comprises 29 conditions: six corruptions and three occlusion paradigms at three severities
each, plus time reversal. EGTEA is severely class-imbalanced, so we report mean-per-class
accuracy alongside top-1.

Three parts of the study do not port, and we state them as scope boundaries rather than
results. EGTEA has no pretend-versus-real subset, so Section~\ref{sec:pretend-actions} has
no direct analogue; we substitute a stricter appearance-controlled test
(Appendix~\ref{app:egtea_direction}). From the temporal battery we run reversal only, the
condition Section~\ref{sec:temporal} relies on. No public EGTEA fine-tuned VideoMAE or
TimeSformer checkpoint exists, so the training-regime comparison of
Appendix~\ref{app:training_regime} remains SSv2-only.

\subsection{Clean accuracy}

\begin{table}[H]
\centering
\small
\caption{Clean accuracy on the full 2{,}022-clip EGTEA test split. The ordering is identical to SSv2. Note that Tables~\ref{tab:egtea_corruption} and~\ref{tab:egtea_occlusion} report clean accuracy on the 1{,}653-clip class-balanced evaluation subset instead, which runs 7--8 points lower for every encoder; the two are not interchangeable and each perturbation retention uses its own matched denominator.}
\label{tab:egtea_clean}
\begin{tabular}{lccc}
\toprule
\textbf{Model} & \textbf{Top-1 (\%)} & \textbf{Mean-per-class (\%)} & \textbf{SSv2 top-1 (\%)} \\
\midrule
V-JEPA\,2.1 & \textbf{81.45} & \textbf{76.12} & 71.8 \\
V-JEPA\,2   & 77.60 & 70.75 & 70.1 \\
V-JEPA\,1   & 68.60 & 59.43 & 70.3 \\
VideoMAEv2  & 59.25 & 48.17 & 51.8 \\
VideoPrism  & 41.54 & 27.85 & 29.9 \\
\bottomrule
\end{tabular}
\end{table}

Both V-JEPA\,2 models were still improving when they reached the epoch cap, so their margin
over VideoMAEv2 is a lower bound; VideoMAEv2 early-stopped having converged. V-JEPA\,1 is
the only encoder that converged outright, which matters for
Appendix~\ref{app:datascale}: the model at the bottom of the generational progression is the
one trained furthest, so the measured progression understates rather than manufactures the
gap.

\subsection{Corruption robustness}

\begin{table}[H]
\centering
\small
\caption{Corruption robustness on EGTEA, averaged over three severities. Retention is
perturbed over clean top-1; clean is measured on the 1{,}653-clip evaluation subset. Cosine
similarity is pooled, matching the convention of Section~\ref{sec:corruption}.}
\label{tab:egtea_corruption}
\begin{tabular}{lcccc}
\toprule
\textbf{Model} & \textbf{Clean (\%)} & \textbf{Retention (all 6)} & \textbf{Retention (excl.\ elastic)} & \textbf{Cosine} \\
\midrule
V-JEPA\,2.1 & \textbf{79.19} & 0.872 & \textbf{0.972} & 0.944 \\
V-JEPA\,2   & 74.65 & 0.857 & 0.939 & 0.934 \\
V-JEPA\,1   & 64.25 & 0.874 & 0.931 & 0.941 \\
VideoMAEv2  & 53.96 & 0.810 & 0.902 & 0.822 \\
VideoPrism  & 34.06 & 0.663 & 0.704 & \textbf{0.994} \\
\bottomrule
\end{tabular}
\end{table}

V-JEPA\,2.1 holds the highest accuracy at every severity (62.9\% at severity 5, against
56.6, 48.9, 37.0 and 14.9 for the others), reproducing
Figure~\ref{fig:severity-curves}. We report retention with and without elastic transform
because that corruption resamples the spatial grid itself and is the universal outlier
already documented in Appendix~\ref{app:corruption}: it drives every encoder towards the
floor, and because retention is a ratio, the strongest encoders have furthest to fall.
Excluding it, the ordering is strictly monotone across all five encoders and matches the
clean-accuracy ordering exactly.

\subsection{Occlusion robustness}

\begin{table}[H]
\centering
\small
\caption{Occlusion retention on EGTEA, averaged over three severities per paradigm.}
\label{tab:egtea_occlusion}
\begin{tabular}{lccccc}
\toprule
\textbf{Model} & \textbf{Moving block} & \textbf{Temporal dropout} & \textbf{Patch dropout} & \textbf{Mean} & \textbf{Cosine} \\
\midrule
V-JEPA\,2.1 & \textbf{0.969} & \textbf{0.886} & \textbf{0.859} & \textbf{0.905} & 0.932 \\
V-JEPA\,2   & 0.936 & 0.879 & 0.712 & 0.843 & 0.910 \\
V-JEPA\,1   & 0.905 & 0.849 & 0.567 & 0.774 & 0.880 \\
VideoMAEv2  & 0.922 & 0.817 & 0.540 & 0.759 & 0.817 \\
VideoPrism  & 0.807 & 0.840 & 0.300 & 0.649 & \textbf{0.991} \\
\bottomrule
\end{tabular}
\end{table}

V-JEPA\,2.1 leads all three paradigms. Spatiotemporal patch dropout is again the decisive
panel, since it mirrors the masked-prediction objective itself: at $\gamma=0.5$ V-JEPA\,2.1
retains 61.1\% top-1 while VideoMAEv2 collapses to 8.8\%. Both are pretrained with patch
masking and differ mainly in whether the prediction target is a latent representation or raw
pixels, which is the comparison Section~\ref{sec:occlusion} rests on.

\subsection{Stability is not usability}

The dissociation of Sections~\ref{sec:corruption} and~\ref{sec:occlusion} is the paper's
most counter-intuitive claim, and it is the result that transfers most exactly. On EGTEA
VideoPrism holds the highest pooled cosine similarity of any encoder under corruption
(0.994) together with the lowest accuracy retention (0.663), while V-JEPA shows the converse
pattern. Under the strongest patch dropout the numbers sit almost on top of their SSv2
counterparts.

\begin{table}[H]
\centering
\small
\caption{Spatiotemporal patch dropout at maximum severity, on both datasets.}
\label{tab:egtea_dissociation}
\begin{tabular}{lcc}
\toprule
 & \textbf{SSv2 (174 classes)} & \textbf{EGTEA (106 classes)} \\
\midrule
VideoPrism, cosine similarity & $>0.98$ & 0.976 \\
VideoPrism, top-1            & 2.7\%   & 4.2\% \\
V-JEPA\,2.1, top-1           & 46.1\%  & 61.1\% \\
\bottomrule
\end{tabular}
\end{table}

\subsection{Arrow of time}

\begin{table}[H]
\centering
\small
\caption{Reversal on EGTEA. Cosine similarity between clean and time-reversed embeddings is
computed per token, the convention of Section~\ref{sec:temporal}; lower means more sensitive
to the direction of time. The semantic flip rate and DSCS use EGTEA's own verb-antonym map,
so their absolute values are not comparable across datasets and only the ordering is.}
\label{tab:egtea_reversal}
\begin{tabular}{lcccc}
\toprule
\textbf{Model} & \textbf{$\cos_{\mathrm{rev}}$ (EGTEA)} & \textbf{$\cos_{\mathrm{rev}}$ (SSv2)} & \textbf{$r_{\mathrm{sem}}$} & \textbf{DSCS} \\
\midrule
V-JEPA\,2   & \textbf{0.366} & 0.408 & 0.771 & 0.489 \\
V-JEPA\,1   & 0.367 & \textbf{0.391} & 0.798 & \textbf{0.505} \\
V-JEPA\,2.1 & 0.482 & 0.578 & \textbf{0.811} & 0.420 \\
VideoMAEv2  & 0.565 & 0.624 & 0.609 & 0.265 \\
VideoPrism  & 0.851 & 0.848 & 0.425 & 0.063 \\
\bottomrule
\end{tabular}
\end{table}

This is the most quantitatively precise replication in the study: every encoder lands within
about 0.10 of its SSv2 value on a different domain and a different label space, and the
ordering is preserved exactly. The three latent-prediction models are far more sensitive to
reversal than the pixel-reconstruction and contrastive baselines, and V-JEPA\,1 and
V-JEPA\,2 are indistinguishable on this axis on both datasets. The DSCS ordering reproduces
Figure~\ref{fig:temporal_reversal}, including the V-JEPA\,2 over V-JEPA\,2.1 inversion the
main paper already reports; Appendix~\ref{app:datascale} takes that trend up.

\subsection{Discriminability, and an appearance-controlled direction tier}
\label{app:egtea_direction}

\begin{table}[H]
\centering
\small
\caption{Frozen-feature discriminability on EGTEA. \textbf{Left:} 21 classes (7 different-verb,
14 same-verb), matching the construction of Section~\ref{sec:encoder-discriminability}.
\textbf{Right:} the appearance-controlled direction tier, 17 same-noun antonym-verb pairs
over 34 classes; chance is 50\%.}
\label{tab:egtea_discrim}
\begin{tabular}{lcccc}
\toprule
 & \multicolumn{2}{c}{\textbf{21-class tier}} & \multicolumn{2}{c}{\textbf{Direction tier}} \\
\cmidrule(lr){2-3}\cmidrule(lr){4-5}
\textbf{Model} & \textbf{Linear probe} & \textbf{kNN} & \textbf{Linear probe} & \textbf{kNN} \\
\midrule
V-JEPA\,2.1 & \textbf{83.0} & \textbf{73.7} & 63.12 & 58.57 \\
V-JEPA\,2   & 80.3 & 66.6 & 66.03 & 59.99 \\
V-JEPA\,1   & 82.3 & 71.1 & \textbf{67.76} & \textbf{61.92} \\
VideoMAEv2  & 79.2 & 70.4 & 56.84 & 47.71 \\
VideoPrism  & 66.8 & 50.4 & 60.68 & 57.24 \\
\bottomrule
\end{tabular}
\end{table}

EGTEA has no pretend-versus-real subset, so the fine-grained axis needed a substitute. SSv2's
hard tier is direction-contrastive, pairing \emph{take something from left to right} against
\emph{right to left}, so the faithful analogue holds the object fixed and varies the
direction of the action: we selected 17 pairs sharing a noun but taking antonym verbs, such
as open versus close fridge and take versus put plate. Object, scene and frame content are
held constant by construction, so the discriminating cue is temporal. All three V-JEPA
generations lead, and the pixel-reconstruction model sits closest to chance. The tier is
direction-\emph{dominant} rather than appearance-free, since the end state of a pair still
differs; what it removes outright is the object-identity shortcut.

That shortcut is worth reporting, because our first attempt at the tier ran into it. We
initially built the hard tier as same-verb, different-noun pairs (take plate versus take
cup), and on that tier VideoMAEv2 separated the hard classes \emph{better} than the easy ones,
with a hard-to-easy Fisher ratio of 1.44 against 0.84--0.87 for V-JEPA\,2.1 and V-JEPA\,2.
The discriminating cue there is object identity, precisely what a pixel-reconstruction
objective is strongest at, so we had built the wrong kind of difficulty. The inversion is
itself consistent with the paper's account of what each objective encodes, and it is why the
direction tier above is the analogue we report.

\subsection{Summary}

\begin{table}[H]
\centering
\small
\caption{Every axis that ports from SSv2 to EGTEA Gaze+.}
\label{tab:egtea_summary}
\begin{tabular}{lll}
\toprule
\textbf{Axis} & \textbf{Reproduces?} & \textbf{Note} \\
\midrule
Clean accuracy ordering & Yes & Identical ordering \\
Corruption robustness   & Yes & V-JEPA\,2.1 leads at every severity \\
Occlusion robustness    & Yes & V-JEPA\,2.1 leads all three paradigms \\
Stability vs.\ usability & Yes & More cleanly than on SSv2 \\
Arrow of time           & Yes & Within $\sim$0.10 of each SSv2 value \\
Discriminability        & Yes & V-JEPA leads; VideoPrism last \\
Fine-grained tier       & Yes & Under a stricter control (Appendix~\ref{app:egtea_direction}) \\
Training-regime comparison & Not run & No public EGTEA fine-tuned checkpoints \\
\bottomrule
\end{tabular}
\end{table}

\section{Is the advantage a pretraining-data effect?}
\label{app:datascale}

The most likely alternative explanation for our results is that the latent-prediction models
simply saw more or better pretraining data. We test this directly. V-JEPA\,1 shares
architecture, capacity, embedding width, pretraining objective and probe design with
V-JEPA\,2, and was trained on VideoMix2M ($\sim$2M clips, 90K iterations) against
V-JEPA\,2's VideoMix22M (22M clips, 252K iterations): an eleven-fold difference in data and
2.8-fold in training, with the objective held constant. It is the closest control on data
scale that public checkpoints permit. If data scale drove our results, the two models should
separate.

\begin{table}[H]
\centering
\small
\caption{The V-JEPA generational progression. The final column marks measures computed from
frozen features with no attentive probe anywhere in them. Reversal cosine is per token, lower being
more sensitive to the direction of time; the Fisher ratio is the direction-contrastive hard
tier over the easy tier, so a value above one means the encoder separates the harder,
temporally defined distinction better than the easier one.}
\label{tab:datascale}
\begin{tabular}{lcccc}
\toprule
\textbf{Measure} & \textbf{V-JEPA\,1} & \textbf{V-JEPA\,2} & \textbf{V-JEPA\,2.1} & \textbf{No attentive probe} \\
\midrule
Pretraining clips        & $\sim$2M & 22M & 22M & --- \\
\midrule
SSv2 clean top-1 (\%)    & 70.32 & 70.10 & 71.78 & no \\
SSv2 reversal cosine     & \textbf{0.391} & 0.408 & 0.578 & yes \\
SSv2 hard/easy Fisher    & \textbf{1.17} & 1.11 & 0.95 & yes \\
SSv2 discriminability, linear probe (\%) & \textbf{47.03} & 39.39 & 39.05 & yes \\
EGTEA reversal cosine    & 0.367 & \textbf{0.366} & 0.482 & yes \\
EGTEA direction tier, linear probe (\%) & \textbf{67.76} & 66.03 & 63.12 & yes \\
\bottomrule
\end{tabular}
\end{table}

\paragraph{Data scale buys domain transfer, not temporal structure.}
Eleven times more pretraining data is worth about $+0.2$ top-1 on SSv2 and $+9.0$ on EGTEA,
while moving the reversal cosine by 0.018 on SSv2 and 0.001 on EGTEA. The gain is therefore
a domain-transfer gain: almost nothing on the benchmark both models were pretrained on, and
a great deal on an egocentric domain neither targeted. On the axes that isolate temporal
structure it buys nothing at all. This is a dissociation rather than a null result, and the
capability it fails to buy is the one this paper is about.

The other end of the data range points the same way. VideoPrism is pretrained on by far the
largest corpus of any model here, 36M captioned clips together with a much larger
weakly-labelled set, and places last on every axis that carries this paper's conclusions. An
eleven-fold reduction costs nothing and the largest corpus performs worst, so neither end of
the range supports a scale explanation.

\paragraph{Directional structure is set by the objective, not by scale.}
Semantic accuracy and robustness rise monotonically from V-JEPA\,1 to V-JEPA\,2.1, but
sensitivity to the direction of time is already saturated at the smallest data budget: it is
flat from V-JEPA\,1 to V-JEPA\,2 on both datasets and on metrics that use no probe. Every
V-JEPA generation still exceeds every non-JEPA model on the reversal and direction measures,
including the weakest V-JEPA against the strongest baseline on the direction tier (63.12
against 60.68). The arrow of time is therefore something the objective supplies from the
outset, and that more data, higher resolution and distillation from a larger teacher neither
create nor improve. Those three factors change together across the generations, so we offer
as a hypothesis, rather than a result, that the later models' longer clips and higher
resolution supply more appearance and long-range context to rely on, whereas V-JEPA\,1's
16-frame, 224\,px budget offers little except local motion.

\paragraph{A caveat that travels with these numbers.}
SSv2 makes up roughly 8.4\% of VideoMix2M against 0.76\% of VideoMix22M, so SSv2 is the more
concentrated in V-JEPA\,1's mix; this is the mirror image of the proximity between
V-JEPA\,1's instructional-video corpus and EGTEA's kitchen domain. Neither affects the two
headline results above, reversal cosine and the hard-to-easy Fisher ratio, which contain no
attentive probe at all and which are normalised within each model.

Finally, this experiment doubles as a correctness check on our integration. V-JEPA's own
paper reports 69.5\% frozen attentive-probe accuracy on SSv2 for ViT-L/16, and we measure
70.32\% under our probe. Landing within a point of a published figure is independent evidence
that the checkpoint, tower and preprocessing are correct, and it is the only such check
available in this study, since every other encoder's SSv2 number is ours rather than
published.

\section{Frozen latent-prediction features vs.\ task-adapted baselines}
\label{app:training_regime}

The main paper showed that V-JEPA leads among frozen encoders under a unified attentive probe. A key question is whether this advantage holds against task-optimised models. Since end-to-end fine-tuning uses label supervision, any remaining edge would suggest that the pretraining objective, not the probe, drives the robustness we observe. We therefore compare V-JEPA 2 (frozen backbone with attentive probe) to VideoMAE~\citep{tong2022videomae} and TimeSformer~\citep{bertasius2021space}, both fully fine-tuned on SSv2 (Table~\ref{tab:sec2_models}), across corruption robustness, fine-grained pretend action recognition, and occlusion robustness. These evaluations led to the following major insights.

\begin{table}[H]
\centering
\caption{Models compared under different training regimes using publicly available checkpoints.}
\label{tab:sec2_models}
\small
\begin{tabular}{ll}
\toprule
\textbf{Model} & \textbf{Fine-tuning on SSv2} \\
\midrule
V-JEPA 2    & Attentive-probe only \\
VideoMAE v1 & End-to-end fine-tuned \\
TimeSformer & Fully supervised \\
\bottomrule
\end{tabular}
\end{table}

\begin{figure}[H]
    \centering
    \includegraphics[width=\linewidth]{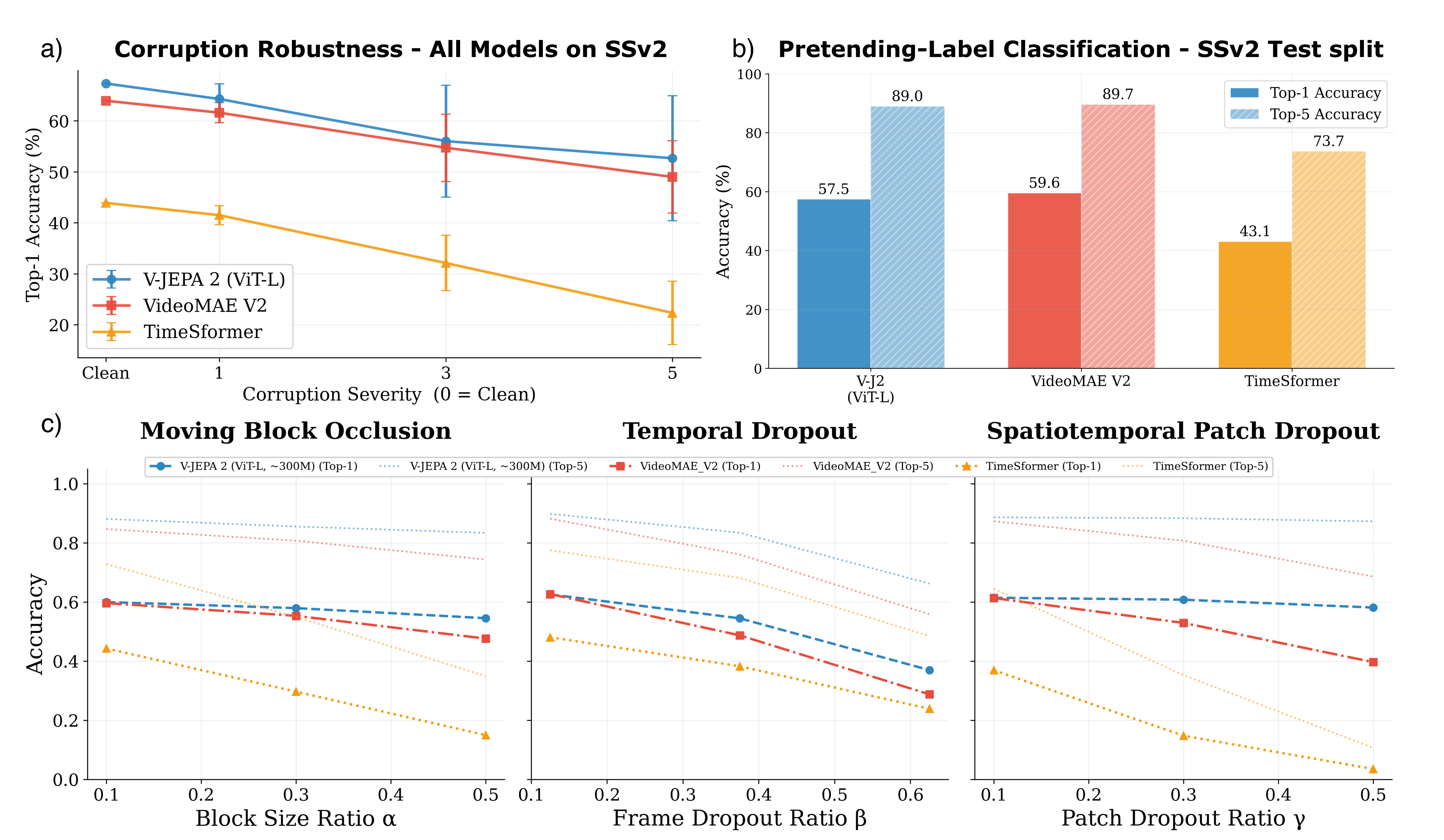}
    \caption{\textbf{(a)} Corruption robustness: Representational stability measured by classification accuracy. \textbf{(b)} Top-1 and Top-5 accuracy on SSv2 pretending action subset. \textbf{(c)} Top-1 and Top-5 accuracy of the models under three occlusion conditions with varying intensity.}
    \label{fig:ccfinetuned}
\end{figure}

\paragraph{Frozen V-JEPA retains its advantage over fine-tuned baselines
for corruption robustness.}
Figure~\ref{fig:ccfinetuned}(a) confirms that V-JEPA\,2 leads the models on
clean accuracy despite its frozen backbone. We carefully fine-tuned the baselines to get their best performance. As corruption
severity increases, the dominance of V-JEPA is  preserved, with TimeSformer
suffering the steepest accuracy collapse and the largest representational
drift, indicating that optimising for clean-condition performance may
actively increase brittleness under distributional shift. Finetuning does help in improving the variance profiles of the baselines over V-JEPA, which is expected.

\paragraph{Frozen V-JEPA remains competitive on fine-grained pretend
action recognition.}
We apply the pretending action evaluation of
Section~\ref{sec:pretend-actions} to the models in
Table~\ref{tab:sec2_models}, obtaining the results shown in Figure~\ref{fig:ccfinetuned}(b).
V-JEPA\,2 closely matches the fine-tuned VideoMAE on top-1 accuracy while
substantially outperforming TimeSformer, despite relying solely on
an attentive probe. In these results, the gap of V-JEPA 2 to a fully optimised model narrows
on the subset demanding physical detail discrimination, which
is consistent with the finding in Section~\ref{sec:pretend-actions}
that latent predictive representations retain fine-grained contact
cues in the frozen state.

\paragraph{Frozen V-JEPA representations are more occlusion-robust
than fine-tuned backbones.}
We repeat the three occlusion paradigms from
Section~\ref{sec:occlusion} on the same models
in Figure~\ref{fig:ccfinetuned}(c). V-JEPA\,2 leads in both Top-1
accuracy and representational stability across all three paradigms
and severity levels. The collapse of the fine-tuned models under
the highest occlusion settings, despite being trained directly on
the evaluation distribution, further supports the view that
task-specific adaptation optimises for clean-condition performance
at the cost of tolerance to missing information.

\section{Model comparison considerations}
\label{app:model_comparison}

Table~\ref{tab:model_details} summarises the key attributes of each encoder evaluated in this study. The four primary models were selected to span the three dominant self-supervised video paradigms at matched ViT-L capacity ($\sim$300\,M parameters), so that observed differences are most naturally attributed to the pretraining objective. Pretraining data composition and training schedule are not equalised across models, and doing so would require retraining each from scratch under identical data and compute budgets, which is beyond our scope. Data \emph{scale} we measure rather than assume: V-JEPA\,1 is listed below as a control that matches V-JEPA\,2 in every respect except an eleven-fold reduction in pretraining data, and Appendix~\ref{app:datascale} reports what that reduction costs. The consistent behavioural clustering we observe---the V-JEPA variants forming one group and the non-JEPA models another across all five evaluation axes---is difficult to attribute to data alone, particularly because V-JEPA 2 and VideoPrism both train on large-scale heterogeneous video corpora yet diverge sharply in every robustness metric.

\begin{table}[H]
\centering
\small
\caption{Detailed comparison of the evaluated encoders. All models use a ViT-Large backbone. $^\dagger$V-JEPA\,1 is the pretraining-data control of Appendix~\ref{app:datascale}; it appears in Figure~\ref{fig:teaser-radar} and that appendix, not in the primary four-encoder comparisons.}
\label{tab:model_details}
\setlength{\tabcolsep}{3.5pt}
\begin{tabular}{lllll}
\toprule
\textbf{Model} & \textbf{Params} & \textbf{Pretraining Data} & \textbf{Objective} & \textbf{Source} \\
\midrule
V-JEPA\,2.1 & $\sim$300M & Internet video ($>$1M hrs) & Latent prediction & Meta \\
V-JEPA\,2   & $\sim$300M & Internet video ($>$1M hrs) & Latent prediction & Meta \\
V-JEPA\,1$^\dagger$ & $\sim$300M & VideoMix2M ($\sim$2M clips) & Latent prediction & Meta \\
VideoPrism  & $\sim$300M & Video + video-text pairs (36M clips) & Contrastive + masked pred. & Google \\
VideoMAEv2  & $\sim$300M & UnlabeledHybrid (mix of public sets) & Pixel reconstruction & Open-source \\
\bottomrule
\end{tabular}
\end{table}

\subsection{Attentive probe training protocol}
\label{app:probe_training}

Every encoder is read out through the same attentive probe: depth 2, 8 heads, MLP ratio 2.0, approximately 15.9M parameters, trained on the same clips under the same optimisation schedule and early-stopping criterion, on both Something-Something\,v2 and EGTEA Gaze+ (encoders therefore stop at different epochs according to when their own validation accuracy plateaus). Holding the readout fixed rather than tuning it per encoder is the stricter control for our purpose: no model is decoded by a larger or better-matched head than another, so the relative ordering across models cannot be an artefact of unequal probe capacity. The probe is deliberately lightweight relative to the frozen backbones ($\sim$5\% of encoder parameters), so it can expose the structure already present in a representation without supplying much of its own.

The experimental pipeline consumed approximately 2{,}000 A100 GPU-hours in total: $\sim$450 to train the attentive probes on SSv2 and $\sim$850 to run the SSv2 perturbation suites (corruption, occlusion, temporal, pretend-action and discriminability evaluations across all encoders), giving $\sim$1{,}300 for SSv2; $\sim$150 and $\sim$250 for the corresponding probes and suites on EGTEA Gaze+; and $\sim$300 for the predictor training and closed-loop planning of Appendix~\ref{app:downstream}.

\subsection{Fine-tuned model checkpoints}
\label{app:finetuned_checkpoints}

For the training-regime comparison in Appendix~\ref{app:training_regime}, we rely exclusively on publicly released checkpoints that have been adapted to Something-Something\,v2, in order to avoid confounds introduced by our own fine-tuning hyperparameters. Because no single provider releases matched ViT-L fine-tuned checkpoints across all three training regimes on SSv2, the backbones differ slightly in capacity and input configuration; we describe each below.

\begin{itemize}
    \item \textbf{V-JEPA\,2 (frozen backbone, attentive-probe only).} We use the ViT-L/16 V-JEPA\,2 encoder released by Meta, with 64 frames at $256\times256$ resolution.

    \item \textbf{VideoMAE (end-to-end fine-tuned).} We use the \texttt{videomae-base-finetuned-ssv2} checkpoint released by MCG-NJU, which is pre-trained for 2400 epochs with masked pixel reconstruction and then end-to-end fine-tuned on SSv2 with full supervision. This is the only VideoMAE SSv2 checkpoint publicly available at the time of writing; it is a ViT-B/16 backbone consuming 16 frames at $224\times224$, and reaches 70.6\% top-1 and 92.6\% top-5 on the SSv2 test set.
    \item \textbf{TimeSformer (fully supervised).} We use the high-frame-count TimeSformer variant trained end-to-end on SSv2 from ImageNet-21k initialisation, consuming 64 frames at $224\times224$ with divided space-time attention. No self-supervised stage is involved; the backbone is optimised jointly with the classification head under label supervision.
\end{itemize}

All three models are evaluated on the same perturbation suite used for the frozen encoders (Sections~\ref{sec:corruption}, \ref{sec:pretend-actions}, and \ref{sec:occlusion}), with perturbations applied to the pre-processed input at the resolution and frame count native to each checkpoint. Because the VideoMAE checkpoint is ViT-B rather than ViT-L, the comparison in Appendix~\ref{app:training_regime} is a capability-level rather than a capacity-matched one, and the question it answers is whether a frozen latent-prediction encoder is competitive with the best publicly available task-adapted models rather than which backbone size is preferable. The TimeSformer baseline is the capacity-comparable reference point: it consumes 64 frames, is optimised end-to-end under full supervision on the evaluation task itself, and is nonetheless the model that collapses fastest under corruption and occlusion.

\section{Feature discriminability: supplementary analysis}
\label{app:intrinsic}

\subsection{Action class list}

Details are shown in Table \ref{tab:class_list}
\begin{table}[t]
\centering
\caption{The 30 SSv2 action classes used in the encoder discriminability analysis, grouped by semantic category and ordered by increasing semantic difficulty (different-verb $<$ same-verb $<$ pretend-vs-real).}
\label{tab:class_list}
\small
\begin{tabular}{ll}
\toprule
\textbf{Semantic Category} & \textbf{Action Class} \\
\midrule
\textit{Different Verbs} & Burying something in something \\
 & Closing something \\
 & Folding something \\
 & Hitting something with something \\
 & Plugging something into something \\
 & Removing something, revealing something behind \\
 & Spinning something so it continues spinning \\
 & Stacking number of something \\
 & Tearing something into two pieces \\
 & Trying but failing to attach something to something \\
\midrule
\textit{Same Verb} & Moving away from something with your camera \\
 & Moving part of something \\
 & Poking a hole into something soft \\
 & Poking a stack of something so the stack collapses \\
 & Pulling something from behind of something \\
 & Pulling something from left to right \\
 & Pushing something from left to right \\
 & Pushing something from right to left \\
 & Putting something and something on the table \\
 & Putting something behind something \\
\midrule
\textit{Pretend vs Real} & Poking a hole into some substance \\
& Pretending to poke something \\
 & Pouring something into something \\
 & Pretending to pour something out of something \\
 & Putting number of something onto something \\
 & Pretending to put something behind something \\
 & Throwing something \\
 & Pretending to throw something \\
 & Turning something upside down \\
 & Pretending to turn something upside down \\
\bottomrule
\end{tabular}
\end{table}

\subsection{Architecture attribution}

\begin{figure}[H]
    \centering
    \includegraphics[width=\textwidth]{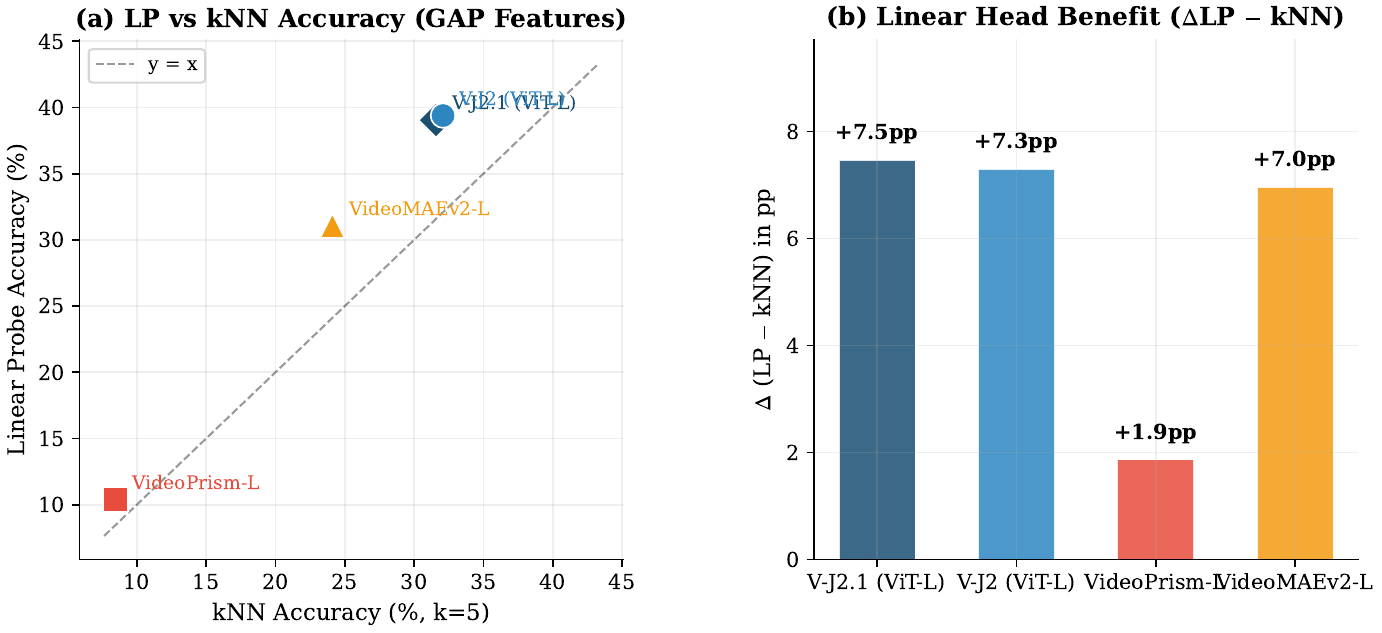}
    \caption{Architecture attribution via GAP features. \textbf{(a)}~Linear probe vs kNN accuracy;
    points above the diagonal indicate a linear head adds value beyond geometric proximity.
    \textbf{(b)}~The gap $\Delta$(LP $-$ kNN) is largest for V-JEPA models, suggesting their
    representations encode direction-dependent semantic structure exploitable by a linear classifier.}
    \label{fig:app_attribution}
\end{figure}

\section{Corruption robustness: supplementary analysis}
\label{app:corruption}

\subsection{Visual examples of corruptions}
\begin{figure} [H]
  \centering
  \includegraphics[width=\linewidth]{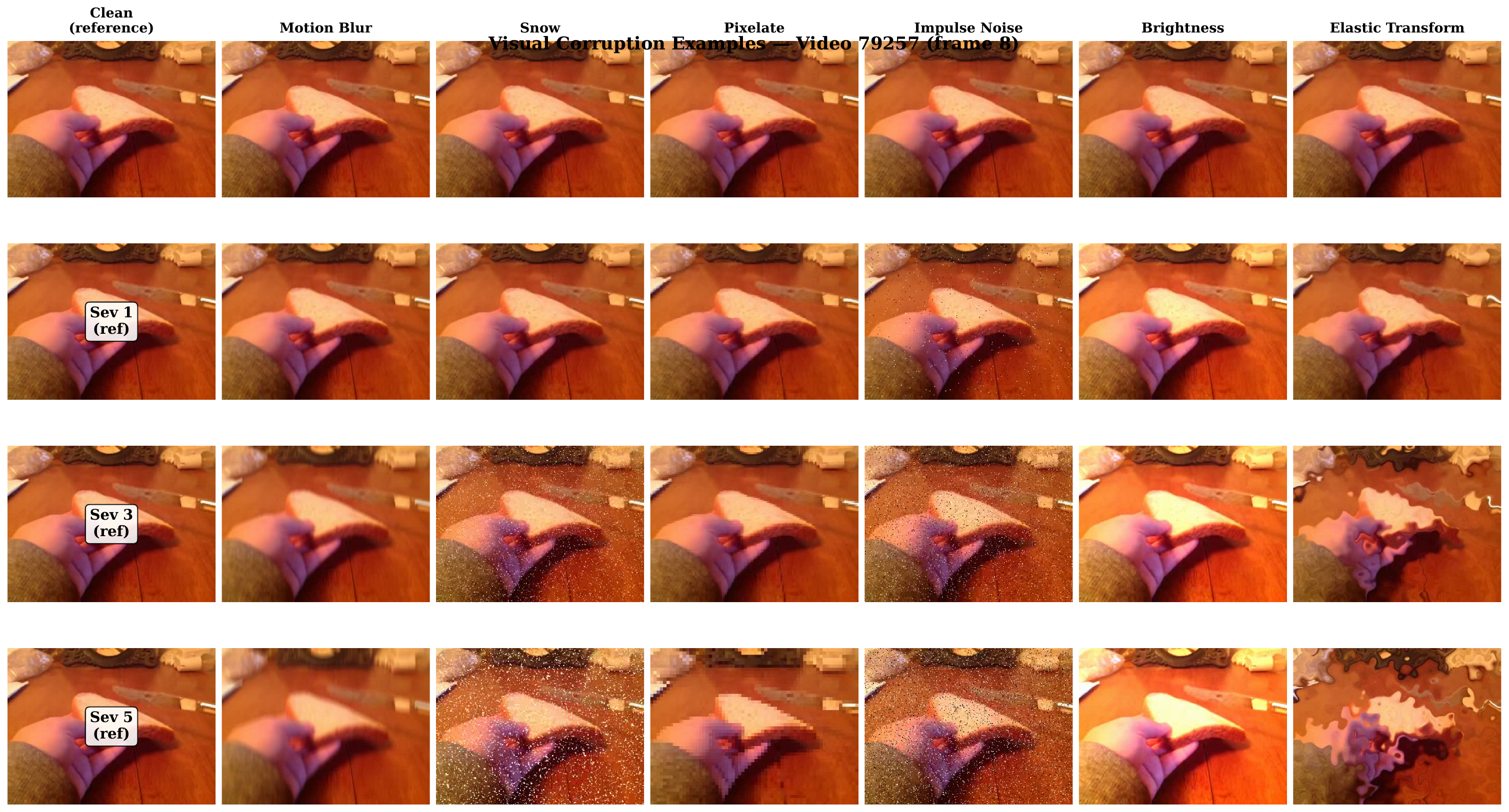}
  \caption{Visual examples of the six corruption types applied to a single SSv2 video frame
  at severities 1, 3, and 5. Columns correspond to corruption types and rows to increasing
  severity, illustrating the progressive degradation of visual content.}
  \label{fig:corruption-examples}
\end{figure}

Figure~\ref{fig:corruption-examples} illustrates the six corruption types at all three severity levels applied to a sample video frame, providing visual reference for the perturbation intensities used throughout the experiment.

\subsection{Per-corruption accuracy retention at maximum severity}
\begin{figure}[H]
\centering
\includegraphics[width=0.85\linewidth]{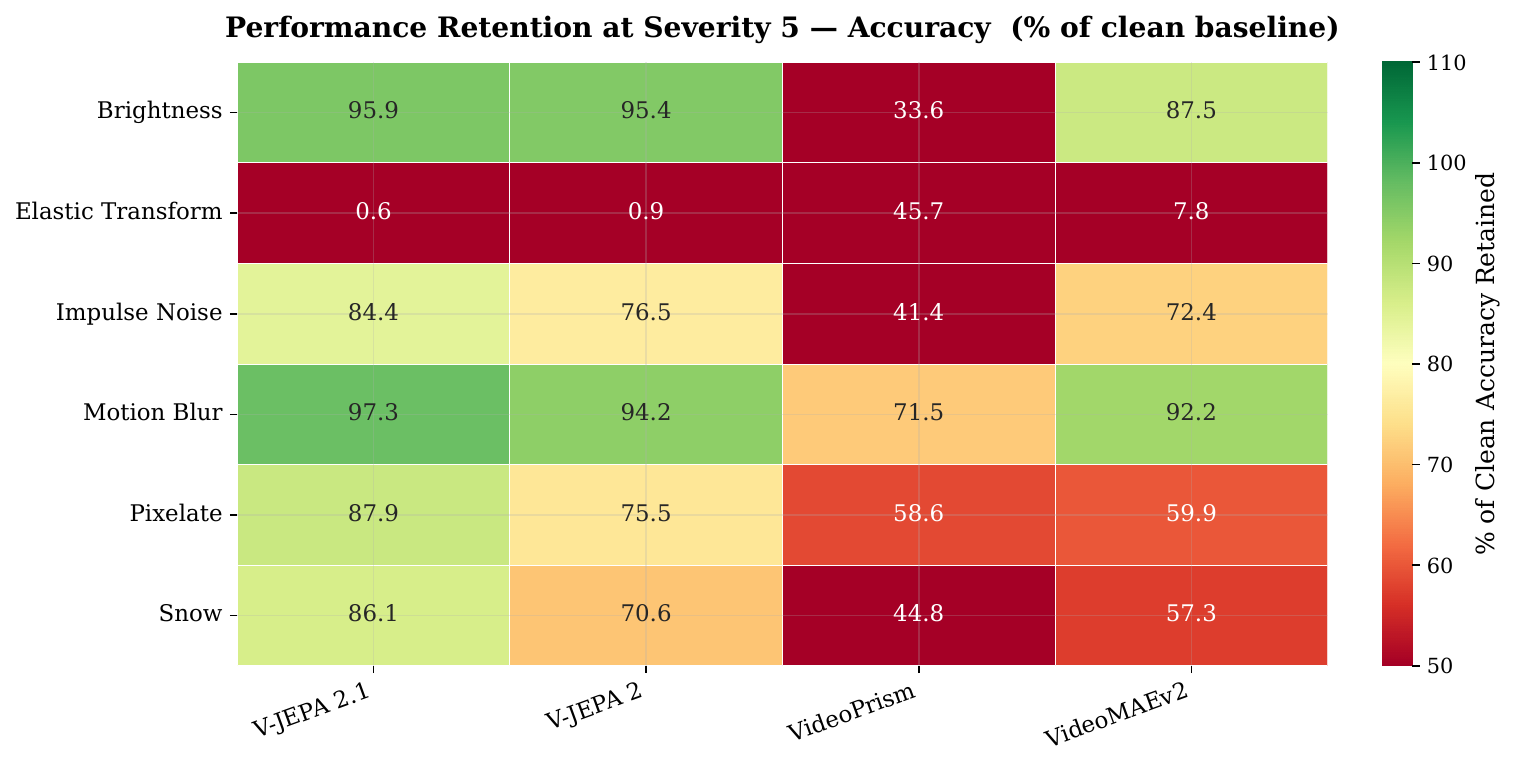}
\caption{Per-corruption accuracy retention at severity 5 (\% of clean baseline retained). V-JEPA 2.1 leads on five of six corruption types, while elastic transform is universally the most damaging.}
\label{fig:retention-heatmap-app}
\end{figure}

Figure~\ref{fig:retention-heatmap-app} reports accuracy retention at severity 5 as a percentage of each model's clean baseline. V-JEPA 2.1 leads on five of six corruption types. Elastic transform is the universal outlier, reducing all models to near-zero or single-digit retention, with VideoMAEv2 exhibiting the most severe collapse. Brightness and motion blur are the least disruptive corruptions for both V-JEPA variants, consistent with the expectation that photometric perturbations preserving temporal structure are well handled by joint-embedding objectives.

\subsection{Cosine similarity retention}
\begin{figure}[H]
\centering
\includegraphics[width=0.85\linewidth]{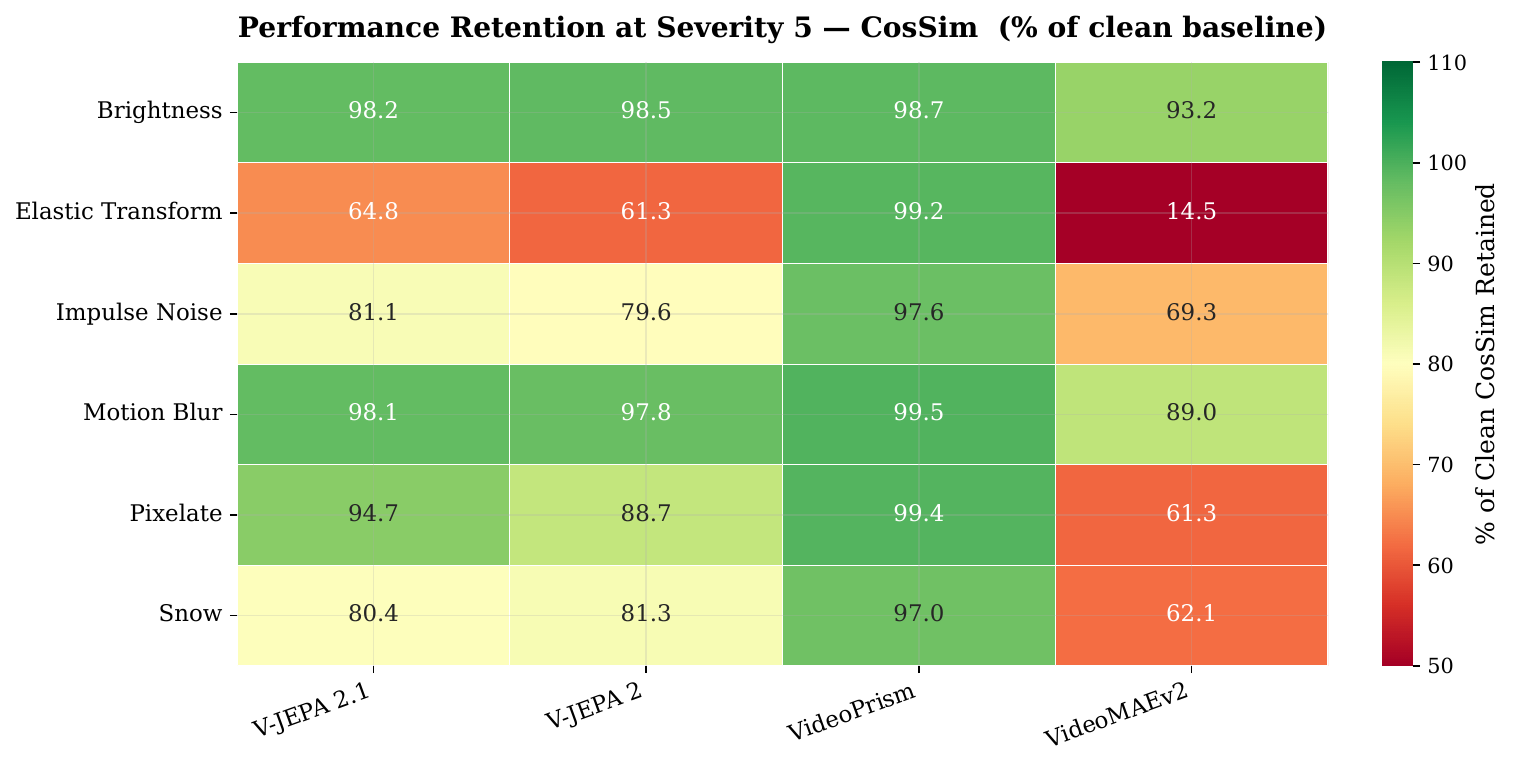}
\caption{Per-corruption cosine similarity retention at severity 5 (\% of clean baseline). VideoPrism maintains near-perfect representational stability despite declining accuracy, while VideoMAEv2 exhibits severe feature drift under elastic transform.}
\label{fig:cossim-heatmap-app}
\end{figure}

Figure~\ref{fig:cossim-heatmap-app} shows cosine similarity retention at severity 5. VideoPrism retains near-perfect similarity across all corruptions, yet this stability does not prevent its classification accuracy from declining sharply (Figure~\ref{fig:retention-heatmap-app}). VideoMAEv2 shows the largest representational drift under elastic transform, confirming that pixel-reconstruction encoders are particularly sensitive to geometric distortions at the feature level.

\subsection{Per-model degradation profiles}
\begin{figure}[H]
\centering
\includegraphics[width=\linewidth]{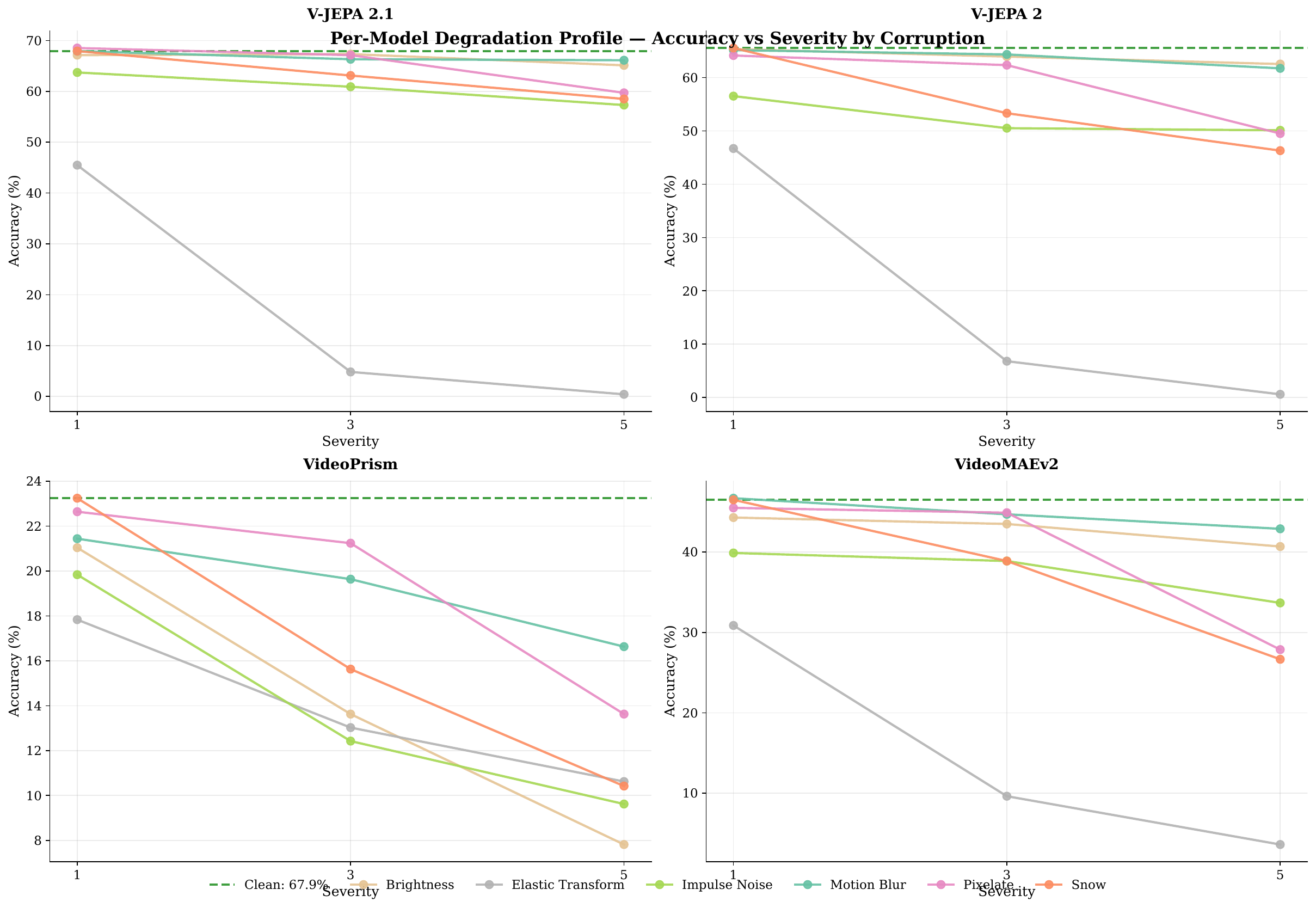}
\caption{Per-model accuracy as a function of severity for each corruption type. Most corruptions cluster together for V-JEPA variants, while elastic transform is a clear outlier for all models except VideoPrism.}
\label{fig:degradation-profiles-app}
\end{figure}

Figure~\ref{fig:degradation-profiles-app} decomposes the aggregate severity curves into per-corruption trajectories for each model. For V-JEPA 2.1 and V-JEPA 2, most corruptions cluster in a narrow band of gradual decline, with elastic transform as a clear outlier dropping steeply. VideoMAEv2 shows a wider spread among corruption types, reflecting greater sensitivity to the nature of the perturbation. VideoPrism exhibits uniformly low absolute accuracy across all corruptions, making relative differences less pronounced.

\subsection{Degradation slopes}
\begin{figure}[H]
\centering
\includegraphics[width=\linewidth]{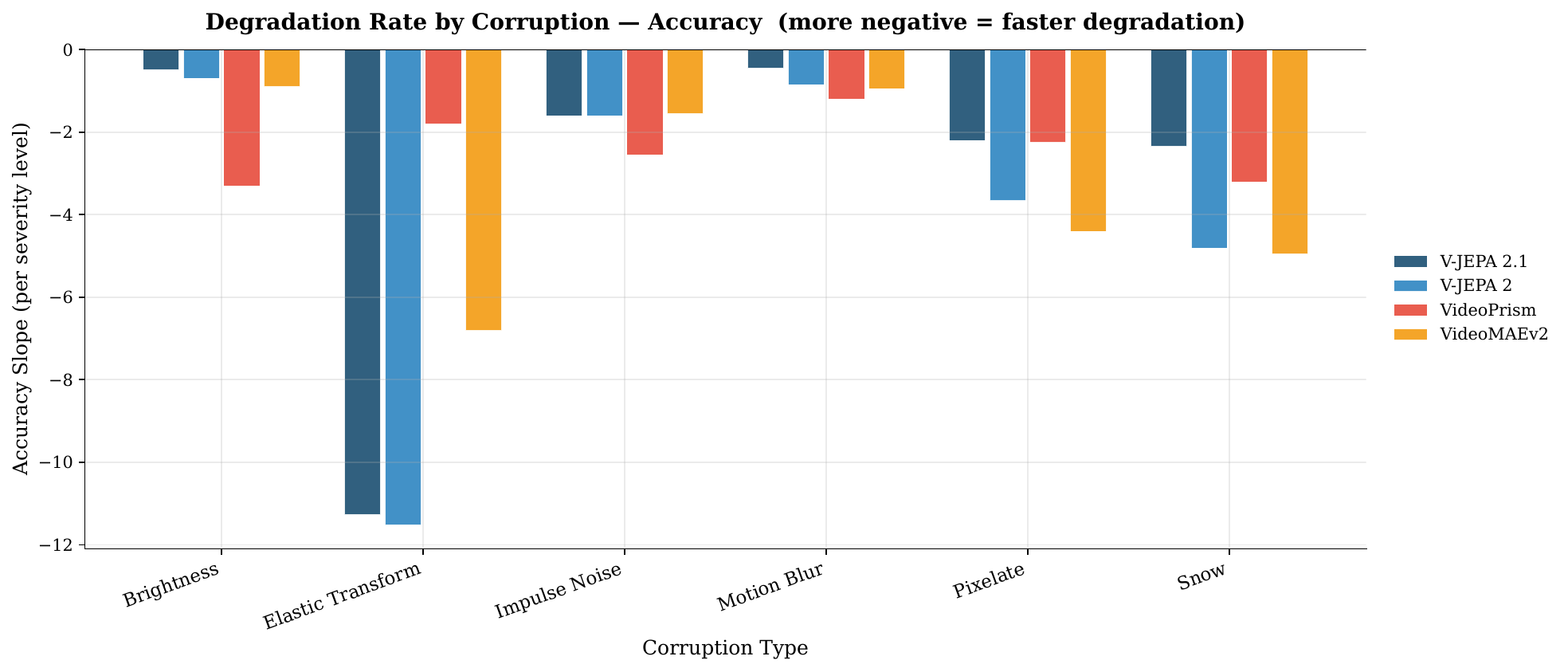}
\caption{Accuracy degradation slope per corruption type (pp per severity level; more negative indicates faster decline). V-JEPA 2.1 consistently shows the shallowest degradation.}
\label{fig:degradation-slopes-app}
\end{figure}

Figure~\ref{fig:degradation-slopes-app} reports the accuracy degradation slope for each model and corruption type, measured in percentage points per severity level. Elastic transform produces the steepest slopes for V-JEPA 2, V-JEPA 2.1, and VideoMAEv2, while brightness is the steepest for VideoPrism. V-JEPA 2.1 exhibits shallower slopes than V-JEPA 2 on every corruption type, quantifying the robustness improvement of the newer release.

\section{Fine-grained action discrimination: supplementary analysis}
\label{app:pretend}

\subsection{Class categorisation}
\label{app:pretend-categories}

We partition the 22 pretending action classes along two dimensions using a
verb-based heuristic applied to the SSv2 label text. \textbf{Object size}
is assigned based on the primary verb: actions involving manipulation of
small items (e.g., \textit{pick}, \textit{poke}, \textit{squeeze},
\textit{pour}, \textit{wipe}) are labelled \textit{small}; actions
involving state changes to moderate-scale objects (e.g., \textit{twist},
\textit{tear}, \textit{open}, \textit{close}, \textit{take}) are labelled
\textit{medium}; and actions involving large spatial displacement (e.g.,
\textit{throw}) are labelled \textit{large}. \textbf{Detail sensitivity}
captures whether the pretend variant is distinguishable only through
fine-grained contact cues (\textit{high}) or through coarser whole-body or
object-level motion (\textit{low}). The complete mapping is given in
Table~\ref{tab:pretend-categories}.

\begin{table}[H]
\centering
\small
\caption{Categorisation of the 22 pretend action classes by object size and
detail sensitivity.}
\label{tab:pretend-categories}
\begin{tabular}{lcc}
\toprule
\textbf{Action Class} & \textbf{Object Size} & \textbf{Detail Sensitivity} \\
\midrule
P. or failing to wipe something off of something & small & high \\
P. to pick something up & small & high \\
P. to poke something & small & high \\
P. to pour something out of something\ldots & small & high \\
P. to put something behind something & small & high \\
P. to put something into something & small & high \\
P. to put something next to something & small & high \\
P. to put something on a surface & small & high \\
P. to put something onto something & small & high \\
P. to put something underneath something & small & high \\
P. to spread air onto something & small & high \\
P. to sprinkle air onto something & small & high \\
P. to squeeze something & small & high \\
P. to scoop something up with something & small & low \\
\midrule
P. or trying and failing to twist something & medium & low \\
P. to be tearing something that is not tearable & medium & low \\
P. to close something without actually closing it & medium & low \\
P. to open something without actually opening it & medium & low \\
P. to take something from somewhere & medium & low \\
P. to take something out of something & medium & low \\
P. to turn something upside down & medium & low \\
\midrule
P. to throw something & large & low \\
\bottomrule
\end{tabular}
\end{table}

\subsection{Hardest pretend actions}
\label{app:pretend-hardest}

\begin{figure}[H]
    \centering
    \includegraphics[width=\linewidth]{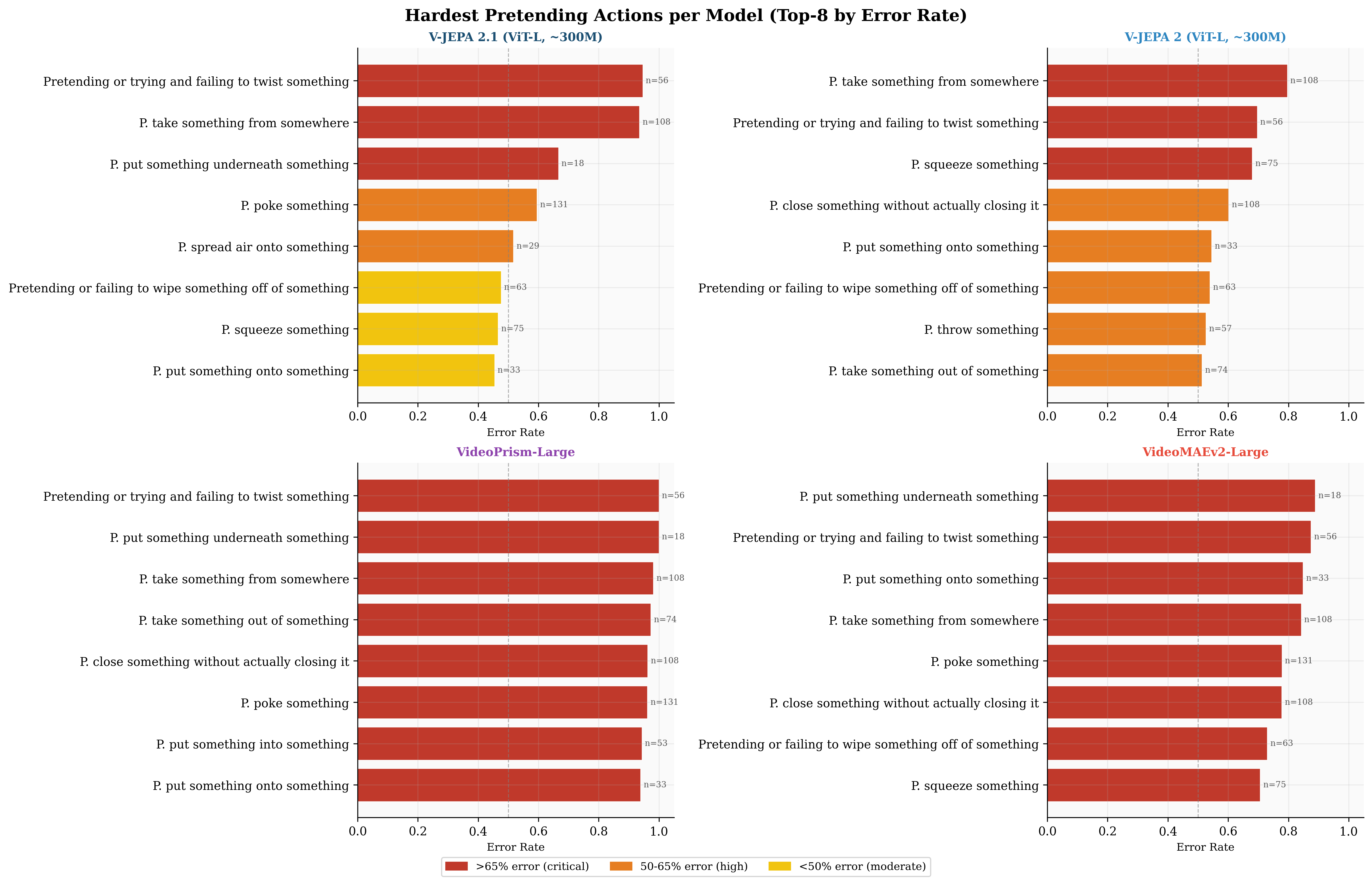}
    \caption{Top eight hardest pretend action classes per model by error rate,
    colour-coded by severity (red: critical, orange: high, yellow: moderate).}
    \label{fig:pretend-hardest}
\end{figure}

Figure~\ref{fig:pretend-hardest} reports the top eight hardest pretending
action classes per model, ranked by error rate. ``Pretending or trying and
failing to twist something'' and ``Pretending to take something from
somewhere'' appear among the top three for all four models, confirming that
these failures are architecture-agnostic and arise from inherent kinematic
ambiguity in the action classes themselves.

\subsection{Per-class accuracy delta}
\label{app:pretend-delta}

\begin{figure}[H]
    \centering
    \includegraphics[width=\linewidth]{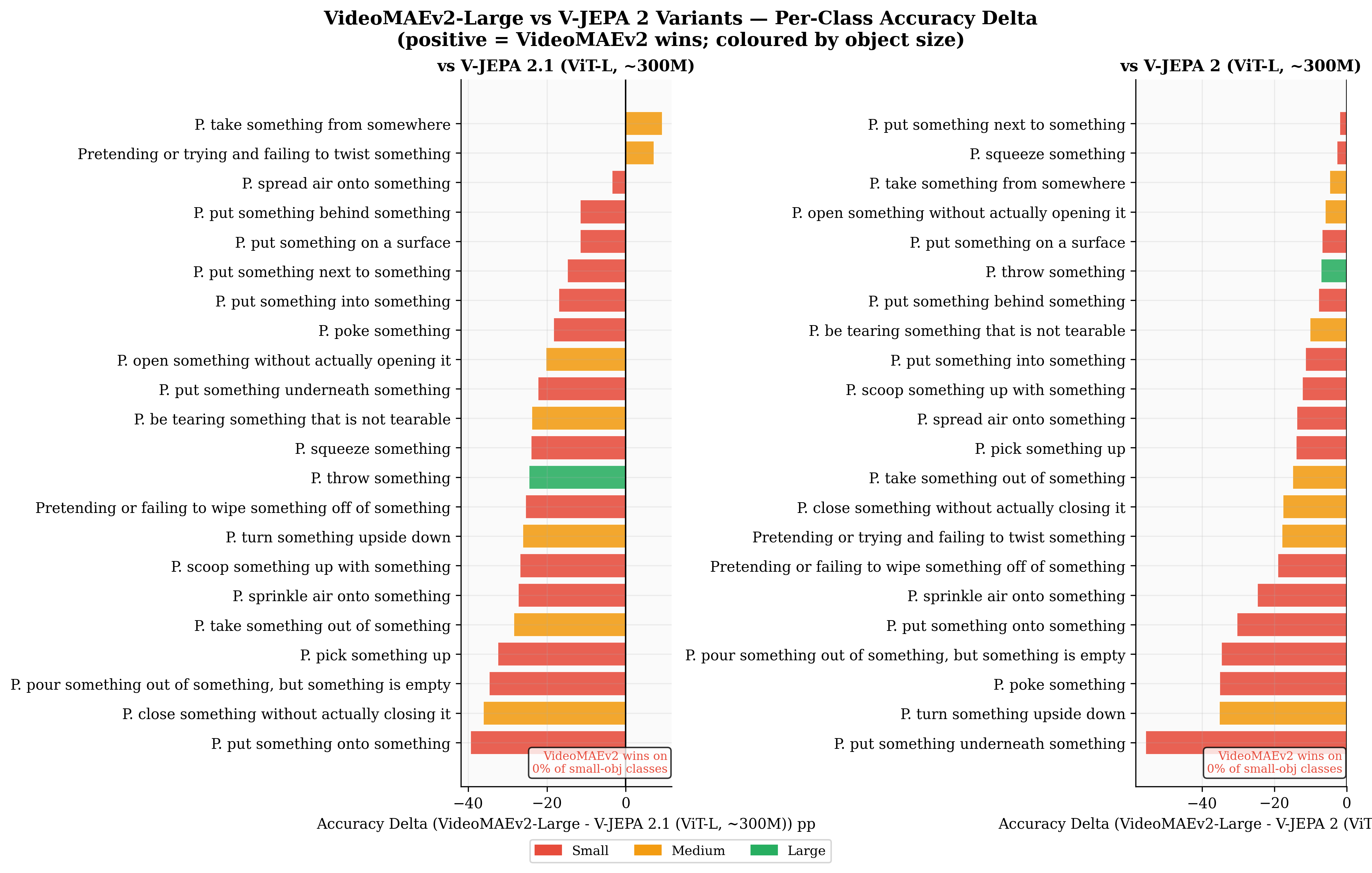}
    \caption{Per-class accuracy delta (VideoMAEv2-Large minus V-JEPA variant;
    positive values indicate VideoMAEv2 wins), colour-coded by object size.}
    \label{fig:pretend-delta}
\end{figure}

Figure~\ref{fig:pretend-delta} presents the per-class accuracy difference
between the appearance-focused model (VideoMAEv2-Large) and each
latent-space variant, with bars colour-coded by object size. Negative values
indicate that the latent model outperforms. The latent-space approach
dominates across virtually all action classes and object size strata, with
the appearance-focused model winning none of the small-object classes against
either V-JEPA variant and at most two medium-object classes against V-JEPA
2.1.

\section{Occlusion robustness: supplementary analysis}
\label{app:occlusion_hparams}

\subsection{Experimental hyperparameters}
\label{app:occlusion_hparams_table}

\begin{table}[H]
\centering
\small
\caption{Hyperparameters for the occlusion robustness benchmark. All severities
are evaluated on the same set of clean clips so that pairs are matched at the
clip level before computing RSI, CCR, and probe accuracy. The three severity
levels are indexed 1/3/5 throughout, matching the corruption scale of
\citet{hendrycks2019benchmarking}.}
\label{tab:occlusion_hparams}
\begin{tabular}{ll}
\toprule
\textbf{Component} & \textbf{Setting} \\
\midrule
Dataset                 & Something-Something\,v2, test split \\
Number of clips         & 1{,}720 \\
Number of classes       & 174 (10 clips per class) \\
Clip sampling           & uniform, at each encoder's native clip length \\
Random seed             & 42 \\
Encoders evaluated      & V-JEPA\,2.1 (ViT-L, $\sim$300M), V-JEPA\,2 (ViT-L, $\sim$300M), \\
                        & VideoPrism-Large, VideoMAEv2-Large \\
\midrule
Moving Block ratio $\alpha$       & $\{0.10, 0.30, 0.50\}$ of $\min(H,W)$ \\
Temporal Dropout ratio $\beta$    & $\{0.125, 0.375, 0.625\}$ of $T$ \\
Patch Dropout ratio $\gamma$      & $\{0.10, 0.30, 0.50\}$ of 3D cuboids \\
Patch Dropout cuboid size         & $(\tau{=}2, p{=}16, p{=}16)$ \\
\midrule
Attentive probe depth   & 2 \\
Attentive probe heads   & 8 \\
Attentive probe MLP ratio & 2.0 \\
kNN probe neighbours    & $k=5$, cosine metric, standardised features \\
\bottomrule
\end{tabular}
\end{table}

\subsection{Metric definitions}
For paired clean and occluded embeddings $f_i^{c}, f_i^{o}\in\mathbb{R}^{D}$ over
$N$ clips:
\begin{align*}
\mathrm{RSI} &= \frac{1}{N}\sum_{i=1}^{N}
   \frac{\langle f_i^{c}, f_i^{o}\rangle}{\lVert f_i^{c}\rVert\,\lVert f_i^{o}\rVert}, \\
\mathrm{CCR} &= \frac{1}{N}\sum_{i=1}^{N}
   \mathbf{1}\!\bigl[\hat{y}_{\mathrm{kNN}}(f_i^{c}) = \hat{y}_{\mathrm{kNN}}(f_i^{o})\bigr], \\
\mathrm{AUC(RSI)} &= \int_{0}^{s_{\max}} \mathrm{RSI}(s)\,ds.
\end{align*}
The kNN classifier is fit on clean features and reused at evaluation, so CCR
measures whether the same clip is routed to the same neighbourhood.

\subsection{Visualisations of each occlusion paradigm}
\label{app:occlusion_viz}

\begin{figure}[H]
  \centering
  \includegraphics[width=0.95\linewidth]{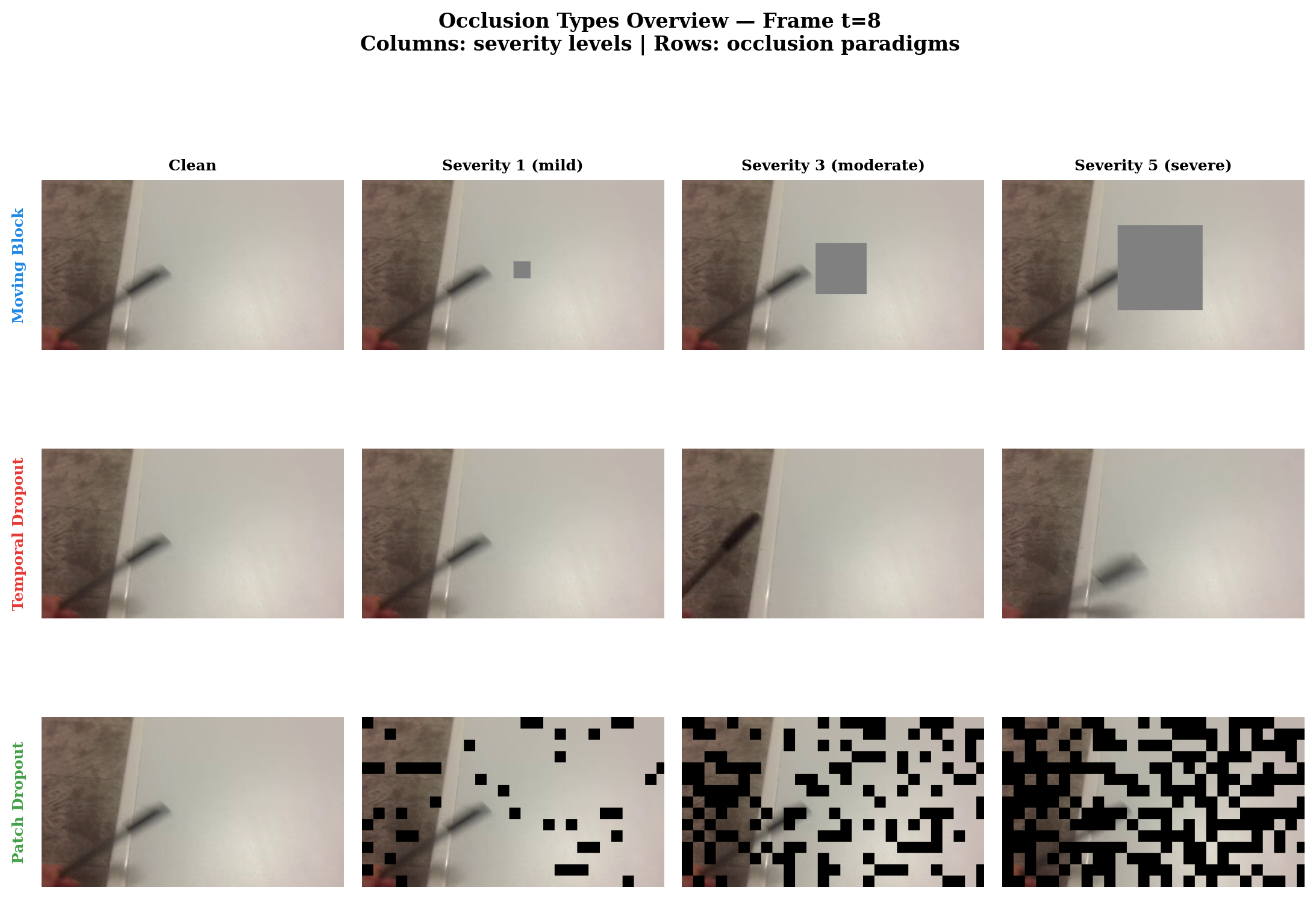}
  \caption{Overview of the three occlusion paradigms at increasing severity on
  a single representative clip.}
  \label{fig:occ_overview}
\end{figure}

\begin{figure}[H]
  \centering
  \includegraphics[width=0.95\linewidth]{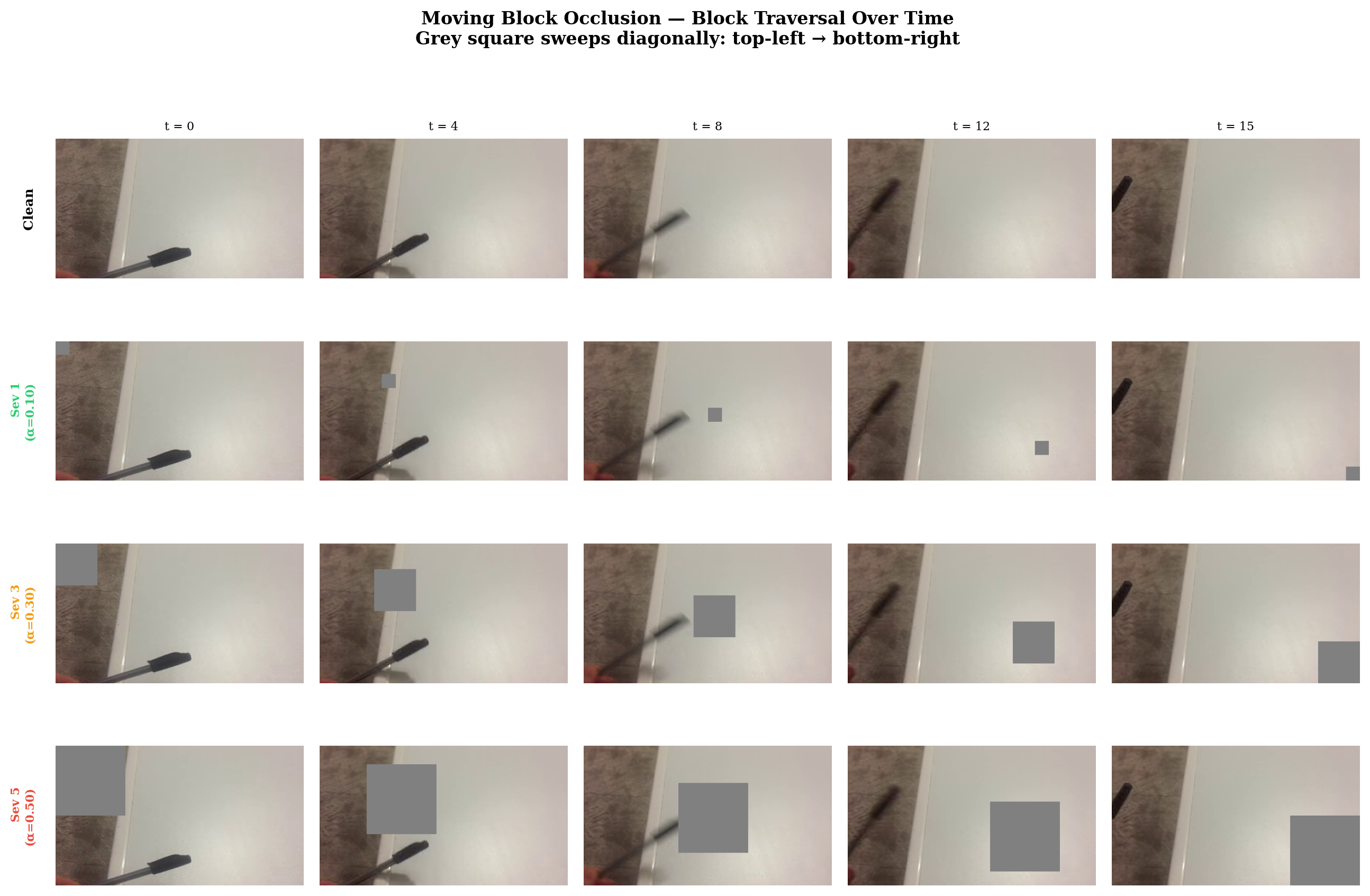}
  \caption{Moving Block at five timestamps. The grey square traverses the
  frame diagonally so that any single moment occludes a different region. The
  block side is $\alpha\cdot\min(H,W)$.}
  \label{fig:viz_moving_block}
\end{figure}

\begin{figure}[H]
  \centering
  \includegraphics[width=0.95\linewidth]{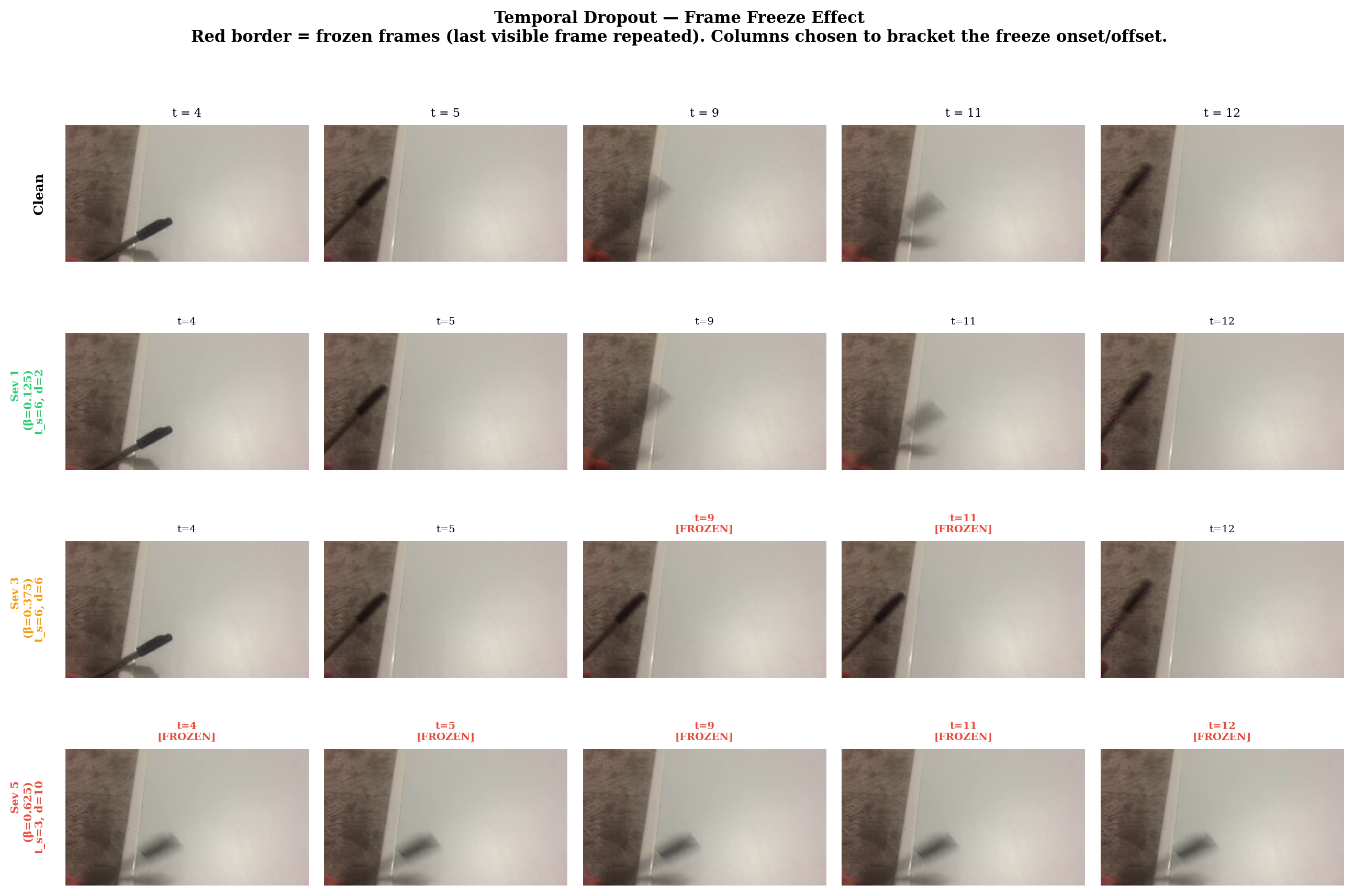}
  \caption{Temporal Dropout. A contiguous block of $\lfloor\beta T\rfloor$
  frames is replaced by repeating the last visible frame. We show frames just
  before the gap, at the onset, mid gap, and at recovery.}
  \label{fig:viz_temporal_dropout}
\end{figure}

\begin{figure}[h]
  \centering
  \includegraphics[width=0.95\linewidth]{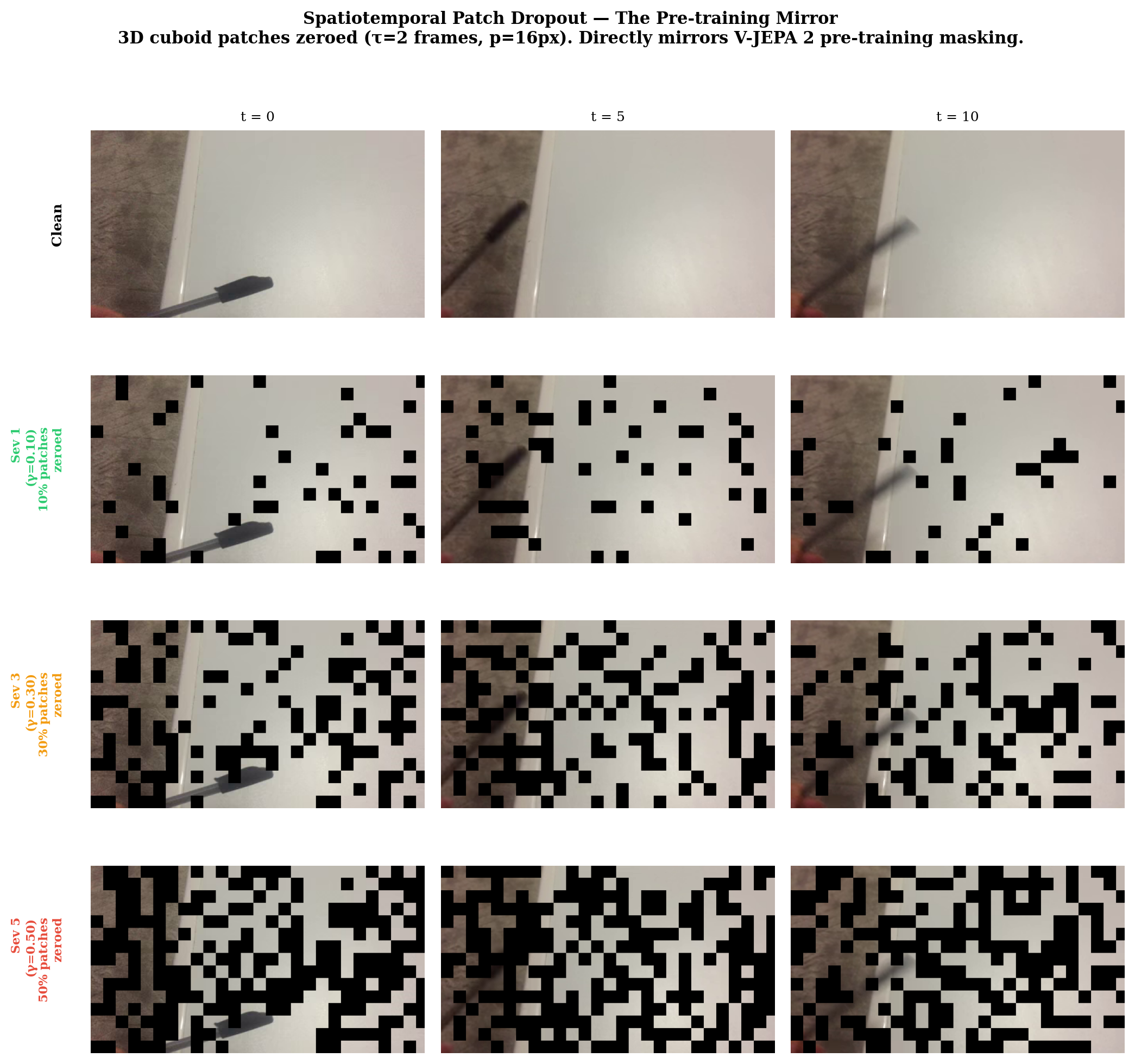}
  \caption{Spatiotemporal Patch Dropout. The volume is partitioned into 3D
  cuboids of size $(2,16,16)$ and a fraction $\gamma$ are zeroed. This is
  the closest analogue at inference time to the masked patches used in
  V-JEPA pre-training.}
  \label{fig:viz_patch_dropout}
\end{figure}

\subsection{Additional analyses}
\label{app:occlusion_extra}

\paragraph{Severity slopes.}
Linear regression of RSI on the severity parameter gives a per model degradation
rate. Steeper negative slopes indicate faster collapse.

\begin{table}[H]
\centering
\small
\caption{RSI degradation slopes per occlusion family. All slopes have $R^2>0.82$.}
\label{tab:slopes}
\begin{tabular}{lccc}
\toprule
\textbf{Model} & \textbf{Moving Block} & \textbf{Temporal Dropout} & \textbf{Patch Dropout} \\
\midrule
V-JEPA\,2.1   & $-0.072$ & $-0.047$ & $-0.437$ \\
V-JEPA\,2     & $-0.089$ & $-0.088$ & $-0.335$ \\
VideoPrism    & $-0.015$ & $-0.020$ & $-0.043$ \\
VideoMAEv2   & $-0.114$ & $-0.329$ & $-0.701$ \\
\bottomrule
\end{tabular}
\end{table}

\paragraph{Wilcoxon signed-rank tests.}
We compare V-JEPA\,2.1 against each baseline across all nine occlusion
$\times$ severity cells. V-JEPA\,2.1 is significantly more stable than both
V-JEPA\,2 and VideoMAEv2 at $p<0.01$. Against VideoPrism the test is
significant in the opposite direction, at the same $p$: VideoPrism's embeddings
move less than anyone's under every occlusion cell. That is the decoupling the
main section describes, in its sharpest form; the accuracy panels of
Figure~\ref{fig:occlusion_composite} show what that stability is worth.

\begin{table}[H]
\centering
\small
\caption{One sided Wilcoxon signed-rank tests on per cell RSI differences. A
positive $\overline{\Delta}$ means V-JEPA\,2.1 is the more stable of the pair.
With $N=9$, $1.95\!\times\!10^{-3}$ is the smallest attainable one-sided $p$.}
\label{tab:wilcoxon}
\begin{tabular}{llcccc}
\toprule
\textbf{Reference} & \textbf{Baseline} & $N_{\text{pairs}}$ & $\overline{\Delta}$ & $W$ & $p$ \\
\midrule
V-JEPA\,2.1 & V-JEPA\,2     & 9 & $+0.0155$ & $45.0$ & $1.95\!\times\!10^{-3}$ \\
V-JEPA\,2.1 & VideoPrism    & 9 & $-0.0446$ & $0.0$  & $1.95\!\times\!10^{-3}$ (reversed) \\
V-JEPA\,2.1 & VideoMAEv2   & 9 & $+0.0616$ & $45.0$ & $1.95\!\times\!10^{-3}$ \\
\bottomrule
\end{tabular}
\end{table}

\paragraph{CCR vs RSI decoupling.}
For each model and occlusion family we compute the decoupling index
$\mathrm{DI} = \mathbb{E}_{s}\,\lvert\mathrm{CCR}(s) - \mathrm{RSI}(s)\rvert$.
A high DI means the classifier disagrees with itself even though the embedding
is stable. VideoPrism has the largest DI under Patch Dropout because its RSI
remains near $0.98$ while CCR drops below $0.10$. This is the quantitative
form of the conclusion in the main section: representation stability does not
imply decision stability.

\begin{figure}[H]
  \centering
  \includegraphics[width=0.7\linewidth]{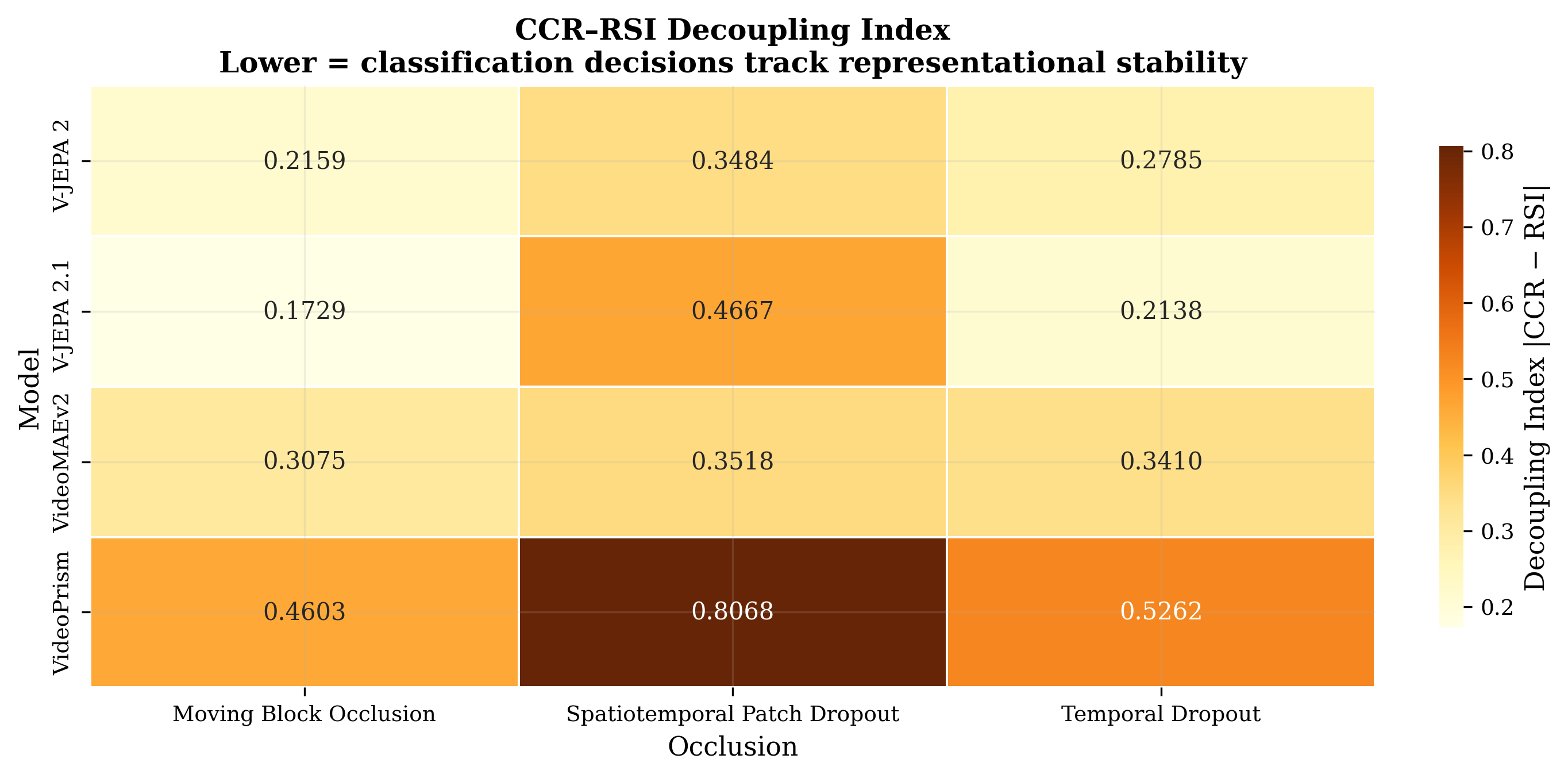}
  \caption{Decoupling index by model and occlusion type. VideoPrism shows the
  largest gap between RSI and CCR under Patch Dropout, despite having the
  highest raw RSI of any model.}
  \label{fig:decoupling}
\end{figure}

\paragraph{Worst-case behaviour.}
For every encoder the worst case is Patch Dropout at severity 5. The pattern there is the
decoupling in its extreme form: VideoPrism, the most stable encoder by RSI, ends with a CCR
of 7.2\% and 2.7\% probe top-1 accuracy, while V-JEPA\,2.1 shifts substantially further in
embedding space yet retains 46.1\%.

\section{Temporal robustness: supplementary analysis}
\label{app:temporal_extra}

\subsection{Experimental setup}
\label{app:temporal_hparams}

We evaluate V-JEPA\,2.1 (ViT-L), V-JEPA\,2 (ViT-L), VideoMAEv2-Large, and VideoPrism-Large on a class-balanced 1{,}738-clip subset of Something-Something\,v2 (174 classes, 10 clips per class, seed 42), with each encoder run at its native clip length. Ten perturbation conditions span four families: \emph{Permutation} (random shuffle, segment shuffle, interleaved reordering), \emph{Static} (first, middle, or last frame repeated), \emph{Noise} (Gaussian, uniform, temporally static Gaussian), and \emph{Reversal}. Clean top-1 accuracies are 63.6\%, 61.6\%, 43.5\%, and 20.6\% respectively, preserving the ordering of the full-split accuracies in Table~\ref{tab:egtea_clean}.

\subsection{Frame permutation}
\label{app:temporal_permutation}

\begin{figure}[H]
  \centering
  \includegraphics[width=\linewidth]{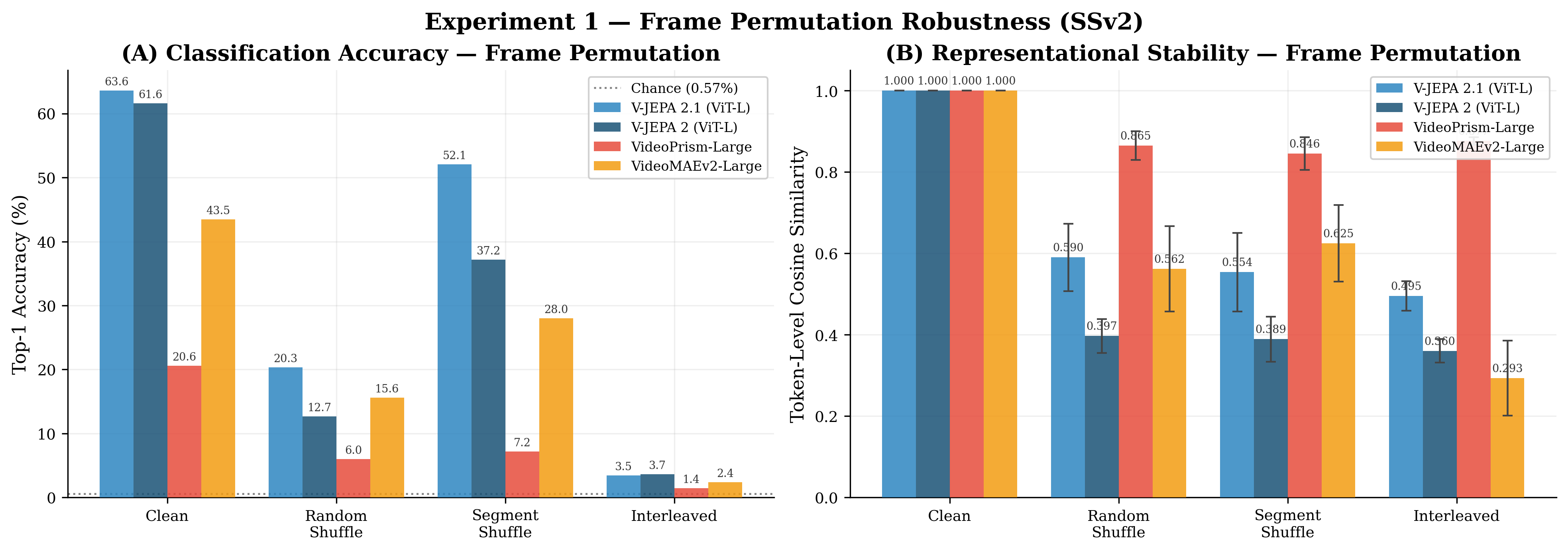}
  \caption{Classification accuracy and representational stability under three
  permutation granularities. Segment-level reordering is tolerated by all
  encoders; interleaving collapses both accuracy and cosine similarity for
  VideoMAEv2.}
  \label{fig:app_permutation}
\end{figure}

Figure~\ref{fig:app_permutation} confirms the macro-versus-micro dissociation
across all four encoders. Segment-level shuffling preserves coarse temporal
context and retains moderate accuracy, whereas interleaving destroys it.

\begin{figure}[H]
  \centering
  \includegraphics[width=0.85\linewidth]{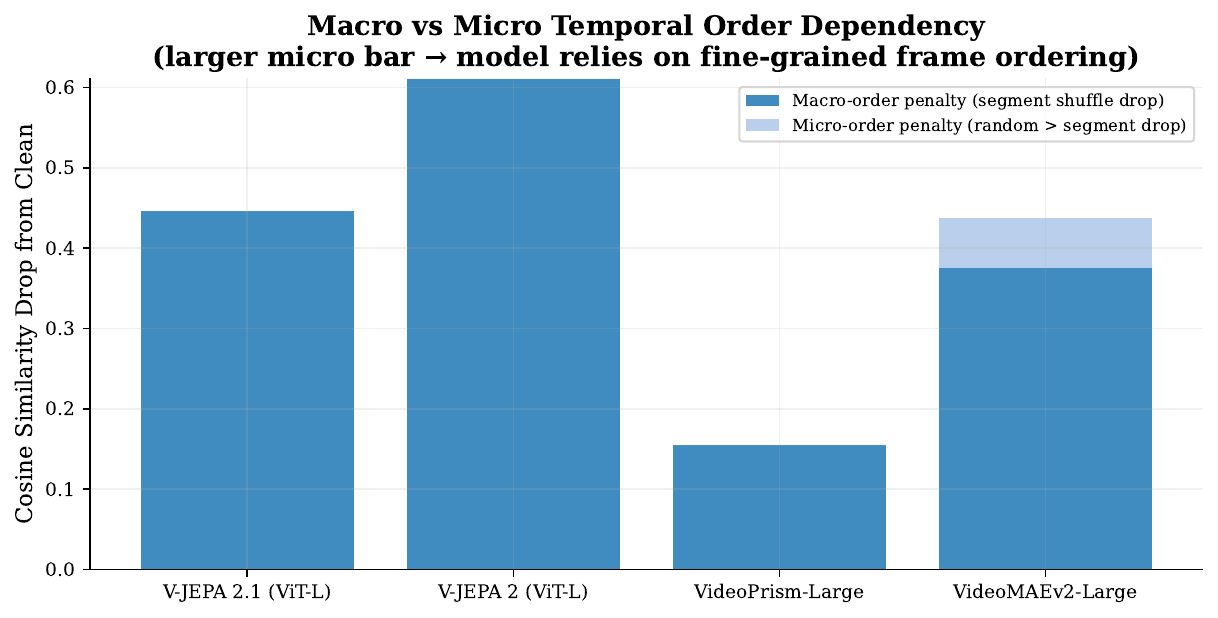}
  \caption{Decomposition of temporal order dependency into macro (segment
  shuffle drop) and micro (additional random shuffle drop) penalties. All
  models are predominantly sensitive to macro-level disruption.}
  \label{fig:app_macro_micro}
\end{figure}

Figure~\ref{fig:app_macro_micro} decomposes the total cosine similarity drop
into macro and micro penalties. The macro component dominates for every
encoder, confirming that segment-level coherence, not frame-to-frame order,
is the primary temporal cue.

\subsection{Static video and spatial grounding}
\label{app:temporal_static}

\begin{figure}[H]
  \centering
  \includegraphics[width=\linewidth]{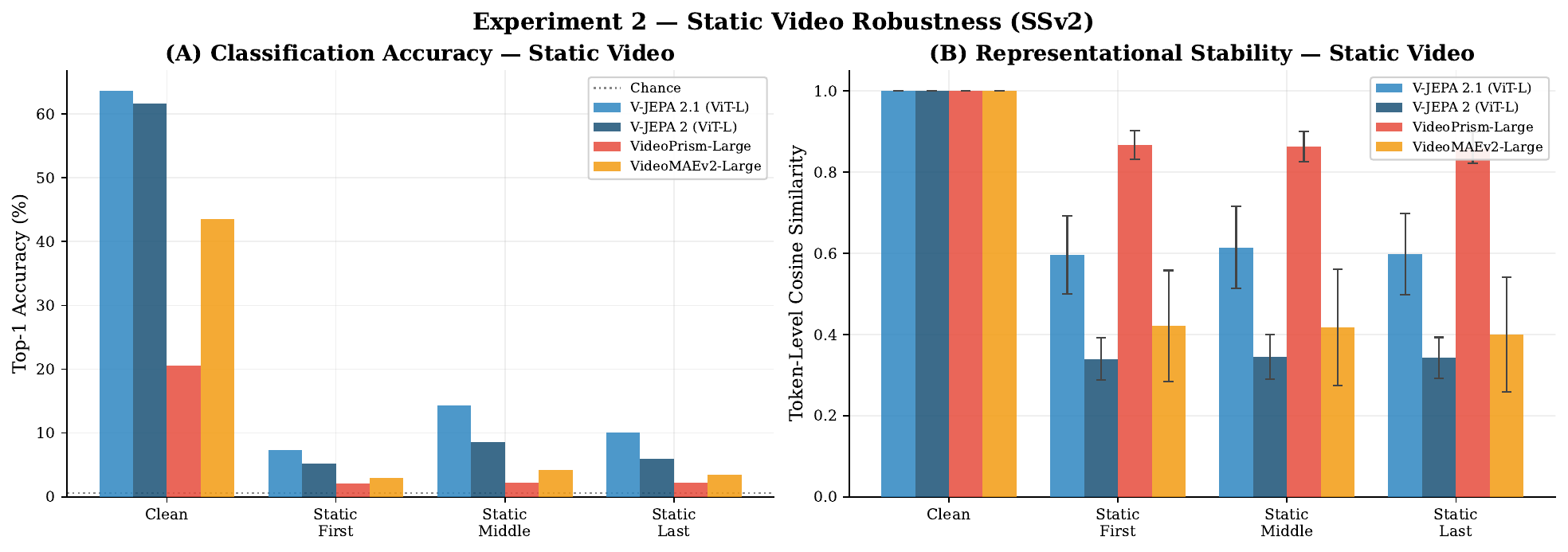}
  \caption{Accuracy and cosine similarity under single-frame static input.
  VideoPrism shows minimal representational change yet near-chance accuracy,
  confirming a spatial-only encoding.}
  \label{fig:app_static}
\end{figure}

\begin{figure}[H]
  \centering
  \includegraphics[width=0.85\linewidth]{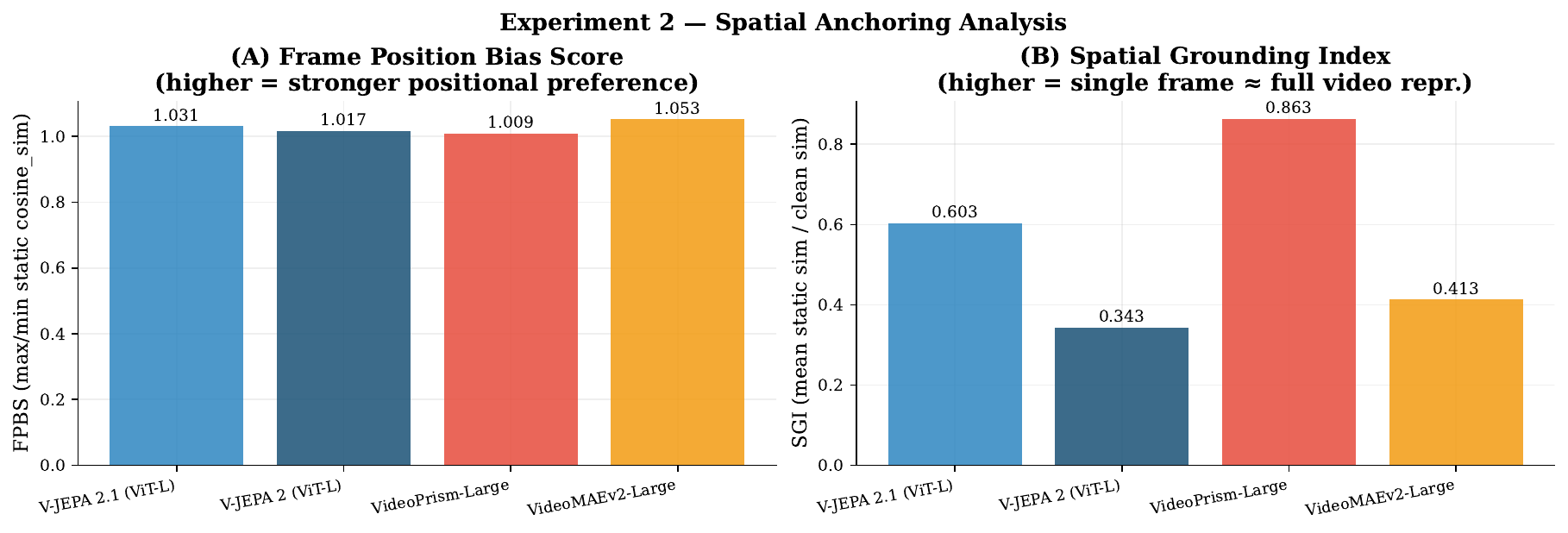}
  \caption{Frame Position Bias Score (left) and Spatial Grounding Index
  (right). V-JEPA\,2.1 anchors to the middle frame; VideoMAEv2 and VideoPrism
  anchor to the first.}
  \label{fig:app_sgi}
\end{figure}

Both quantities are computed from the three static conditions. The Spatial
Grounding Index is the mean cosine similarity between the clean embedding and
the static-video embedding, averaged over the first-, middle- and last-frame
variants, so a high SGI means a single still frame reproduces most of the
latent code. The Frame Position Bias Score is the ratio of the largest to the
smallest of those three similarities, so a value near one means no frame
position is privileged.

V-JEPA\,2.1's middle-frame anchoring (Figure~\ref{fig:app_sgi}) reflects a
temporally integrative representation. VideoPrism's SGI of 0.863 confirms
that a single frame accounts for nearly all of its latent code, against 0.603
for V-JEPA\,2.1 and 0.343 for V-JEPA\,2.

\subsection{Pure noise and temporal consistency}
\label{app:temporal_noise}

\begin{figure}[H]
  \centering
  \includegraphics[width=\linewidth]{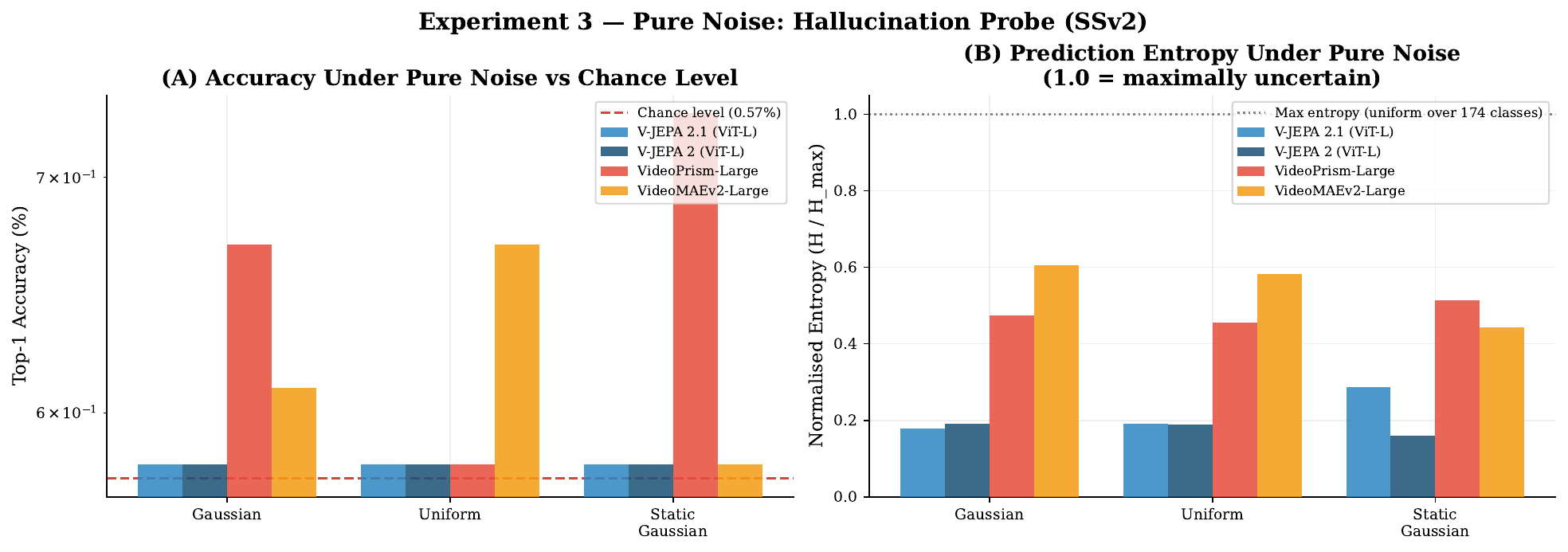}
  \caption{Accuracy and entropy under three noise types. All models fall to
  chance, ruling out semantic hallucination. Representational responses diverge
  sharply: VideoMAEv2 collapses to near-zero cosine similarity while
  VideoPrism's remains above 0.77.}
  \label{fig:app_noise}
\end{figure}

\begin{figure}[H]
  \centering
  \includegraphics[width=0.85\linewidth]{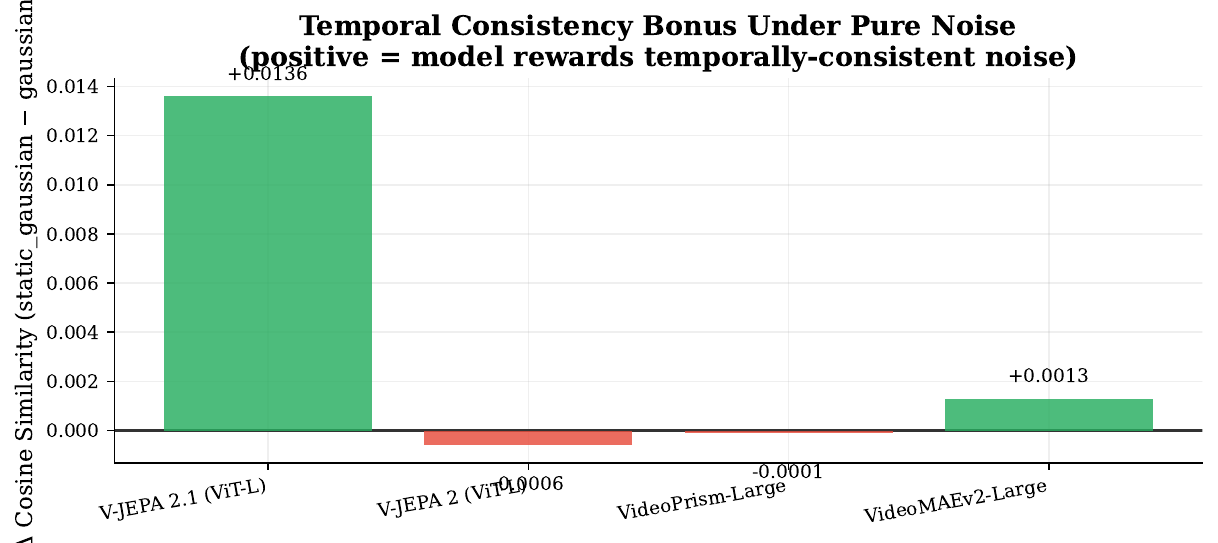}
  \caption{Temporal Consistency Bonus: the difference in cosine similarity between
  temporally static and temporally varying noise. V-JEPA\,2.1 is the only encoder
  that rewards frame-to-frame coherence appreciably in the absence of semantic
  content.}
  \label{fig:app_consistency_bonus}
\end{figure}

The Temporal Consistency Bonus is defined as $\cos_{\mathrm{static\ noise}} -
\cos_{\mathrm{varying\ noise}}$: a positive value means the encoder's latent
space distinguishes a temporally coherent input from an incoherent one even
when neither carries semantic content. V-JEPA\,2.1 is the only encoder with an
appreciable bonus, $+$0.014 against $+$0.001 for VideoMAEv2 and approximately
zero for V-JEPA\,2 and VideoPrism
(Figure~\ref{fig:app_consistency_bonus}), indicating that its latent space
rewards frame-to-frame coherence as a structural prior inherited
from predictive pre-training. VideoPrism's near-zero bonus reflects
insensitivity to temporal structure. VideoMAEv2's near-zero cosine similarity
under noise (Figure~\ref{fig:app_noise}) exposes fragility rather than
temporal sensitivity.

\subsection{Video reversal and directional coherence}
\label{app:temporal_reversal}

\begin{figure}[H]
  \centering
  \includegraphics[width=\linewidth]{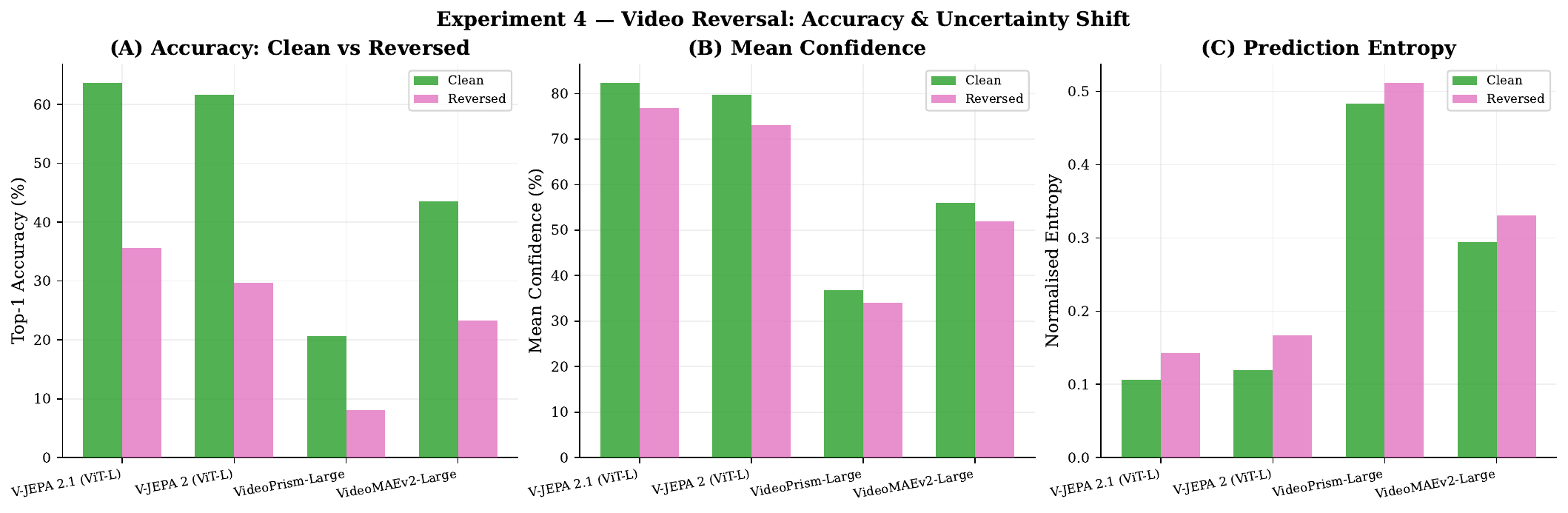}
  \caption{Accuracy, confidence, and entropy under video reversal. The
  V-JEPA models retain the highest accuracy; VideoPrism shows the largest
  entropy increase.}
  \label{fig:app_reversal}
\end{figure}

\subsection{Cross-architecture summary}
\label{app:temporal_summary}

\begin{figure}[H]
  \centering
  \includegraphics[width=0.85\linewidth]{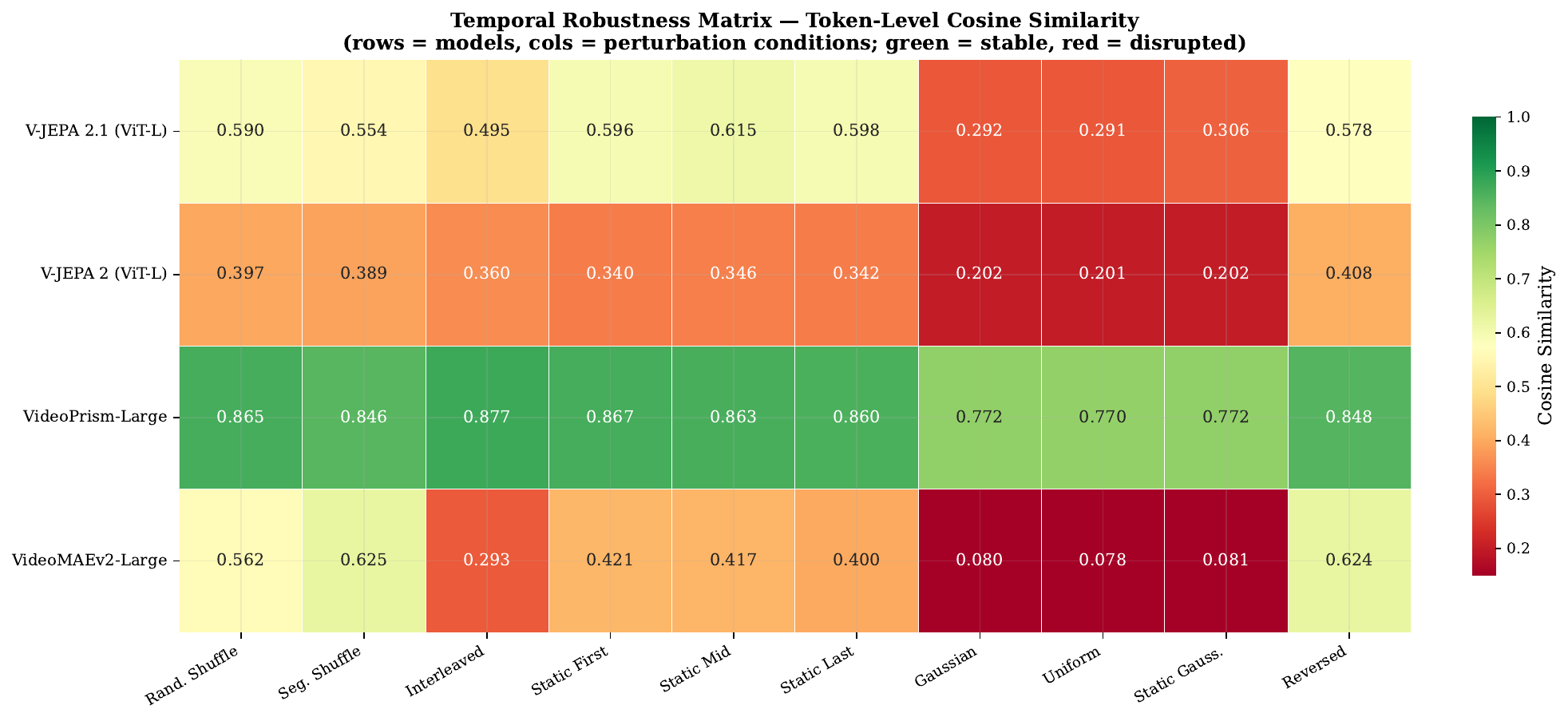}
  \caption{Full cross-architecture robustness heatmap. Green cells denote
  high cosine similarity (stable representation); red cells denote collapse.
  VideoPrism's uniformly green row reflects spatial rigidity.
  VideoMAEv2's red noise columns reflect fragility.}
  \label{fig:app_heatmap}
\end{figure}

\begin{figure}[H]
  \centering
  \includegraphics[width=0.75\linewidth]{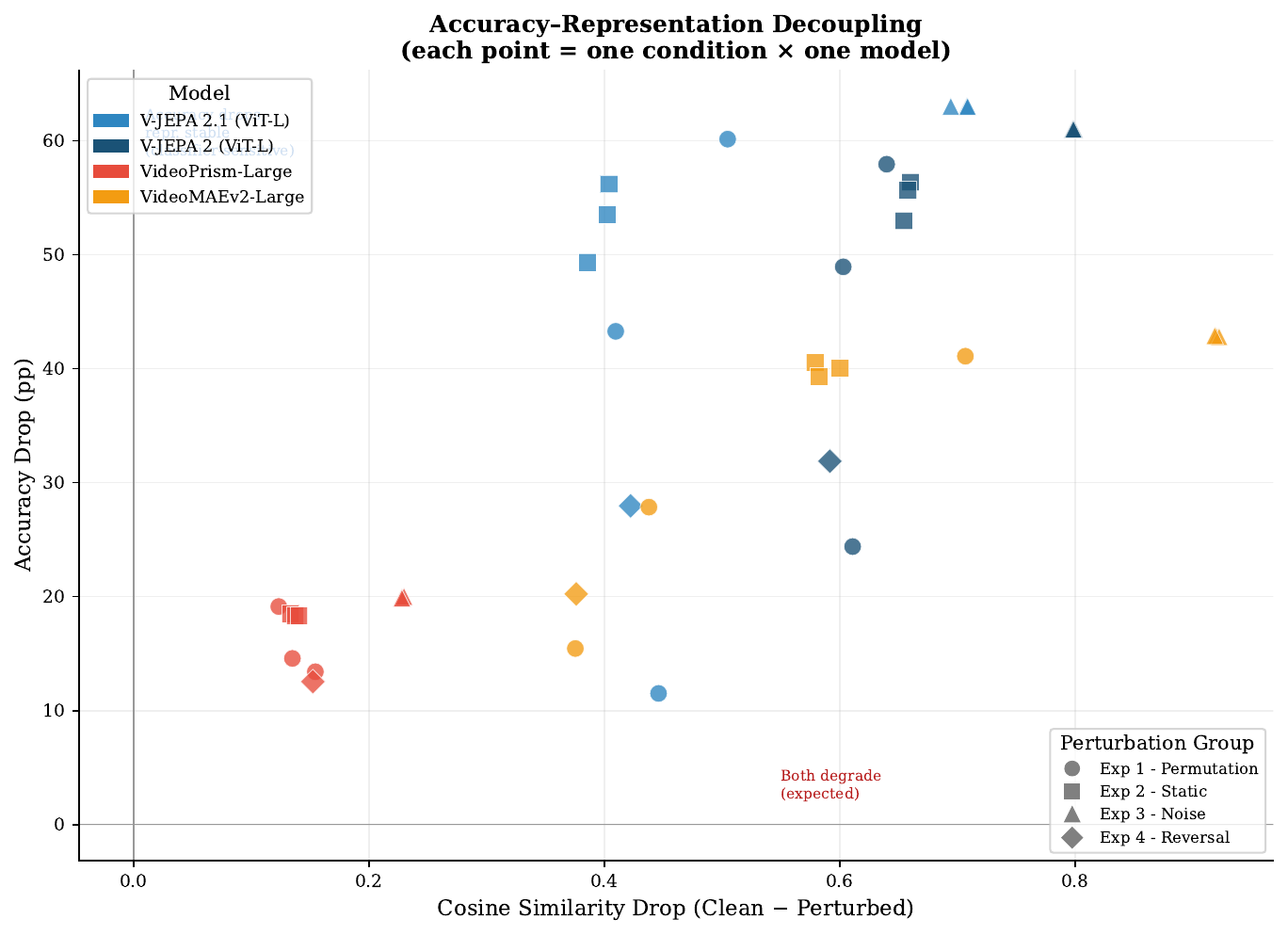}
  \caption{Accuracy drop versus cosine similarity drop across all conditions
  and models. VideoPrism clusters at low cosine drop but high accuracy drop,
  confirming that its stable representations do not support stable decisions.}
  \label{fig:app_decoupling}
\end{figure}

\section{Downstream world modelling: supplementary analysis}
\label{app:downstream}

This appendix documents the closed-loop experiment of Section~\ref{sec:downstream}: the
protocol, the validation checks that precede it, the full per-condition results,
the controls that establish what the planner is actually using, and the limitations.

\subsection{Environment, tasks and protocol}
\label{app:downstream_protocol}

We use Meta-World~\citep{yu2020meta}, the standard benchmark for evaluating frozen real-video
encoders as control representations, so that real-video-to-simulation transfer is an
established setting rather than an assumption of this study. Expert demonstrations are
generated by the scripted policies the benchmark ships with, removing any dependence on an
external robot dataset. Two contact-rich manipulation tasks, \texttt{drawer-open} and
\texttt{door-close}, are used for predictor training and evaluation, and a third task,
\texttt{window-open}, is held out entirely for the zero-shot arm. Every planning cell runs 30
episodes per task (40 for the held-out task).

Each encoder is frozen. A window of the last 16 environment steps is resampled to the encoder's
native clip length, and tokens are kept at each encoder's native spatial grid (576, 256, 196 and
256 tokens) rather than pooled to a common size, so that no encoder is compressed more than
another. Latents are z-scored using training-split statistics only, which prevents a single
shared learning rate from suiting one latent scale and handicapping another.

\begin{table}[H]
\centering
\small
\caption{Predictor and planner configuration. One configuration is used for every encoder; no
per-encoder tuning is performed.}
\label{tab:downstream_config}
\begin{tabular}{ll}
\toprule
\textbf{Component} & \textbf{Setting} \\
\midrule
Predictor & 4-layer causal transformer, $d=256$, 4 heads ($\approx3.2$M parameters) \\
Optimisation & inherited from the attentive-probe protocol (Appendix~\ref{app:probe_training}) \\
Planner & cross-entropy method, 100 samples, 10 elites, 3 iterations \\
Planning horizon & 5, replanning every 3 steps, warm-started, 200 step cap \\
Episodes & 30 per task per cell; 40 for the held-out task \\
\bottomrule
\end{tabular}
\end{table}

Because all four encoders emit $D=1024$ features, the predictor has an identical parameter count
in every case. A sensitivity check at $d=384$ was run to confirm that the ordering does not
depend on this choice.

\paragraph{Harness bounds.} Every cell is bracketed by two references: replaying the scripted
expert through the same evaluation loop, and acting randomly. The expert reaches $1.00$ and the
random policy $0.00$ for all four encoders on all three tasks, so a low planning score reflects
the world model rather than broken environment or goal plumbing.

\subsection{Validation: can these encoders see the scene at all?}
\label{app:downstream_gate}

Before any planning, we check that encoders pretrained on real internet video can localise the
arm and object in a rendered simulation. A linear probe decodes ground-truth end-effector and
object position from frozen features, reported in millimetres against two references: predicting
the dataset mean, and decoding from raw pixels.

\begin{table}[H]
\centering
\small
\caption{Linear state decodability from frozen features (millimetres, lower is better). Every
encoder beats the pixel reference on both targets, so the simulation domain gap does not make
the comparison uninformative.}
\label{tab:downstream_gate0}
\begin{tabular}{lcc}
\toprule
\textbf{Representation} & \textbf{Hand (mm)} & \textbf{Object (mm)} \\
\midrule
Dataset mean (reference) & 210.5 & 189.0 \\
Raw pixels (reference)   & 33.6  & 84.1 \\
\midrule
V-JEPA 2.1 & 12.3 & 11.4 \\
V-JEPA 2   & 11.4 & \textbf{9.8} \\
VideoMAEv2 & 18.4 & 18.3 \\
VideoPrism & 21.1 & 20.8 \\
\bottomrule
\end{tabular}
\end{table}

All four encoders clear the pixel reference by a factor of $1.6$--$4.0$ on object position, and
the two latent-prediction encoders lead. This is a validation check, not a finding: it
establishes that the downstream comparison measures world-model quality rather than an inability
to perceive the scene.

\subsection{Offline predictor quality}
\label{app:downstream_stage1}

Before closing the loop we measure the predictor itself. Rollout error is the decoded object
position error after $k$ predicted steps; the \emph{oracle} column decodes the true future
latent, so the \emph{gap} isolates what the predictor contributed from what the encoder
supports. Directional fidelity is the fraction of rollouts that move the object the correct way,
the downstream form of the arrow-of-time axis in Section~\ref{sec:temporal}.

\begin{table}[H]
\centering
\small
\caption{Offline latent rollout quality. Errors in millimetres, lower is better.}
\label{tab:downstream_stage1}
\begin{tabular}{lccccc}
\toprule
\textbf{Encoder} & \textbf{$k$} & \textbf{Rollout (mm)} & \textbf{Oracle (mm)} & \textbf{Gap} & \textbf{Dir. fidelity} \\
\midrule
V-JEPA 2.1 & 1  & 9.5  & 8.1  & 1.4  & 0.823 \\
V-JEPA 2.1 & 5  & 19.2 & 8.6  & 10.5 & 0.960 \\
V-JEPA 2.1 & 10 & 29.7 & 9.1  & 20.6 & 0.944 \\
V-JEPA 2   & 1  & 9.2  & 7.7  & 1.4  & \textbf{0.867} \\
V-JEPA 2   & 5  & \textbf{14.5} & 8.3 & 6.3 & 0.965 \\
V-JEPA 2   & 10 & \textbf{19.7} & 8.6 & 11.1 & 0.963 \\
VideoMAEv2 & 1  & 15.7 & 15.1 & 0.7  & 0.766 \\
VideoMAEv2 & 5  & 22.3 & 15.8 & 6.5  & 0.933 \\
VideoMAEv2 & 10 & 31.0 & 16.5 & 14.6 & 0.944 \\
VideoPrism & 1  & 16.3 & 15.4 & 1.0  & 0.702 \\
VideoPrism & 5  & 24.1 & 16.6 & 7.5  & 0.965 \\
VideoPrism & 10 & 33.5 & 17.8 & 15.7 & 0.957 \\
\bottomrule
\end{tabular}
\end{table}

The oracle columns separate the two model families cleanly at every horizon, showing that the
latent spaces themselves differ in how well object state can be recovered, before any predictor
is involved. Both V-JEPA encoders also lead on single-step directional fidelity, the metric most
closely aligned with the arrow-of-time result of Section~\ref{sec:temporal}.

\subsection{Full closed-loop results}
\label{app:downstream_stage2}

\begin{table}[H]
\centering
\small
\caption{Planning success per condition. Perturbations are impulse noise and snow at severity 5
and spatiotemporal patch dropout at severity 3, matching the families used in
Sections~\ref{sec:corruption} and~\ref{sec:occlusion}. Each cell is 60 episodes.}
\label{tab:downstream_stage2}
\begin{tabular}{lccccc}
\toprule
\textbf{Encoder} & \textbf{Clean} & \textbf{Impulse} & \textbf{Snow} & \textbf{Occlusion} & \textbf{Mean perturbed} \\
\midrule
V-JEPA 2.1 & 0.983 & \textbf{0.417} & \textbf{0.467} & \textbf{0.500} & \textbf{0.461} \\
V-JEPA 2   & 0.967 & 0.183 & 0.267 & 0.217 & 0.222 \\
VideoMAEv2 & 0.350 & 0.000 & 0.017 & 0.000 & 0.006 \\
VideoPrism & 0.317 & 0.033 & 0.000 & 0.167 & 0.067 \\
\bottomrule
\end{tabular}
\end{table}

Clean success is at ceiling for both V-JEPA encoders: 59/60 and 58/60 are indistinguishable from
a perfect run at this episode count (Fisher exact $p=1.00$), so the clean column does not order
them however the decimals happen to sort. What clean success does establish is the split between
the two groups: the closest cross-group pair separates at $p=1.3\times10^{-13}$. The ordering in
Section~\ref{sec:downstream} therefore rests on the perturbed conditions, where all three
adjacent pairs separate, at $p=2.6\times10^{-6}$, $3.6\times10^{-5}$ and $2.9\times10^{-3}$ from
top to bottom.

The two tasks respond differently to perturbation. Grasping a drawer handle is
precision-limited and degrades sharply once the observation is corrupted, whereas pushing a
door closed tolerates substantial visual corruption, so a task set with more tolerant contact
geometry would raise every number in the table. The ordering between encoders is preserved
within each task.

\subsection{Is the world model doing the work?}
\label{app:downstream_control}

A planner that follows visual subgoals could in principle succeed by matching latents greedily,
without the learned dynamics contributing anything. We test this directly by re-running the
identical planner with a \emph{randomly initialised} predictor, holding the encoder, subgoals,
cost, search budget and latent normalisation fixed, so any difference is attributable to the
learned dynamics alone.

\begin{table}[H]
\centering
\small
\caption{Trained versus untrained predictor, per encoder, condition and task. Reported per task
rather than averaged, because the task average conceals the one cell where the dynamics are not
required. $p$ is Fisher exact.}
\label{tab:downstream_control}
\begin{tabular}{lllccl}
\toprule
\textbf{Encoder} & \textbf{Condition} & \textbf{Task} & \textbf{Trained} & \textbf{Untrained} & \textbf{$p$} \\
\midrule
V-JEPA 2.1 & snow  & door-close  & 28/30 & 0/30  & $8.4\times10^{-15}$ \\
V-JEPA 2   & clean & drawer-open & 30/30 & 0/30  & $1.7\times10^{-17}$ \\
V-JEPA 2   & clean & door-close  & 28/30 & 27/30 & $1.00$ \\
VideoMAEv2 & clean & drawer-open & 14/30 & 0/30  & $1.7\times10^{-5}$ \\
VideoMAEv2 & clean & door-close  & 7/30  & 0/30  & $1.1\times10^{-2}$ \\
\bottomrule
\end{tabular}
\end{table}

The control confirms that the results we rely on are results about world models. Under snow,
\texttt{door-close} drops from 28/30 to 0/30 when the predictor is untrained: the cell that
carries the perturbed ranking is entirely dependent on the learned dynamics, which is what
licenses reading that ranking as a measure of world-model quality. The same holds for both
VideoMAEv2 cells and for V-JEPA\,2 on \texttt{drawer-open}. One cell behaves differently:
clean \texttt{door-close} is solved just as well with an untrained predictor, because pushing
a door closed from a clean observation admits a greedy latent-matching solution. That cell is
not evidence of world modelling and we do not treat it as such. It is also the reason the
paper's ordering is drawn from the perturbed conditions rather than the clean ones, where the
task itself is easy enough to mask the difference between encoders.

\subsection{Zero-shot transfer}
\label{app:downstream_zeroshot}

\begin{table}[H]
\centering
\small
\caption{Planning on \texttt{window-open}, absent from every predictor's training set, 40
episodes. Retention is zero-shot success relative to in-distribution clean success.}
\label{tab:downstream_zeroshot}
\begin{tabular}{lccc}
\toprule
\textbf{Encoder} & \textbf{Zero-shot} & \textbf{In-distribution} & \textbf{Retention} \\
\midrule
V-JEPA 2.1 & \textbf{0.875} & 0.983 & \textbf{89\%} \\
V-JEPA 2   & 0.850 & 0.967 & 88\% \\
VideoMAEv2 & 0.000 & 0.350 & 0\% \\
VideoPrism & 0.125 & 0.317 & 39\% \\
\bottomrule
\end{tabular}
\end{table}

This arm is deliberately kept separate from the perturbed averages, since it answers a different
question. Both V-JEPA encoders carry roughly nine tenths of their in-distribution success onto a
mechanism the predictor never saw, while VideoMAEv2 transfers nothing despite solving 21 of 60
in-distribution episodes. The two adjacent pairs within each group do not separate
(35/40 versus 34/40, $p=1.00$; 5/40 versus 0/40, $p=0.055$), so this arm establishes the split
but does not rank within it.

\subsection{Agreement with the representation-level axes}
\label{app:downstream_agreement}

The temporal axes are the ones that predict downstream behaviour most directly. DSCS and
reversal cosine, both measured on frozen features in Section~\ref{sec:temporal}, reproduce the
encoder ordering of offline rollout error at $k=5$ and of single-step directional fidelity
(Table~\ref{tab:downstream_stage1}) exactly. This is the correspondence the paper predicts: an
encoder that represents the direction of time supports a predictor that moves objects the
right way.

The paradigm-level split is what transfers most robustly. Every representation-level axis
separates the latent-prediction encoders from the rest, and so does every closed-loop
measurement: latent prediction supports planning under degraded input and zero-shot transfer,
and the alternatives largely do not.

\subsection{Limitations}
\label{app:downstream_limitations}

The planner is demonstration-guided: it follows visual subgoals taken from the expert
trajectory, which is more privileged information than a single goal image. This is held
identical across encoders, so the comparison is fair, but the absolute success rates should not
be read as goal-image-conditioned planning. The study uses two training tasks and one held-out
task in a single simulated environment, and two of the six corruption families from
Section~\ref{sec:corruption}. Success rates are binomial with 60 episodes per cell (180 pooled
across perturbations), so differences of one or two episodes are not resolved; the group-level
split we draw conclusions from is far larger than that, but adjacent checkpoints within a
group need not be. Finally, as the control in Appendix~\ref{app:downstream_control} shows,
clean success is a weaker probe of world modelling than the perturbed conditions we rely on.

\end{document}